\pdfoutput=1
\documentclass[11pt]{article}

\usepackage{acl}

\usepackage{times}
\usepackage{latexsym}

\usepackage[T1]{fontenc}
\usepackage[utf8]{inputenc}

\usepackage{microtype}
\usepackage{mathptmx} % <--- better than "times"
\usepackage{latexsym}
\usepackage{xcolor}
\usepackage{stackengine}

\usepackage{subfigure}
\usepackage{algorithm2e}
\usepackage{setspace}
\usepackage{breqn}
\usepackage{bbm}
\usepackage{dirtytalk}
\usepackage{graphicx}
\usepackage{float}
\usepackage{listings}
\usepackage{booktabs}
\usepackage{caption}
\usepackage{enumitem}
\usepackage{amsmath,stackengine}
\usepackage{cleveref}
\usepackage{float}
\stackMath

\usepackage{microtype}

\newcommand{\RefSection}[1]{\hyperref[#1]{\S\,\ref{#1}}}
\newcommand{\RefFigure}[1]{\hyperref[#1]{Figure~\ref{#1}}}
\newcommand{\RefTable}[1]{\hyperref[#1]{Table~\ref{#1}}}

\title{Local Structure Matters Most: Perturbation Study in NLU}
\author{Louis Clouâtre\textsuperscript{1,3}
  Prasanna Parthasarathi\textsuperscript{2,3}
  Amal Zouaq\textsuperscript{1} and
  Sarath Chandar \textsuperscript{1,3,4} \\
  \textsuperscript{1} Polytechnique Montréal\\
  \textsuperscript{2} School of Computer Science, McGill University\\
  \textsuperscript{3} Quebec Artificial Intelligence Institute (Mila) \\
  \textsuperscript{4} Canada CIFAR AI Chair 
}
\date{}
\DeclareUnicodeCharacter{2212}{-}
\begin{document}
\maketitle

\begin{abstract}

Recent research analyzing the sensitivity of natural language understanding models to word-order perturbations has shown that neural models are surprisingly insensitive to the order of words.
In this paper, we investigate this phenomenon by developing order-altering perturbations on the order of words, subwords, and characters to analyze their effect on neural models' performance on language understanding tasks.
We experiment with measuring the impact of perturbations to the local neighborhood of characters and global position of characters in the perturbed texts and observe that perturbation functions found in prior literature only affect the global ordering while the local ordering remains relatively unperturbed.
We empirically show that neural models, invariant of their inductive biases, pretraining scheme, or the choice of tokenization, mostly rely on the local structure of text to build understanding and make limited use of the global structure.

\end{abstract}

\section{Introduction}

% Large pretrained (PT) models have become an inevitability in modern Natural Language Processing (NLP) applications~\citep{Transformers, BERT, radford2019language, RoBERTa, lewis-etal-2020-bart, brown2020language, ELMO, ULM-FIT}. 
% Pretraining techniques such as Masked Language Modeling (MLM) and Causal Language Modeling (CLM) have aided in making the best use of language statistics learned from large corpora to achieve improved performance on several NLP benchmarks~\citep{glue2, glue8, glue7, glue3, glue4, GLUE, SuperGLUE,glue1,glue5,glue6,glue9,glue10,glue11,SQUADv1,SQUADv2}. 
% With the increase in applications of these large models also came a growing interest in evaluating the way these models learn to solve natural language tasks. 

Recent research has shown that neural language models have an understanding of well-formed English syntax in recurrent neural networks, convolutional neural networks, and in large pretrained (PT) Transformers~\citep{gulordava2018colorless,zhang2018language, chrupala-alishahi-2019-correlating, lin-etal-2019-open, belinkov-glass-2019-analysis,liu2019linguistic,jawahar-etal-2019-bert,rogers2020primer}.
Other studies, however, take a critical stance with experiments suggesting that models may be insensitive to word-order perturbations~\citep{pham2020out,sinha2021masked,sinha2020unnatural,  gupta2021bert,o2021context}, showing that shuffled word-order has little to no impact during training or inference with neural language models. 
While some research show that models learn some abstract notion of syntax, further probing into their insensitivity to the perturbation of syntax is necessary.
Specifically, \emph{What are the underlying mechanisms causing those unintuitive, or unnatural, results from neural models} is still a largely unanswered question.

Recent research exploring the sensitivity to syntax of pretrained models has primarily been applying perturbations to text through perturbing the order of words~\citep{pham2020out,sinha2021masked,sinha2020unnatural,  gupta2021bert,o2021context}.
Perturbations applied and quantified at this granularity of text offer only a limited understanding of the learning dynamics of the neural language models.
Analyzing perturbations at a finer granularity such as subwords~\citep{bojanowski2017enriching} or characters~\citep{gao2018black,ebrahimi2017hotflip}, may provide a deeper insight into the insensitivity to word-order of neural models.
% Consider \autoref{fig:sample-perturbations}, which shows an unperturbed sentence, a word-level perturbed sentence, a subword-level perturbed sentence, and a character-level perturbed sentence. 
% An average reader may find it possible to parse and infer the meaning of the word-level perturbed sentence but would have issues inferring any meaning from subword and character-level perturbed text. 

In this paper, we define two types of structure\footnote{Structure here relates to the organization of characters in the text.} in text, global which relates to the absolute position of characters, and local, which relates to the relative position of characters to their immediate neighbors.
We observe from our experiments (\RefSection{sec:experiments}) that most perturbations proposed and analyzed in the literature will perturb the global structure with different reordering of words, while the amount of disturbance to the local structure remains limited.
\emph{We hypothesize that the local structure, more so than the global structure, is necessary for understanding in natural language tasks.}
By applying perturbations of varying degrees to the local structure, while controlling for the amount of global perturbations, we are able to measure how essential it is to a neural model understanding of text.
We demonstrate the sensitivity to local structure of model performances in English natural language understanding (NLU) (GLUE~\citep{GLUE}) and their relative insensitivity to the global structure, and control for many potential confounding factors that would otherwise provide an alternative explanation to our results.
%\RefFigure{fig:pretrained_metrics} shows some of our experimental results, demonstrating that DND is very correlated with a models degradation in performance, and compares very favorably to other traditional metrics such as the Levenshtein distance and the BLEU score as an explanation of the degradation in performance.

Our contributions are as follows:

\begin{itemize}
    \item We show that the performance of neural models -- Transformers and others, pretrained or not -- on perturbed input strongly correlates with the amount of preserved local structure of text.
    \item We identify possible confounding factors for this phenomenon and construct experiments controlling for them.
    \item We provide analysis on implications derived from our large array of empirical findings.

    % \item We observe that DND has a strong correlation with GLUE scores across different architectures, suggesting that neural language understanding models generally are sensitive to distortions in local structures more so than global structures.
    % \item We show that commonly used lexical perturbations distort the global structures and seldom affect the local structures explaining the insensitivity of large models to such perturbations, and showing that intuitions relating to the importance of the order of words in neural language models may not be appropriate. 
    % \item We show that DND has a weak correlation to other metrics -- BLEU, Levenshtein -- indicating that the common metrics used for evaluation do not measure the dimension captured by DND.
    % \item We find that the lack of correlation to performance of non-pretrained Transformers to IDC is useful in detecting when models do not make use of the positional information present in text, defaulting to bag-of-word models.
\end{itemize}

\section{Related Work}
\label{sec:related-work}

\paragraph{Importance of Syntax} 
Discussions on 
semantics~\citep{
culbertson2014language, futrell2020dependency} 
%peters1973generative,pinker1979formal
agree on specific orders of words to be necessary for comprehending text.
Psycholinguistic research~\citep{hale2017models} corroborates this through evaluating sentence comprehension mechanisms of humans. 
Hence, interpreting language as a bag-of-words could limit the expressions conveyed through the word-orders \cite{harris1954distributional,le2014distributed} and understanding syntax\footnote{Preference to a specific word-order over the other and the preference complying with the choices of an average human speaking that language.} becomes an essential artifact.
Recently, \citet{composition_neurobiology} found that humans were robust to word-ordering perturbations in text as long as local ordering of text was roughly preserved.

% Will add those back in, need to see  where they fit
% \citet{warstadt-etal-2019-investigating, richardson2020probing} show that model can understand the grammar of language by learning to use the syntactic structures.  

Prior works have explored the relationship between neural models and syntax.
\citet{assessing_syntax_1, assessing_syntax_2} both show that BERT~\citep{BERT} models have some syntactic capacity.
\citet{BertHierachicalSyntaxicRepresentation} show that BERT represents information hierarchically and concludes that BERT models linguistically relevant aspects in a hierarchical structure.
\citet{syntax1, syntax2} show that the contextual embeddings that BERT outputs contain syntactic information that could be used in downstream tasks.

While it seems that syntax is both important, and to an extent, understood by the recent family of PT models, it is unclear how much use they make of it.
\citet{SyntaxSupervisedTraining} showed that pretraining BERT on syntax does not seem to improve downstream performance much.
\citet{LearningWhichFeaturesMatter} showed that while models such as BERT do understand syntax, they often prefer not to use that information to solve tasks.
\citet{ettinger-2020-bert, pham-etal-2019-improving, sinha2020unnatural, gupta2021bert} show that large language models are insensitive to minor perturbations highlighting the lack of syntactic knowledge used in syntax rich NLP tasks. 
\citet{sinha2021masked} show that pretraining models on perturbed inputs still obtain reasonable results on downstream tasks, showing that models that have never been trained on well-formed syntax can obtain results that are close to their peers.
% \citet{sinha2020unnatural} show that models can perform better with specific perturbed word order inputs. % Need to explain how this relates to syntax here

While syntactic information seems vital to language, and large PT models seem to be at least aware of syntax, the lack of sensitivity of neural models to perturbation of syntax motivates further probing.

\paragraph{Text Perturbations} 
Several different types of reordering perturbation functions and schemes have been explored to understand and study neural architectures' (in)sensitivity to word-order.
The class of perturbation analysis could broadly be split into three categories: deletion, paraphrase injection, and reordering of tokens.  
\citet{sankar-etal-2019-neural} explore utterance and word-level perturbations applied to generative dialogue models to highlight their insensitivity to the order of conversational history.
On natural language classification tasks, \citet{pham2020out} define $n$-grams for different values of $n$ and shuffle them to highlight the insensitivity of PT models.
They show that shuffling larger $n$-grams has a lesser effect than shuffling smaller $n$-grams, suggesting that preserving more local structure causes less performance degradation.
Studying textual entailment tasks, \citet{sinha2020unnatural} perform perturbations on the position of the words, with the criteria that no word remains in its initial position.

\citet{hsieh-etal-2019-robustness} propose a suite of adversarial attacks that replace one word in the input to cause a model to flip its correct prediction.  
\citet{gupta2021bert} combine several types of destructive transformations --- such as sorting, reversing, shuffling words --- towards removing all informative signals in a text.
Along similar lines, \citet{wang-etal-2019-improving} inject noise by reordering or deleting articles towards injecting artificial noise to measure the robustness of PT language models. 
Character-level perturbations that perform minimal flips to cause a degenerate response have been explored by \citet{ebrahimi2017hotflip, gao2018black}. 
\citet{gao2018black} quantify the perturbation in Levenshtein distance and draw a correlation to the model's performance. 
This work is closely related to our own.
We demonstrate that our hypothesis, the importance of local ordering, is a much more robust explanation of the degradation in performance of models than the Levenshtein distance.

% Although the recent literature on perturbation analysis in PT language models was able to observe the extremes-- sensitivity and insensitivity, understanding the attributes of text to which neural models are sensitive too requires a detailed study. 
%Going against the recent trend of $n$-gram perturbations or word-level explanations,
% We speculate that the perturbation analyses done at the granularity of sub-words and characters is necessary for properly probing the neural models' insensitivity to word-order phenomenon. 

\paragraph{Quantifying Perturbations}
Several popular similarity metrics can be used to measure perturbations.
Metrics like BLEU~\citep{papineni2002bleu} and ROUGE~\citep{lin2004rouge} will treat text as a sequence of words, from which a measure of overlap is computed.
The Levenshtein distance~\citep{OG_Levenshtein, yujian2007normalized}, or the \textit{edit} distance, measures the minimum amount of single-character edits (insertions, deletions, or substitutions) necessary to match two strings together.
In the context a shuffling text, it will roughly count the amount of characters that have been displaced.
\citet{parthasarathi2021sometimes} observed that learned metrics like BERT-Score~\citep{zhang2019bertscore} and BLEURT~\citep{sellam-etal-2020-bleurt} are often unaffected by minor perturbations in text which limits their usefulness in measuring perturbations. 
% \citet{sinha2020unnatural} propose a POS mini-tree overlap score to interpret the results of the perturbation analysis. 
% The score computes the part-of-speech (PoS) tags neighborhood for every word and estimates an average overlap in the neighborhood for all the tokens before and after applying the  perturbation. 
% The authors, however, find that the working range of the proposed metric to be small and explain the effect only for PT Transformer architectures.
Character-level metrics, such as the character $n$-gram F-score (chrF)~\citep{chrf} offer a character-aware approach to measuring similarity of $n$-gram overlap between two texts.
In the context of shuffling this, this will represent roughly the amount of character $n$-gram that have been changed by the shuffling.

\section{Measuring Local and Global Pertubations}

To properly analyze different perturbations to the local and global structure of text, we first require a way to measure perturbations to said structures.
The \textit{global} structure here relates to the absolute position of characters in a text, and the \textit{local} structure relates to the neighboring character of any other character in a text. 

\subsection{Character bigram F-score (chrF-$2$)}
To measure local perturbations, we use the chrF~\citep{chrf} metric. 
chrF is an $n$-gram overlap metric that is applied to characters.
The goal here is to isolate the smallest unit of local structure that we can quantify, character $2$-grams being preserved after perturbations.
We therefore use a minimal and maximal $n$-gram length of $2$.
We use the default $\beta$ value of $3$.
Our metric is equivalent to calculating the F$3$-score of character $2$-gram overlap between the unperturbed text and the perturbed text, taking whitespaces into account.

\subsection{IDC}
To measure the global perturbations, we introduce the \textbf{I}ndex \textbf{D}isplacement \textbf{C}ount (IDC) metric, which measures the average distance traversed by every character after perturbations.

Let a string, $x_i = (c)_k^i$, be denoted by a sequence of characters $c_0, \ldots, c_k$, where $k$ is the length of the string in characters and $p^{x_i}$ denote the positions of characters in $x_i$. 
Let $\eta(\cdot)$ be a perturbation operation.

\begin{equation}
  x_i'\gets \eta\left(x_i\right),
\end{equation}

where $x_i'$ denote the perturbed string with positions of the characters specified by $p^{x'_i}$.

\begin{equation}
    IDC \gets \frac{1}{k^2}\sum_{j=1}^{k}\left\Vert p^{x'_i}\left(j\right) - p^{x_i}\left(j\right) \right\Vert_{1}
\end{equation}

The denominator $k^2$ normalizes the average by the length of the text\footnote{$k^2$ is used to normalize as we sum $k$ times a number that is between $0$ and $k$, where $k$ is the text length.}.
Intuitively, an IDC of 0.3 would imply that characters in the perturbed text have moved 30\% of the text length on average.
The values of IDC will lie in the range $\left[0,0.5\right]$, where $0.5$ would be obtained by reversing a text at the character level. 
% \subsection{Levenshtein Distance (Lev)}
% To measure perturbations to the global structure, we will use the Levenshtein Distance, normalized by the length of the text.
% This measure of perturbation only takes into account the change in absolute positions of characters and not its local neighborhood, which makes it ideal to measure the changes to the global structure.
% With the normalization, a levenshtein distance of 0.8 would imply that 80\% of characters need to be edited once to retrieve the original text.

% \begin{equation}
%   x_i'\gets \eta\left(x_i\right),
% \end{equation}

% where $x_i'$ denote the perturbed string with positions of the characters specified by $p^{x'_i}$.

% \begin{equation}
%     IDC \gets \frac{1}{k^2}\sum_{j=1}^{k}\left\Vert p^{x'_i}\left(j\right) - p^{x_i}\left(j\right) \right\Vert_{1}
% \end{equation}

% The denominator $k^2$ normalizes the average by the length of the text\footnote{$k^2$ is used to normalize as we sum $k$ times a number that is between $0$ and $k$, where $k$ is the text length.}.
% Intuitively, an IDC of 0.3 would imply that characters in the perturbed text have moved 30\% of the text length on average.
% The values of IDC will lie in the range $\left[0,0.5\right]$, where $0.5$ would be obtained by reversing a text at the character level. 

\subsection{Compression Rate (Comp)}
Finally, to measure local perturbations to words and subwords, we could count the rate of out-of-vocabulary (OOV) tokens introduced by the perturbations.
As our experiments make use of a subword vocabulary~\citep{BPE} which can represent any string of English characters without OOV tokens, the compression rate~\citep{xue2021byt5}, as measured by the length of the original string in characters divided by the length of the tokenized string, will serve as a proxy to measuring OOV tokens.
As more local perturbations are applied, more and more subwords will be broken into smaller subwords which will yield a lesser compression of text through tokenization.
The tokenizer of the RoBERTa-Base model~\citep{RoBERTa} is used to calculate the compression rate in all cases. 

\section{Perturbation Functions}
\label{sec:Perturbations}

Towards conducting a detailed analysis on the effect of perturbations on the performance of neural language models, we define three granularities of perturbation functions --- \emph{word-level}, \emph{subword-level} and \emph{character-level}. 
The subwords are taken from the RoBERTa-Base vocabulary.
We define the perturbation functions as generic operations that can be applied across the different levels of granularity\footnote{Pseudo-code and examples for all perturbations are shown in Appendix \ref{app:pseudocode_perturbations}.}.

\paragraph{Full Shuffle} randomly shuffles the position of every word, sub-word, or character, according to the level it is applied to. 
This transformation should cause a great amount of perturbation to the \textit{global} and \textit{local} structure for the specific granularity.

\begin{figure}[ht]
    \centering
    \fbox{\begin{minipage}{0.9\columnwidth}
    \setstretch{1.5}
    \parbox{\columnwidth}{
    	    \centering
    
            \large
            
            \textcolor{red}{The} \textcolor{blue}{scholar} \textcolor{green}{is} \textcolor{black}{typesetting}.
            
            \textcolor{blue}{scholar}
            \textcolor{black}{typesetting}
            \textcolor{green}{is}
            \textcolor{red}{The}.
        }
     \end{minipage}
     }
    \caption{Example for word-level full shuffling. 
    The perturbed sentence has a IDC of 0.29 and a chrF-$2$ of 0.92.}
    \label{fig:sample-perturbations_word_full_shuffle}
\end{figure}

\paragraph{Phrase Shuffle} creates chunks of contiguous tokens of variable length, controlled by a parameter $\rho$, and shuffles the phrases of word, subword, or characters.
This perturbation has, on average, the same impact as the full shuffling on the \textit{global} structure as the absolute positions of characters tend to change just as much as full shuffling while preserving a controllable amount of \textit{local} structure.
%, since phrases maintain more of such structure.
\begin{figure}[ht]
    \centering
    \fbox{\begin{minipage}{0.9\columnwidth}
    \setstretch{1.5}
    \parbox{\columnwidth}{
    	    \centering
          
            \large
            
            \textcolor{red}{The} \textcolor{blue}{scholar} \textcolor{green}{is} \textcolor{black}{typesetting}.
            
            \textcolor{green}{is} \textcolor{black}{type}\textcolor{red}{The} \textcolor{blue}{scho}\textcolor{black}{setting} 
            \textcolor{blue}{lar}.
           
        }
     \end{minipage}
     }
    \caption{Subword-level phrase shuffling.
    The perturbed sentence has an IDC of 0.35 and a chrF-$2$ of 0.84. 
    }
    \label{fig:sample-perturbations_subword_n_gram}
\end{figure}

% Phrase shuffling uses a parameter $\rho$ that controls the average size of the randomly defined contiguous chunks of tokens. 
To randomly define our phrases, we traverse the text sequentially on the desired granularity. 
The entire text is assumed as a single large phrase and is truncated at a token with probability $\rho$ into smaller phrases.

A lower value of $\rho$ leads to longer on average phrases, thus preserving more of the \emph{local} structure while destroying roughly the same amount of \emph{global} structure. 
In the extreme case with $\rho = 1.0$, phrase shuffling will be equivalent to full shuffling as phrases will all be one token long.
%of length $1$.

\paragraph{Neighbor Flip Perturbations} flip tokens of the chosen granularity with the immediate right neighbor with probability, $\rho$.
This function has, on average, a smaller impact on the \emph{global} structure, as the absolute positions of tokens do not change much but can have an arbitrary large effect on disturbing the \emph{local} structure.

\begin{figure}[ht]
    \centering
    \fbox{\begin{minipage}{0.9\columnwidth}
    \setstretch{1.5}
    \parbox{\columnwidth}{
    	    \centering
            \large
            
            \textcolor{red}{The} \textcolor{blue}{scholar} \textcolor{green}{is} \textcolor{black}{typesetting}.
            
            \textcolor{red}{heT} \textcolor{blue}{cshlor}\textcolor{green}{i s}\textcolor{blue}{a} \textcolor{black}{typeesttnig}.
           
        }
     \end{minipage}
     }
    \caption{Character-level neighbor flip.
    The perturbed sentence has an IDC of 0.04 and a chrF-$2$ of 0.32. 
    Due to a greater distortion to the local order, the model has a greater chance to be sensitive to this perturbation.}
    \label{fig:sample-perturbations_char_neighbour}
\end{figure}

The perturbation is applied by traversing the string from left-to-right on the desired granularity and, with a probability $\rho$, switching the current attended token with the following token. 
The lower the $\rho$ is, the less perturbation happens, thus preserving more of the \textit{local} structure. 
This transformation has a limited impact on the \textit{global} metric, thus letting us isolate the impact of perturbations to the different structures.

\section{Experiments}\label{sec:experiments}

\subsection{Dataset}
We experiment with the GLUE Benchmark~\citep{GLUE} datasets, a popular NLU benchmark.
% Of the GLUE's suite, we evaluate on 8 tasks -- Multi-Genre NLI (MNLI), Corpus of Linguistic Acceptability (CoLA), Quora Question Pairs (QQP), Microsoft Research Paraphrase Corpus (MRPC), Question NLI (QNLI), Recognizing Textual Entailment (RTE), Stanford Sentiment TreeBank (SST),  Semantic Textual Similarity Benchmark (STS-B)--~\citep{glue1, glue2, glue3, glue4, glue5, glue6, glue7, glue8}. 
% The model agnostic tasks in GLUE that have diverse textual contexts, dataset sizes, and varying degrees of difficulty are designed to evaluate language understanding components of neural language models.
We create perturbed versions of the validation set for all tasks with the different perturbation functions defined in~\RefSection{sec:Perturbations}.
In total, 50 different variations of our perturbation functions are applied by varying the granularity as well as the $\rho$ values, including an unperturbed benchmark version\footnote{The hyperparameters used for the perturbation functions are detailed in Appendix \ref{app:experiments}.}.

\subsection{Confounding Variables}
We have identified several confounding variables that we will attempt to control for in our experimental setup.

\paragraph{Inductive Biases} of the neural architecture may yield models that rely on different types of structure.
Intuitively, it may be that Transformer-based models, through global self-attention, rely more on global structure than ConvNets which are limited to local information.
% We will experiment with BiLSTMs, Transformers, and ConvNets to have an appropriate breadth of inductive biases.

\paragraph{Pretraining} may have a large impact on the level of sensitivity to different types of structure. 
It may be that global structure simply requires more training to be understood and that pretrained models leverage it to a much higher degree than non-pretrained (NPT) models.
The specific method used for pretraining may also impact the sensitivity to different types of structures, such as adding permutations to the pretraining objectives.
% Therefore, we will experiment with both PT and NPT Transformers and three different flavors of PT Transformers to compare various pretraining schemes.

\paragraph{Tokenization} schemes may be the most significant confounding variable.
By perturbing the local ordering of characters, we also perturb the vocabulary of models that rely on the precise order of characters.
% It is also likely that the amount of tokens perturbed would correlate heavily with our measure of local structure.
% Additional experiments to understand which of the two factors, local structure perturbation and vocabulary perturbation, cause the degradation in performance and which is simply a byproduct of the other will be required.
% We will experiment with character-level tokenization to verify whether the removal of a vocabulary impacts the performance decay.
% If character-level models exhibit the same type of correlations between local perturbations and downstream performance without any use of words, it is safe to assume that the cause for the degradation in performance is not the destruction of learned word-embeddings.
% Experiments with word and subword perturbations also serve to isolate the impact of local structure perturbations from vocabulary perturbation, as they do not impact the vocabulary while still impacting the local structure somewhat. 

\begin{figure*}[h!t]
    \centering
    \includegraphics[width=0.35\textwidth]{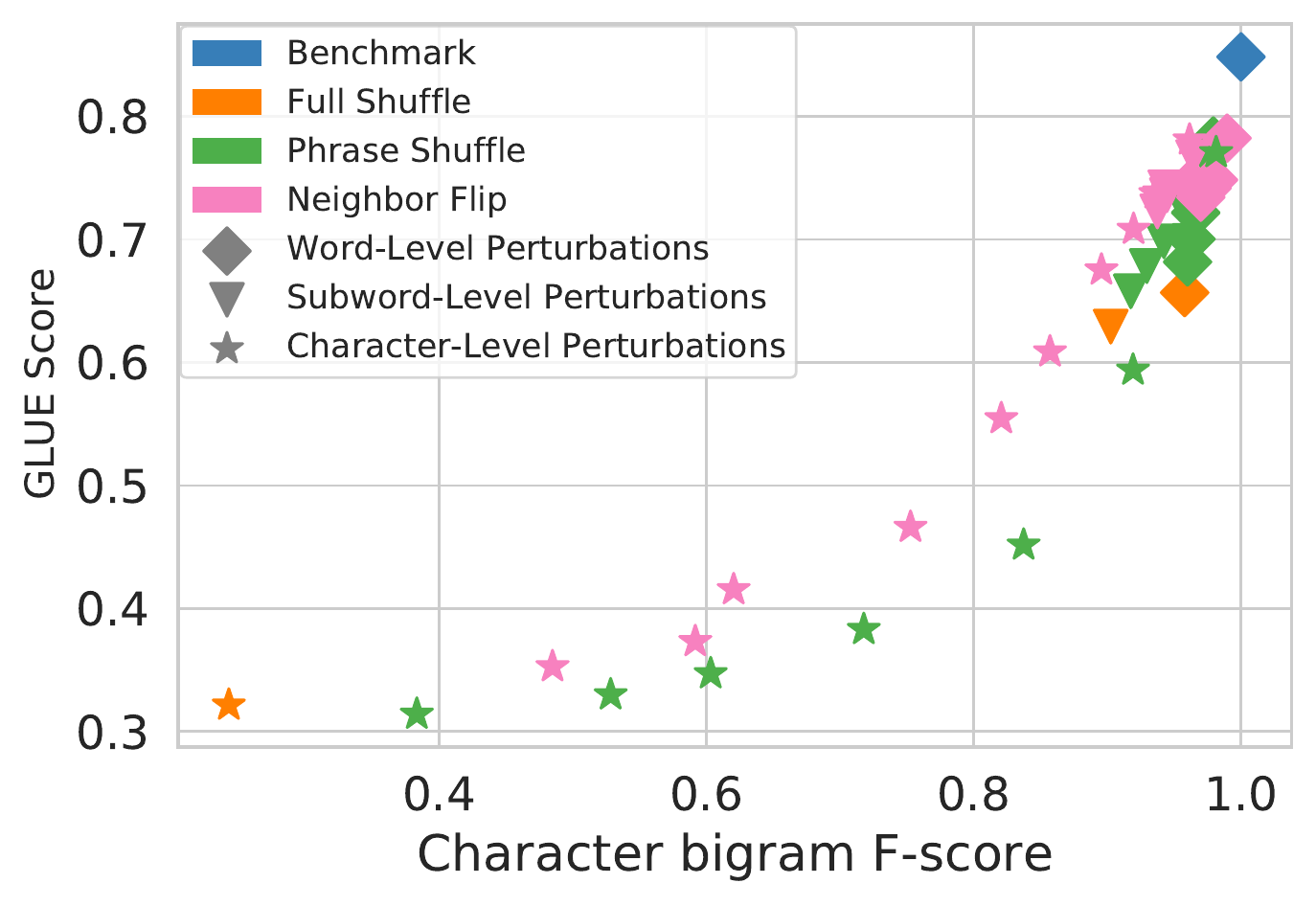}\hfill
    \includegraphics[width=0.31\textwidth]{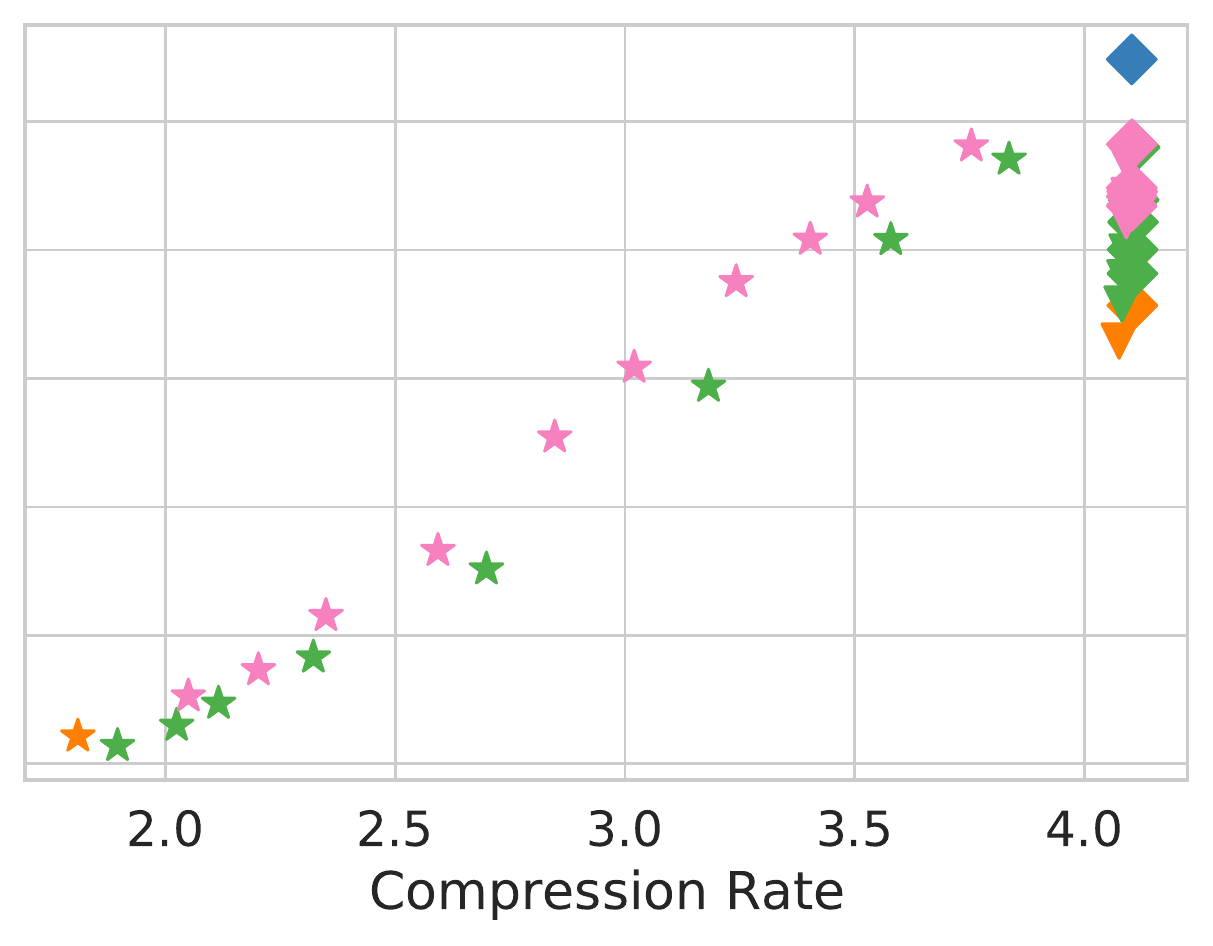}\hfill
    \includegraphics[width=0.31\textwidth]{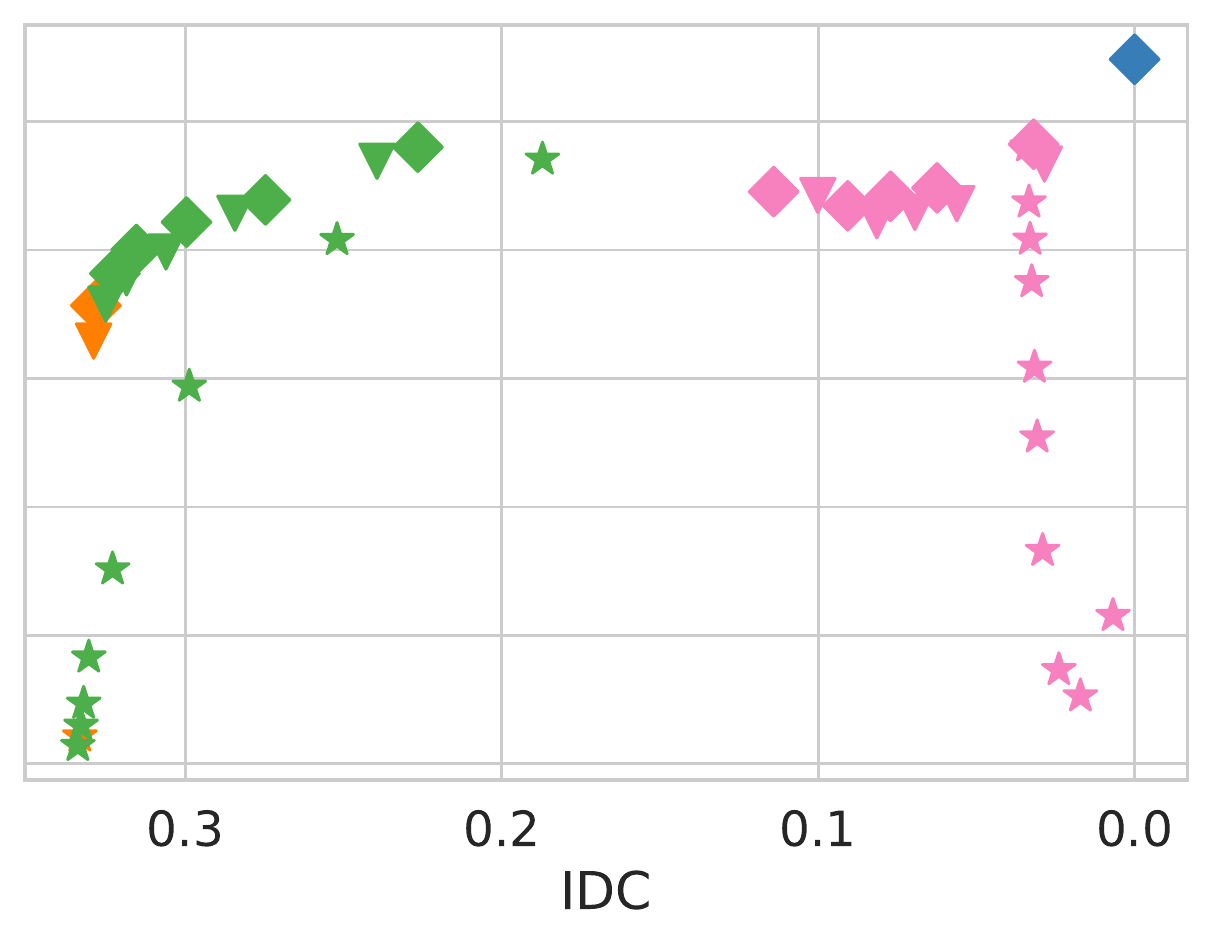}

    \caption{
    Plotted are the relations between the different choices of metrics measuring the amount of perturbation and the performance of PT RoBERTa-Base model tested on the perturbed data.
    Left is more perturbed, up is better performance.
    The X-axis of the IDC metric is inverted for clearer comparison.
    }
    \label{fig:pretrained_transformer_metrics}
\end{figure*}

\subsection{Models}
We experiment with BiLSTMs~\citep{Bi-LSTM}, Transformers~\citep{Transformers}, and ConvNets to have an appropriate breadth of neural inductive biases.
We experiment with three flavor of PT Transformers (RoBERTa-Base~\citep{RoBERTa}, BART-Base~\citep{BART} and CharBERT-Base~\citep{CharBERT}), and a NPT Transformer (RoBERTa-Base architecture) to verify the impact of pretraining. 
We also experiment with different tokenization schemes, using byte-pair encoding (BiLSTMs, ConvNet, RoBERTa-Base, BART-Base, NPT Transformer) as well as character-level tokenization (BiLSTMs, ConvNet, CharBERT-Base~\citep{CharBERT}), to isolate the impact of the destruction of a model's vocabulary.
 
The tokenization for PT Transformer models use their corresponding vocabulary, while NPT models (BiLSTM, ConvNet, Transformer) use the RoBERTa-Base vocabulary and the character-level models use characters exclusively as vocabulary\footnote{The CharBERT model uses a mix of characters and subword vocabulary.}.
Training is done once on the unperturbed dataset until convergence and evaluation is done on the perturbed version of the validation datasets. The training details can be found in Appendix \ref{app:experiments}. 

\section{Analysis}

\subsection{Metrics and GLUE Performance}
\label{sec:metric-vs-glue}

% \begin{figure}[h]
%     \centering
%     \includegraphics[width=0.95\columnwidth]{./res/img/AllModelPlaceholder_unreadable.pdf}
%     \caption{Comparison of different neural architectures' performances with different level of perturbation as measured by DND.}
%     \label{fig:DND-GLUE-AllModels}
% \end{figure}

We compute the average GLUE score of different models applied to the validation data perturbed with our different perturbation functions.
The PT RoBERTa-Base results are plotted in \RefFigure{fig:pretrained_transformer_metrics}\footnote{Results for all individual models can be found in Appendix \ref{app:all_results}}.

\begin{figure}[ht]
    \centering
    \includegraphics[width=0.95\columnwidth]{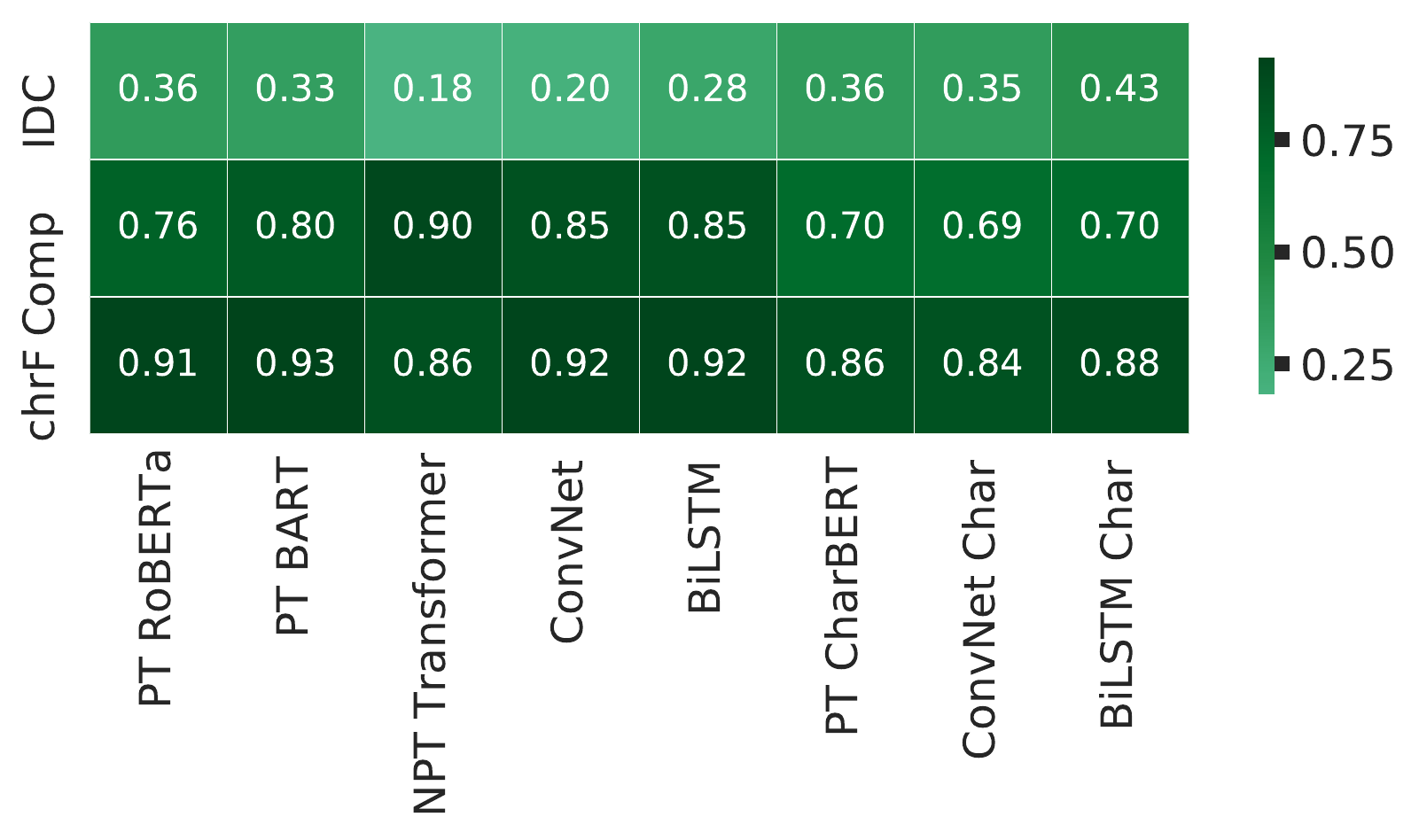}
    \caption{Rank correlation matrix between the models' performance to perturbed samples on the GLUE benchmark and the perturbation quantified by the different metrics.
    The higher the value the better the metric explains the degradation in performance.}
    \label{fig:models-metric-correlation}
\end{figure}

First, we observe that word and subword-level perturbations are very limited in their impact on the local structure, but can affect the whole spectrum of global structure.
We observe the general trend that the chrF-$2$ metric strongly correlates with neural models' loss in performance on the GLUE benchmark tasks across all perturbations and granularity of perturbations. 
While the IDC metric correlates somewhat with performance, it fails to distinguish between neighbor flipping perturbations and phrase shuffle perturbations.
The compression rate is strongly correlated with performance on character-level perturbations but does not hold explanatory power for word and subword-level perturbations, as they do not affect the vocabulary, leading to the overall lower rank correlation with performance degradation.
% The compression rate also correlates well for part of the experiments but act much differently for the character-level perturbations and the word and subword-level perturbations.

By computing the rank correlation between the GLUE score of the different models on the perturbed samples and the metric measuring the perturbations (\RefFigure{fig:models-metric-correlation}), we see that the correlation of GLUE score with the chrF-$2$ metric holds for every single architecture and setting tested.
On the other hand, the IDC metric is only weakly correlated with performance decay.
This implies that local structure, more so than global structure, is necessary for models to perform NLU.
A model being evaluated on a perturbed text with a chrF-$2$ of $0.7$ can be assumed to have much lower performance than on a perturbed text with a chrF-$2$ of $0.95$, irrespective of the granularity or the type of perturbations that yielded those metrics.
This is not true of any of the other metrics.

\begin{figure}[ht]
    \centering
  
    \includegraphics[width=0.95\columnwidth]{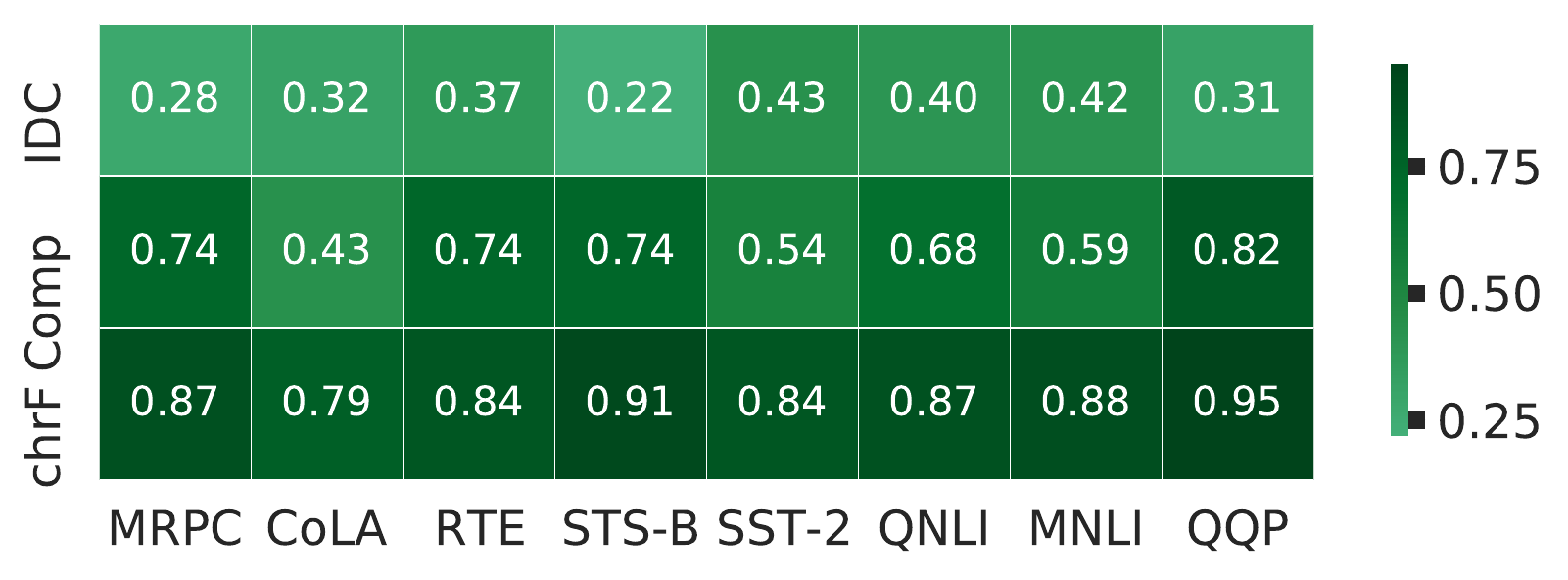}
    \caption{Rank correlation matrix between perturbations measured by different metrics and the performance on the different GLUE tasks of the PT RoBERTa model.
    }
    \label{fig:task_to_scores_roberta}
\end{figure}

Looking at the individual tasks more closely, as in \RefFigure{fig:task_to_scores_roberta}, we see that the conclusions regarding the overall GLUE benchmark do hold for every task individually.

% However, a few 
% The semantic acceptability task, CoLA, correlates very strongly with the BLEU-4 metric and only weakly with chrF-$2$.
% It is also the only task in GLUE benchmark that reaches chance-level performance with word-level perturbations~\citep{pham2020out}.
% Hence, word-level perturbations, as measured by BLEU, being more important for the task than the local structure, as measured by DND, is expected.

% The STS-B task, a semantic textual similarity task, is barely correlated with IDC but strongly correlated with DND.
% It does not seem possible to affect this task performance with word or subword-level perturbations, implying that the bag-of-words information is sufficient to obtain good textual similarity estimates.
% DND, being able to measure distortions to the bag-of-words information, is able to provide an explanation for degradation in performance on a task that does not seem to rely on syntactic information.
% Whereas IDC does not seem to provide an explanation for the decay of performance in this task.

\subsection{Effect of Perturbations on Metrics}

As intended, the different perturbations have different impact on our metrics, as shown in \RefFigure{fig:pretrained_transformer_metrics}.
Thee neighbor flip perturbations objective was to obtain an arbitrary amount of local perturbation for a relatively small amount of global perturbation.
We can observe that the IDC metric, which measures the impact to the global structure, is smaller for the neighbor flip than for the phrase shuffle, even when the amount of local perturbation, as measured by the chrF-$2$ metric, is roughly equivalent.
The compression rate is closely tied to the measure of local structure on character-level perturbations, but is static  for word and subword perturbations as the tokens are never impacted.

\subsection{Correlation between metrics}

\begin{figure}[ht]
    \centering
  
    \includegraphics[width=0.8\columnwidth]{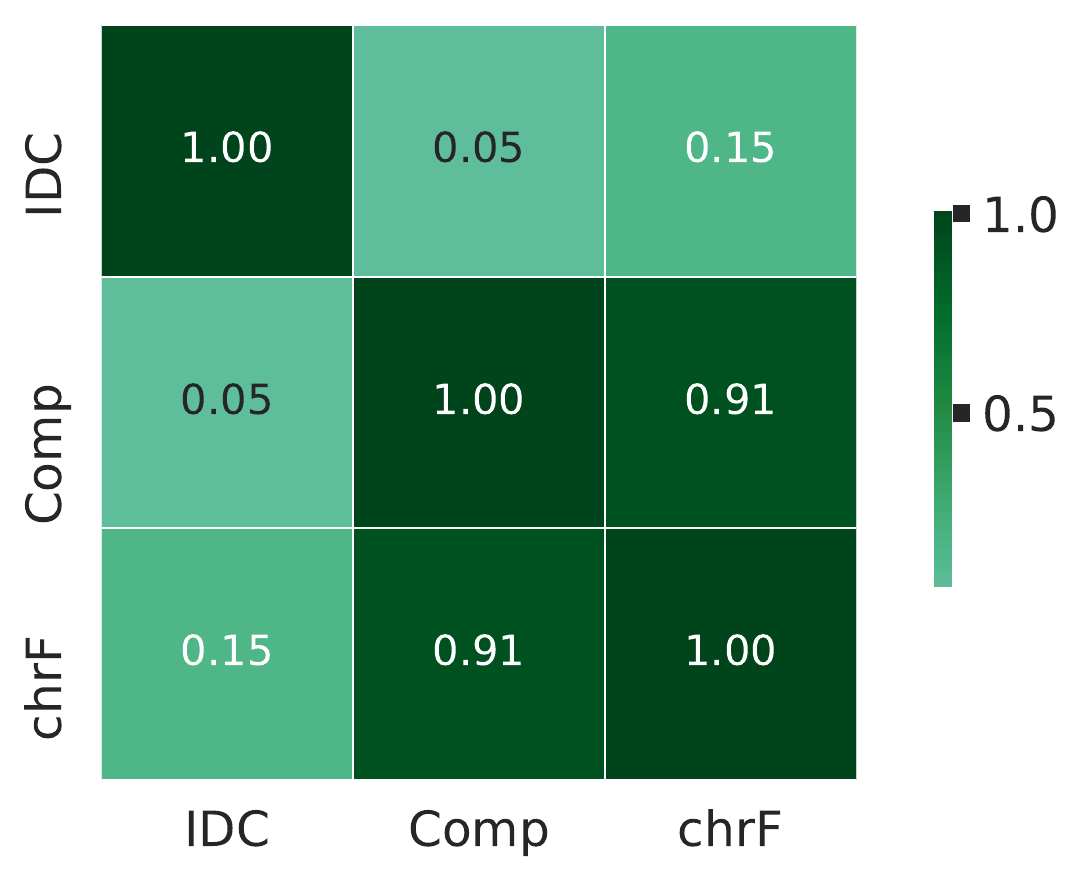}
    \caption{Correlation matrix between the different metrics on the GLUE tasks.
    }
    \label{fig:metrics-correlation}
\end{figure}

To confirm that the chrF-$2$ metric and the IDC metric do measure orthogonal aspects of structure, we compute their pairwise pearson correlation in the GLUE validation set in \autoref{fig:metrics-correlation}\footnote{For every correlation, we inverted the value of the IDC metric by flipping its signs to make the comparison of the different correlations more straightforward.
It is a measure of perturbation and not similarity and is therefore inversely correlated to the GLUE score and the other metrics.}
We also include the compression rate. 
Specifically, for every sample in the validation set of the GLUE tasks, we perturb them using the different perturbation functions and compute their scores with the different metrics. 

We observe that chrF-$2$ and IDC have a fairly low correlation, suggesting that the metrics measure different aspects of the perturbations. 
We also observe a very high correlation between the chrF-$2$ measure and the compression rate, which motivates experiments that perturb one without impacting the other to isolate the main component causing the performance degradation.

% \subsection{Comparison of Perturbation Functions}

% We populate an assorted list of 16 word-level linguistically motivated perturbation functions described and analyzed by \citet{parthasarathi2021sometimes} that can be applied to examples in GLUE tasks. 
% % The 16 different word-level perturbations are categorized as PoS-Tag perturbations, Dependency Tree perturbations, and Random shuffles that include perturbing with different traversal orders of a text dependency tree, such as Pre-Order, Post-Order or In-Order, swapping verbs, adverbs, nouns in a text, reversing sentences among other perturbations.
% We perturb the samples across GLUE tasks for every perturbation function, compute the IDC and DND scores and compare with the perturbations defined in \S~\ref{sec:Perturbations}. 
% The distribution of scores computed  by the metrics for the different perturbations functions shown in \autoref{fig:spectrum-perturbation-functions} indicates that the word-level and subword-level perturbations do not disturb the local structure as measured by the DND metric. 
% As the metric is observed to strongly correlate with model performance on perturbed samples (in \S \ref{sec:metric-vs-glue}), 
% the analysis provides a reasonable explanation to the insensitivity observed due to word-level perturbations studied in the literature~\citep{sinha2020unnatural,pham2020out,gupta2021bert}. 

\subsection{Model specific analysis}

The loss in performance of models in GLUE tasks shows a greater degree of correlation with the chrF-$2$ metric than any other metric, as shown in \RefFigure{fig:models-metric-correlation}, with the exception of the NPT Transformer which we discuss in \RefSection{sec:npt_embeddings}.
% We found our results consistent across PT Transformers, ConvNets, BiLSTMs, subword tokenization, and character tokenization.
% This indicates that our results generalize to neural language models across different inductive biases, pretrained or not, and to different pretraining techniques.

\subsubsection{Pretrained vs Non-Pretrained models}

\RefFigure{fig:models-metric-correlation} demonstrates that perturbations to the local structure explain much of the degradation in performance for both PT and NPT models.
Despite the different pretraining schemes used, the PT RoBERTa and BART model have a comparable level of degradation across the different perturbations, showing that the choice of pretraining scheme has a relatively small impact on perturbation resistance.

All NPT models exhibit a strong correlation between the chrF-$2$ metric and their degradation in performance on the GLUE tasks, which indicates that the sensitivity to local structure is not an artifact of pretraining.
% In contrast with PT models, NPT models have very low correlations between all metrics and performance on the RTE task.
% This is explained by the fact that most NPT models do not obtain significantly above chance-level performance on the RTE task.
% As the performance quickly degrades to chance-level once any perturbation is applied, it is hard to measure correlations between the metrics and the task performance.

\subsubsection{NPT Transformer and Positional Embeddings}\label{sec:npt_embeddings}

Interestingly, the NPT Transformer bucks the overall trend by having very little correlation between its performance and IDC and being more correlated to the compression rate than to the chrF-$2$ metric.
As IDC will roughly measure the distance traversed by characters from their initial position, it having little correlation with performance in NPT Transformers implies that the absolute position of tokens is not taken into account by the NPT Transformers. 
We hypothesize that learning the positional embeddings requires much more data than is present in a single NLU task, leading the NPT model to act as a bag-of-words model.
This would explain why perturbations to the vocabulary are so impactful to the NPT Transformer, as it is unable to correct minor disturbances in words with the context of neighboring words.

\begin{figure*}[h!t]
    \centering
    \subfigure[PT Transformer]{
    \includegraphics[width=0.37\textwidth]{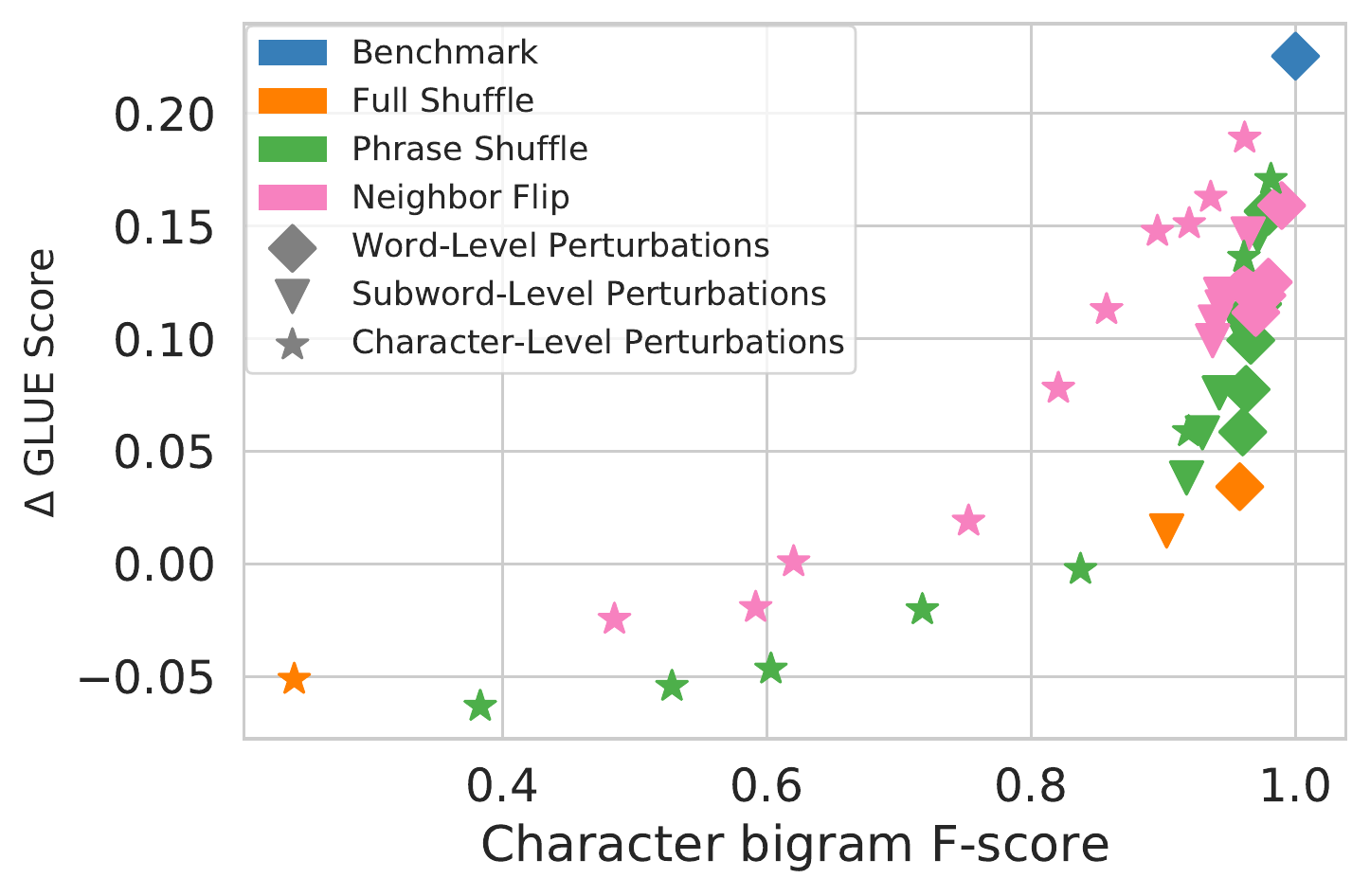}\hfill
    \includegraphics[width=0.31\textwidth]{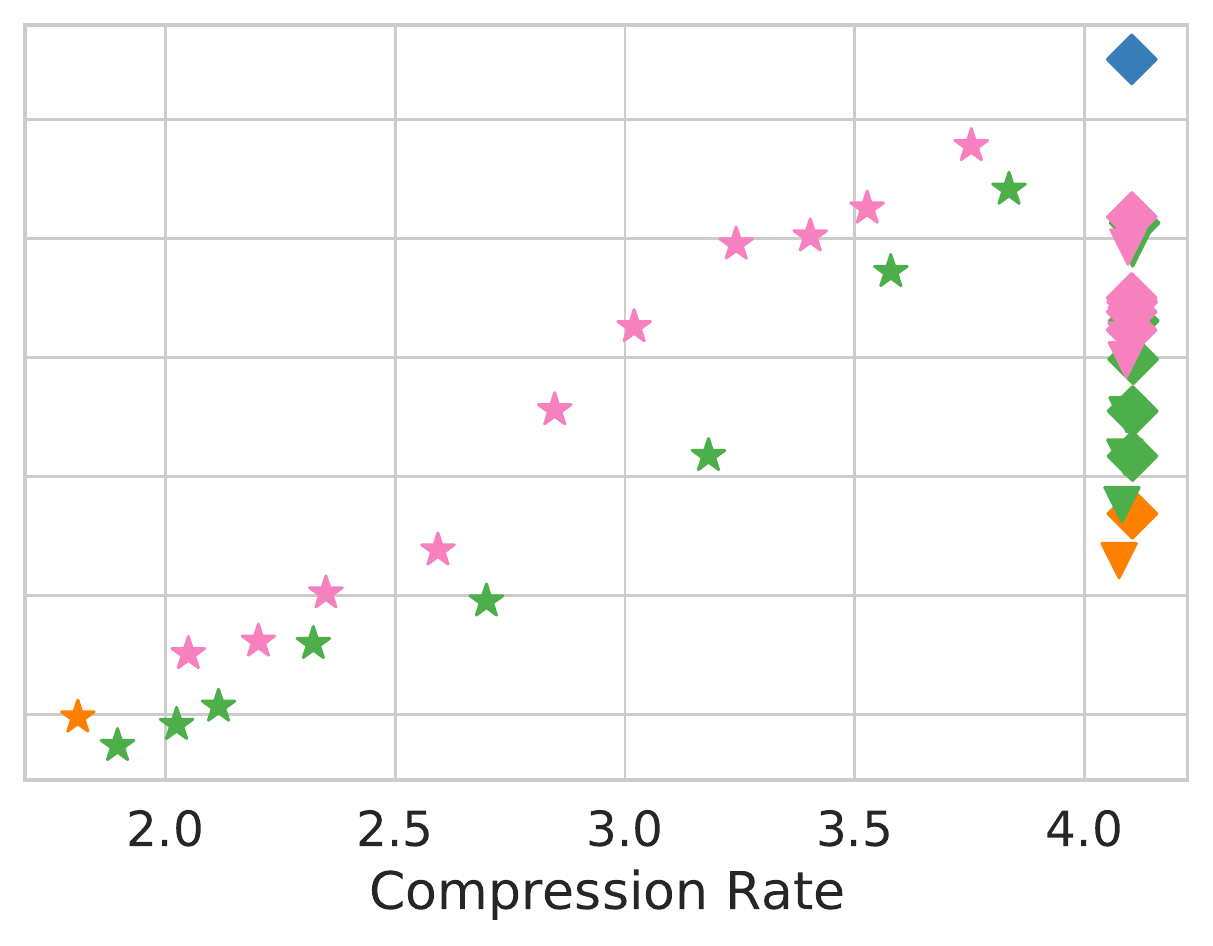}\hfill
    \includegraphics[width=0.31\textwidth]{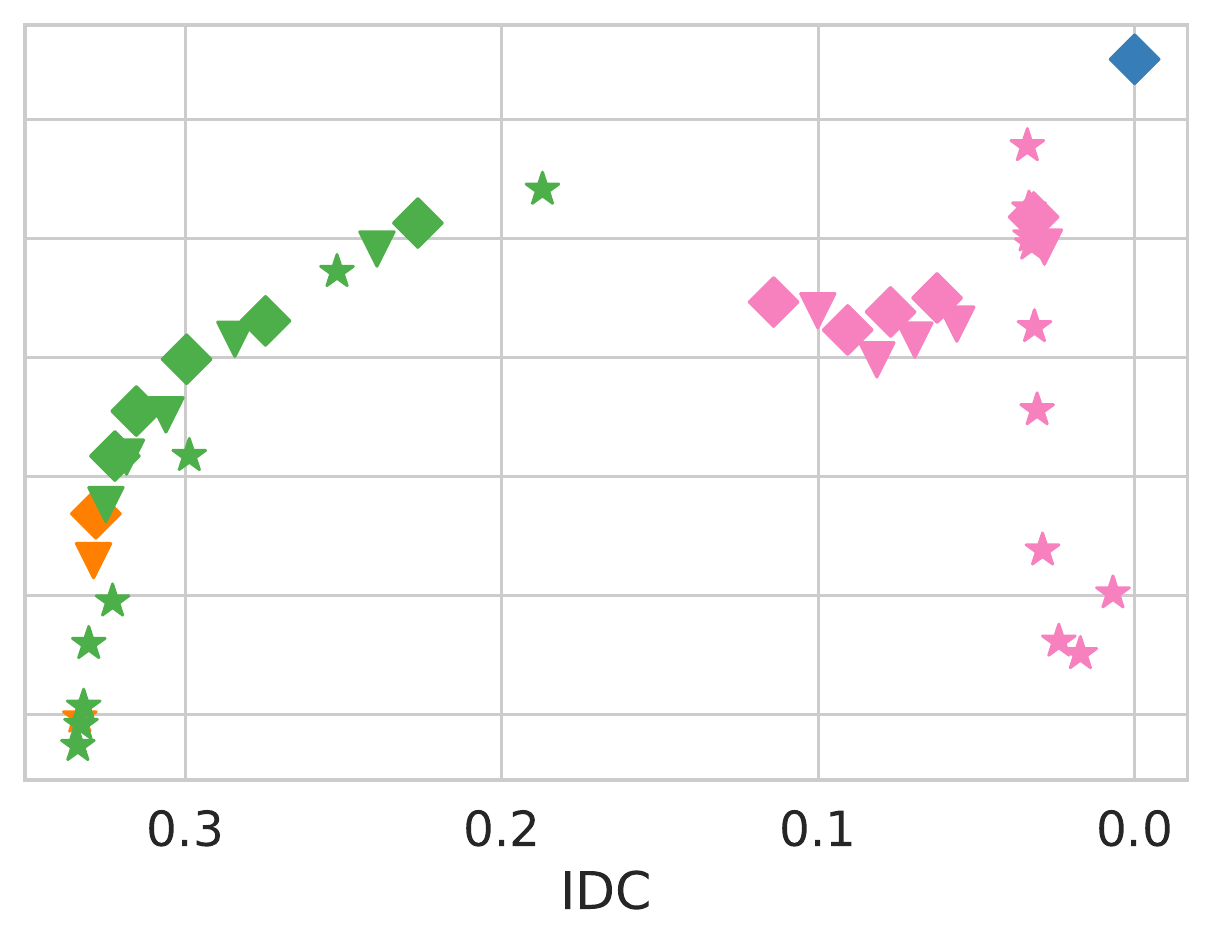}}
   
    \subfigure[NPT Transformer]{
    \includegraphics[width=0.37\textwidth]{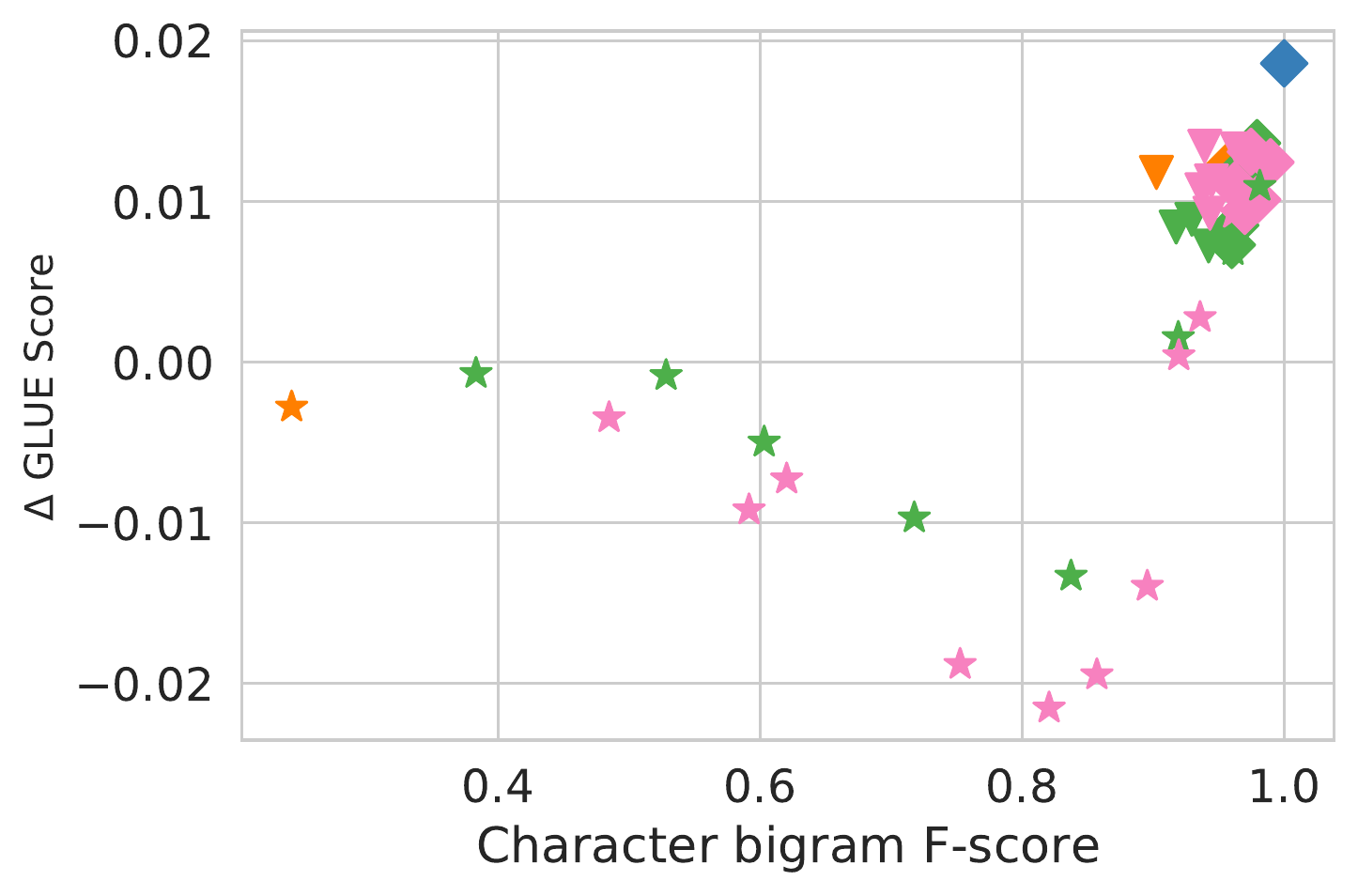}\hfill
    \includegraphics[width=0.31\textwidth]{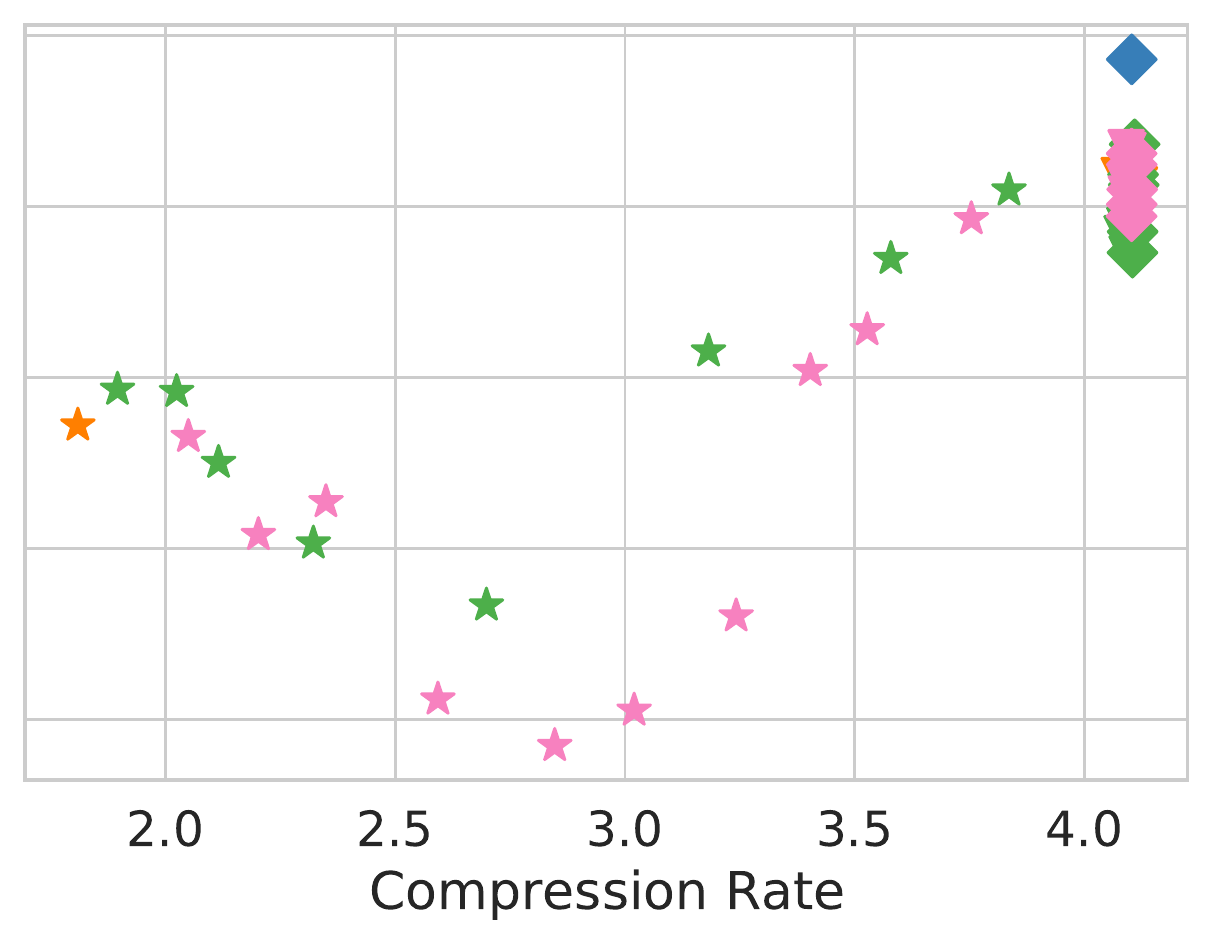}\hfill
    \includegraphics[width=0.31\textwidth]{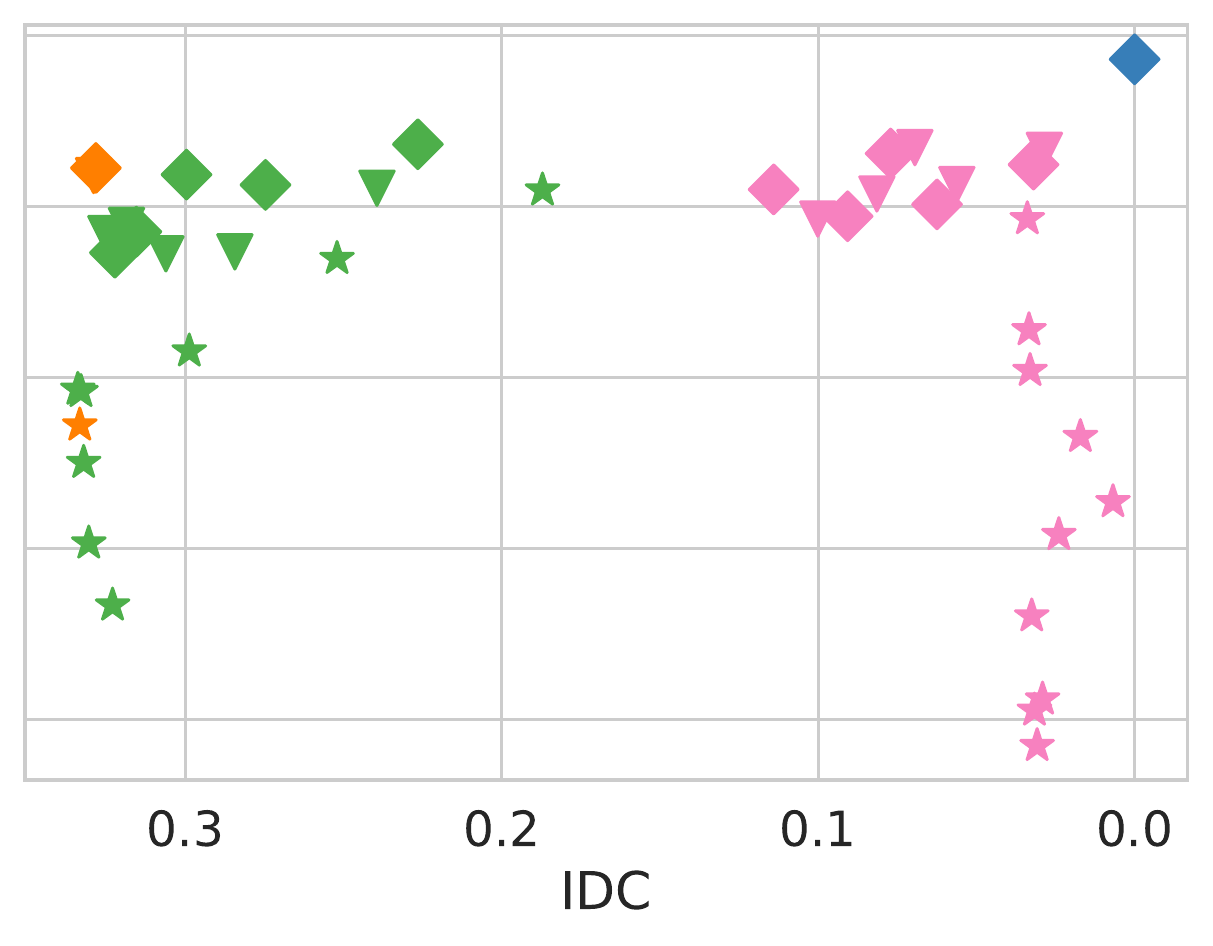}}

    \caption{Difference in GLUE scores between a Transformer and the same Transformer trained and tested with positional embeddings frozen at $0$.
    Results for NPT and PT models are shown.
    }
    \label{fig:no_emb_metrics}
\end{figure*}

% \begin{figure*}[h!t]
%     \centering
%     \subfigure[CHRF NPT]{
%         \includegraphics[width=0.45\textwidth]{res/img/nonpretrained_noemb_diff_local_binary_metric_normalized.pdf}
%         }
%     \subfigure[IDC NPT]{
%         \includegraphics[width=0.45\textwidth]{res/img/nonpretrained_noemb_diff_global_metric_normalized.pdf}
%     }
    
%      \subfigure[DND PT]{
%         \includegraphics[width=0.45\textwidth]{res/img/pretrained_noemb_diff_local_binary_metric_normalized.pdf}
%         }
%     \subfigure[IDC PT]{
%         \includegraphics[width=0.45\textwidth]{res/img/pretrained_noemb_diff_global_metric_normalized.pdf}
%     }
   
%     \caption{Difference in GLUE scores between a Transformer and the same Transformer trained and tested without positional embeddings. 
%     Results for NPT and PT models are shown.
%     }
%     \label{fig:no_emb_metrics}
% \end{figure*}

Towards studying this, we conduct an ablation study on the impact of positional embeddings with NPT and PT Transformers.
We freeze the weights of the positional embeddings to $0$, making them have no contribution to the overall output of the model.
As we are interested in the marginal utility of positional embeddings with relation to NPT Transformers, we report the difference in performance between the model that has access to those embeddings and the model that does not ($\Delta$ GLUE Score). 
Without positional embeddings, a model has no information on the relative position of inputs and is forced to use only the bag-of-word information.
In \RefFigure{fig:no_emb_metrics}, we can see that the performance of the NPT Transformer without positional embedding varies about $\pm2\%$, consistent across all levels of perturbations, while the PT model performance is strongly improved by the presence of the positional embeddings. 
This suggests that NPT Transformers barely make any use of the positional embeddings on those tasks\footnote{Further analysis is presented in Appendix~\ref{app:npt_transformer}}.

\subsection{Character-Level Experimentation}

As the results presented from experiments so far use subword tokenization, it is possible that the local perturbations being directly correlated with performance decay could be caused by the perturbation to the vocabulary.
% We indeed see a large correlation between the compression rate and the performance decay.
%Indicating that higher DND metric correlates with greater destruction to the vocabulary of a model. 
To control for vocabulary destruction as a possible explanation for the observed phenomenon, we train character-level BiLSTMs, ConvNets and finetune a PT CharBERT model on all tasks to evaluate whether the correlations between metrics and performance hold without multi-character vocabulary.
Results shown in \RefFigure{fig:models-metric-correlation} demonstrate that even when using a single-character vocabulary, the correlations between performance for ConvNets, BiLSTMs, and PT Transformers remains roughly static.
This implies that the destruction of the specific tokens used by the model is not the main driver for the degradation in performance leaving perturbation to the local structure as the most likely explanation.

\section{Discussion}

\paragraph{Significance of Results}
While our results at the extremes may be trivial, such that completely shuffling the order of characters of a text removes all the structure necessary for understanding, and that destroying the local structure to an extreme also prohibits models from building a useful representation of the text, it is not trivial that performance correlates to this degree to local structure across the whole spectrum of perturbations.
In \RefFigure{fig:pretrained_transformer_metrics}, fully shuffling the subwords of a text and randomly flipping characters with their neighboring character 10\% of the time obtains roughly the same GLUE score and chrF-$2$ metric despite much different perturbations being applied and much different IDC and compression rate.
The removal of any amount of local structure correlating directly to an equivalent drop in performance, with little concern for the granularity or mechanics of that removal of local structure, allows us to make interesting conclusions on the kind of structure that is used by neural models to build understanding.

\paragraph{Adversarial Attacks}
By better understanding the specific mechanics that can induce failure in neural language models, it is possible to develop models that are more resistant to adversarial attacks.
As current models performances can be directly related to the preservation of character $2$-grams in all studied variations, this study demonstrates a very likely vector of adversarial attacks that may be important to explore further.
\citet{gao2018black} use the Levensthein distance to measure and limit perturbations of black box adversarial attacks, similar research relying on chrF-$2$ instead may be interesting.

% \paragraph{Neural Architecture}
% Our results may be used to guide architectural research in NLP.
% The relative importance of local structure would motivate research into models that treat local structure separately from other types of structure.
% Hierarchical Transformers~\citep{hiearchical-long-classification, hiearchical-multi-document, swin-transformers}, and other hierarchical architectures, seem especially apt at representing this concept, by focusing on creating local meaning first. 

\paragraph{Tokenization}
Our results on the importance of local structure could bear some implications for tokenization.
Recent research trends~\citep{xu27vocabulary,clark2021canine} look at alternatives and improvements to BPE. %attempt at dethroning BPE as the default tokenization scheme.
The current research appears to be pushing towards smaller vocabulary at finer granularity, even exploring simple byte-level representations~\citep{xue2021byt5,tay2021charformer}.

We find that local clumps of characters contain the most essential structural information required to solve several NLU problems.
As a large part of the complexity of NLU seems to be contained within the meaning of the specific order of clumps of characters, by having more of that local structure fixed through tokenization, it is possible to inject additional useful inductive biases into the model.
%it is not clear -- the word problem: 
%It may be that a word-level or even phrase-level tokenization scheme may remove a larger amount of complexity from the problem, by providing a direct representation for specific clumps of characters, and yield better performances.
The perturbation analysis discussed in our work could be used for better construction of vocabulary with improved heuristics. 
%\citet{xu27vocabulary} formulates vocabulary learning as an optimal transport problem towards constructing vocabularies that balances both the corpus entropy and model learning. 

% \paragraph{Learning Positional Embeddings is Data Hungry}

% Our experiments indicate that learning useful positional embeddings may require huge amounts of data and are mostly unused by Transformers that were not pretrained with an extensive corpora.
% This suggests that for problems where the input order is important, and there is limited training data, models with stronger inductive biases such as ConvNets and LSTMs may be a better choice than Transformers that were not pretrained~\citep{tran-etal-2018-importance}. 
% We leave experimenting with models from specific breakpoints to study the evolution of the utility of positional embeddings to be explored as future work.

\section{Conclusion}

% \paragraph{Local, Global, and Bag-of-Words}
Our results on the relative importance of local structure in relation to global structure hint at the possibility that much of the tested NLU tasks can be solved with a bag-of-words formulation.
Intuitively, local structure mainly relates to building meaningful words from the characters of a text whereas the global structure relates to the general order and word-level syntax being maintained.
From our experiments, we observe that as long as the local structure is roughly maintained, a majority of NLU tasks can be solved without requiring the global structure. 
This correlates with similar findings by~\citet{o2021context}.
In essence, the structure required to build words seems to be necessary, but much of NLU can be solved with the information of which words (or subwords) are present in the text, without regard to their relative positions.
% This adds further credibility to similar research that attempts to understand the success of Transformers in NLP through hypothesizing that the global attention makes the architectures particularly apt at reflecting over a set of items, like a bag-of-words.

In this work, we have provided empirical results demonstrating that, for deep learning models in English NLU, perturbations to the local structure, as measured by the chrF-$2$ metric, is highly correlated to downstream model performance which implies that much of the information obtained from the structure of text comes from the local structure.
Perturbations to the global structure, as measured by IDC, seems to only have a limited correlation to performance, implying that models don't generally rely on it to build understanding. 
Reflecting on our results, we observe that perturbations on a local level explains the (in)sensitivity of neural language models to perturbations at different granularities on a variety of NLU tasks. 
This paper hopefully provides useful intuitions on the importance of different types of structures in text for researchers looking into tokenization, neural architectures and adversarial attacks.
Although the paper primarily focuses on the effects of perturbations on English texts, extending the study to neural models on other languages will be beneficial.
% Especially, studying whether perturbations have a similar effect on other languages could help in deepening our understanding of cross-language tasks, like machine translation.

\section*{Acknowledgements}
We thank Saujas Vaduguru for the useful comments and discussions on early drafts. 
This research was supported by Apogée Canada, Canada First Research Excellence Fund program and École Polytechnique Startup Fund PIED. 
SC is supported by a Canada CIFAR AI Chair and an NSERC Discovery Grant.

\newpage

\nocite{glue1, glue2, glue3, glue4, glue5, glue6, glue7, glue8, glue9, glue10, glue11}
\nocite{HuggingFace}
\nocite{pytorch}

\bibliographystyle{acl_natbib}
\bibliography{anthology,acl2021}

\clearpage
\appendix

\onecolumn
\section{Experiment Details}\label{app:experiments}

\paragraph{Model Hyperparameters}\label{app:hyperparameters}
The results in the paper are averaged over 5 random seeds.
We train 5 individual model on all tasks and apply a different random seed to the perturbations to each trained model once.
Early stopping was performed after 2 full epochs not resulting in better results on the validation set.
All models had similar model sizes, containing between 100 million and 130 million parameters.
The ConvNet architecture is the one described in \citet{convnet_architecture} and the BiLSTM architecture is the one described in \citet{bilstm_architecture}.
The character embedding ConvNet uses a kernel of size 12 instead of 3, to offset the much longer character sequences.
Both the ConvNet and BiLSTM use the same hidden size, dropout and word embedding size as the RoBERTa-Base model.
Pretrained models used a learning rate of 2e-5, a batch size of 32, a maximum of 5 epochs and a weight decay of 0.1.
Non-pretrained models used a learning rate of 1e-4, a batch size of 128, a maximum of 50 epochs and a weight decay of 1e-6.
All experiments used a warmup ratio of 0.06, as described in \citet{RoBERTa}.
Experiments using characters as input used a maximum sequence length of 2048 inputs.
All other experiments used a maximum sequence length of 512.
The Winograd Schema Challenge (WNLI) task was omitted from all experiments as it contains well known issues and is often omitted~\citep{RoBERTa, BERT, GPT-1}. 
The validation set, instead of the test set, is used as the test set is kept private for the GLUE benchmark.

\paragraph{Perturbations}\label{app:perturbations}
Subword-level perturbations were all done with the RoBERTa-Base tokenization.
On all level of granularity, we perform one experiment with in the full shuffling setting.
On the word and subword-level perturbations we perform phrase-shuffling with $\rho$ values of: $[0.8$, $0.65$, $0.5$, $0.35$, $0.2]$ and neighbour-flip shuffling with $\rho$ values of: $[0.8$, $0.6$, $0.5$, $0.4$, $0.2]$.
On the character-level perturbations we perform phrase-shuffling with $\rho$ values of: $[0.975$, $0.95$, $0.9$, $0.8$, $0.65$, $0.5$, $0.4$, $0.3$, $0.2$, $0.15$, $0.1$, $0.075$, $0.05]$ and neighbour-flip shuffling with $\rho$ values of: $[0.8$, $0.65$, $0.5$, $0.4$, $0.3$, $0.2$, $0.1$, $0.075$, $0.05$, $0.035$, $0.025$, $0.01]$.
A total of 11 word-level experiments, 11 subword-level experiments, 27 character-level experiments and the unperturbed benchmark are evaluated for a grand total of 50 different perturbation settings.

\section{Pseudocode for Metric and Perturbations}
\label{app:pseudocode_perturbations}
\newcommand{\myalgorithma}{%
      \begin{algorithm}[h]
      \small
      \SetAlgoLined
      \SetKw{Kw}{\KwInput}
      \SetKwInOut{Input}{input}
      \SetKwInOut{Output}{output}
      \SetKwProg{Fn}{Function}{:}{\KwRet}
      
      \SetKwFunction{Func}{IDC}
      \Fn{\Func{$X_p$}}{
       $X_p^{len} \leftarrow X_p.\texttt{length()}$\;
      IDC\_list $\leftarrow$ \texttt{list()}
      
      \For{ $i \leftarrow\ 0$ {\bf and} $i \leq X_p^{len}$}{
      abs\_distortion $\leftarrow$ \texttt{abs(i-$X_p\left[i\right]$)}\;
      IDC\_list.\texttt{append}(abs\_distortion)\;
      }
      IDC\_agg $\leftarrow$ IDC\_list.\texttt{mean}()\;
      
      IDC $\leftarrow \frac{IDC\_agg}{X_p^{len}}$\;
        }
      %\caption{Pseudocode for calculating IDC metric. \texttt{length} and \texttt{mean} are assumed in-built functions that computes the length and mean of a list respectively.}
    \caption{Pseudocode to compute IDC metric.}
    \end{algorithm}
}
\newcommand{\myalgorithmb}{
\begin{algorithm}[h]
\small
\SetAlgoLined
  \SetKw{Kw}{\KwInput}
  \SetKwInOut{Input}{input}
  \SetKwInOut{Output}{output}
  
  \SetKwProg{Fn}{Function}{:}{\KwRet}
  
  \SetKwFunction{Func}{DND}
  \Fn{\Func{$X_p$}}{
  $X_p^{len} \leftarrow$ $X_p.\texttt{length()}$\;
  DND\_list $\leftarrow$ \texttt{list()}
  
  \For{ $i \leftarrow\ 0$ {\bf and} $i \leq X_p^{len} - 1$}{
  \eIf{$X_p[i] - X_p[i+1]==$1}{
  DND\_list.\texttt{append}($0$)\;
  }{
  DND\_list.\texttt{append}($1$)\;
  }
  }
  DND\_agg $\leftarrow$ DND\_list.\texttt{sum}()\;
  
  DND $\leftarrow \frac{DND\_agg}{(X_p^{len} - 1)}$\;
    }
  %\caption{Pseudocode for calculating DND metric. \texttt{length} and \texttt{sum} are assumed in-built functions that computes the length and mean of a list respectively.}
\caption{Pseudocode to compute DND metric.}
\end{algorithm}
}

% \subfigure[]{
%   \includegraphics[width=.5\textwidth]{
\newcommand{\myalgorithmc}{
\begin{algorithm}[h]
  \small
  \SetKw{Kw}{\KwInput}
  \SetKwProg{Fn}{Function}{:}{\KwRet perturbed\_text}
  \SetKwFunction{Func}{PhrasePerturbation}
  \Fn{\Func{$\rho \leftarrow 0.5$, text$\leftarrow$\texttt{list}}}{

  all\_phrases $\leftarrow$ \texttt{list()}\;
  phrase $\leftarrow$ \texttt{list(text[0])}
  
  \For{token {\bf in} $\rm{text}[1:]$}{
  %p$\leftarrow$ \texttt{random()}\;
  p $\sim Unif\left(\left[0,1\right]\right)$\;
  \eIf{$p < \rho$}{
  all\_phrases.\texttt{append}(phrase)\;
  phrase $\leftarrow$ \texttt{list}(token)
  }{
  phrase $\leftarrow \left[
  \rm{phrase,token} \right]$\;
  }
  }
  all\_phrases.append(phrase)\;
  perturbed\_text $\leftarrow$ `'.\texttt{join}(shuffle(all\_phrases))
  }
  \caption{Pseudocode for PhraseShuffle.}
\end{algorithm}
}
% \subfigure[]{
%   \includegraphics[width=.5\textwidth]{

\newcommand{\myalgorithmd}{
\begin{algorithm}[h]
\small
  \SetKw{Kw}{\KwInput}
  \SetKwProg{Fn}{Function}{:}{\KwRet perturbed\_text}
  \SetKwFunction{Func}{NeighborFlip}
  \Fn{\Func{$\rho \leftarrow 0.5$,text$\leftarrow$\texttt{list}}}{

  perturbed\_tokens $\leftarrow$ \texttt{list()}\;
  held\_token $\leftarrow$ \texttt{list(text[0])}
  
  \For{token {\bf in} $\rm{text}[1:]$}{
  %p$\leftarrow$ \texttt{random()}\;
  p $\sim Unif\left(\left[0,1\right]\right)$\;
  \eIf{$p < \rho$}{
  perturbed\_tokens.\texttt{append}(held\_token)\;
  held\_token $\leftarrow$ \texttt{list}(token)
  }{
  perturbed\_tokens $\leftarrow \left[
  \rm{perturbed\_tokens,token} \right]$\;
  }
  }
  perturbed\_tokens.append(held\_token)\;
  perturbed\_text $\leftarrow$ `'.\texttt{join}(perturbed\_tokens)
  }
  \caption{Pseudocode for NeighborFlip.}
\end{algorithm}
}
% }

% \myalgorithma

% \myalgorithmb

\myalgorithmc

\myalgorithmd

% \begin{figure*}[htb]
%   \null\hfill \myalgorithma \hfill \myalgorithmb \hfill\null\par \medskip
%   \null\hfill \myalgorithma \hfill \myalgorithma \hfill\null
%   \caption{Here are some algorithms.}
%   \end{figure*}

\newpage
\section{Other Results}\label{app:all_results}
In this section, we add for all other tested models the results that were presented for the RoBERTa-Base model.
They were not included in the main paper for simple economy of space.

\subsection{PT BART}
The PT BART model has results that are very much inline with the PT RoBERTa model.

\begin{figure}[H]
    \centering
    \includegraphics[width=0.35\textwidth]{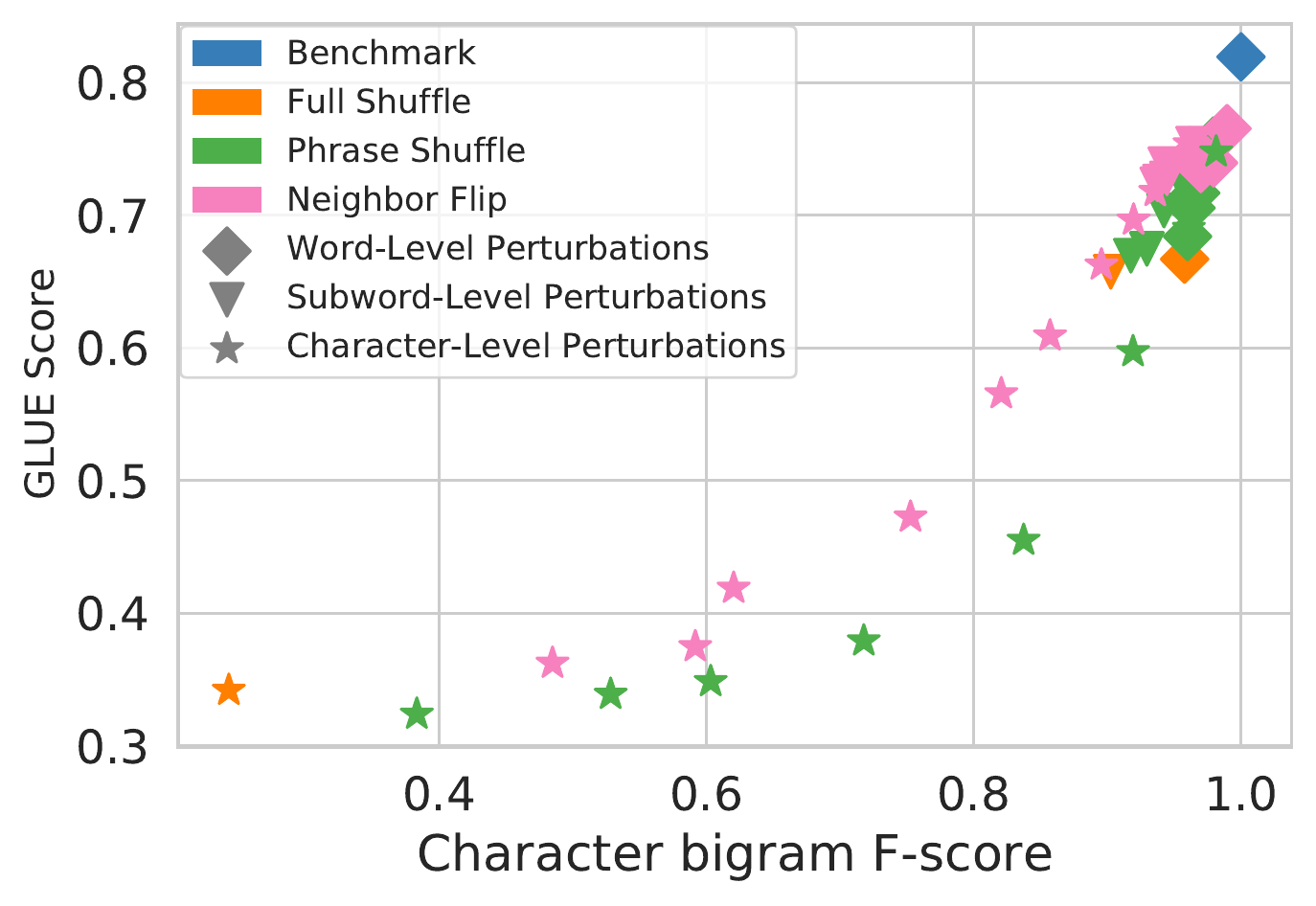}\hfill
    \includegraphics[width=0.31\textwidth]{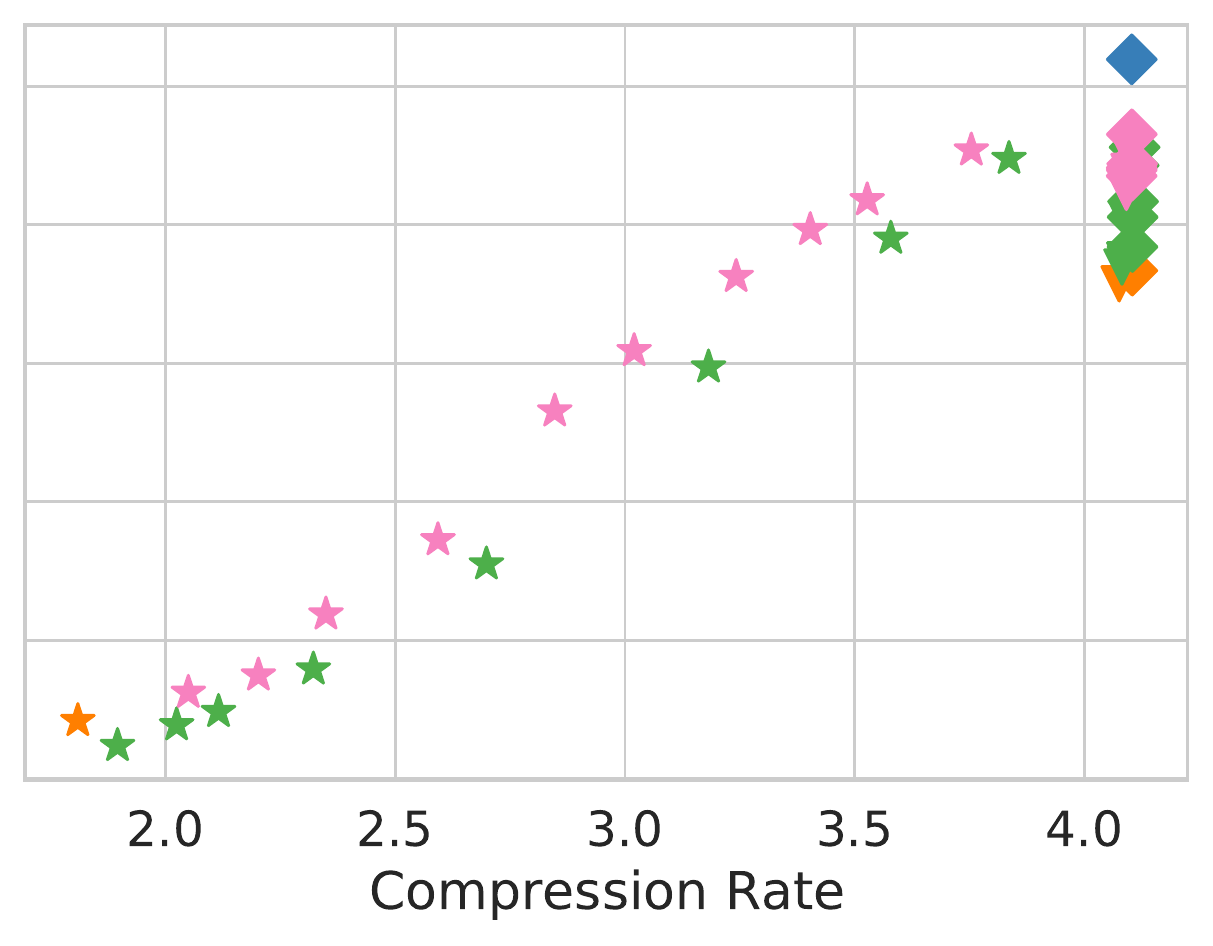}\hfill
    \includegraphics[width=0.31\textwidth]{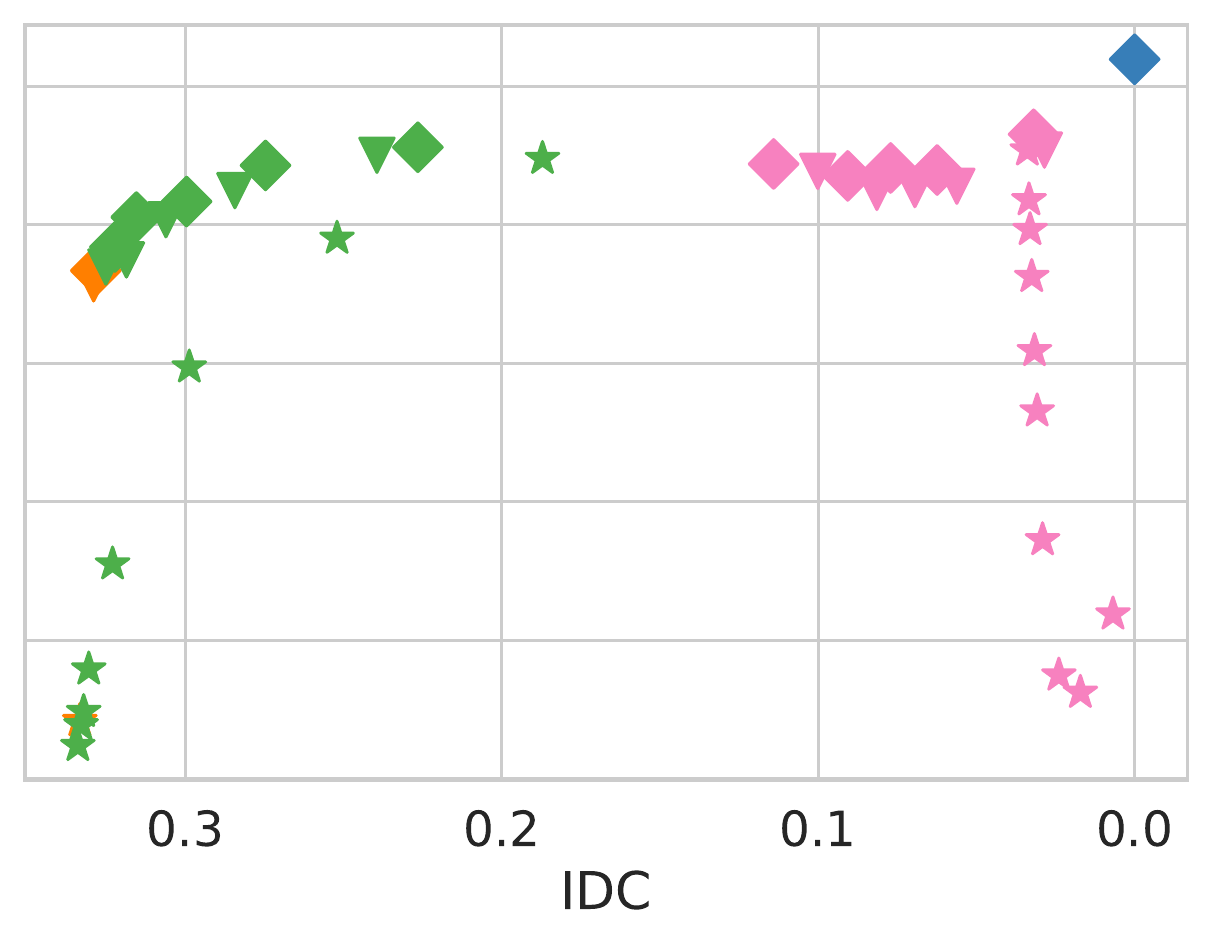}

    \caption{
    Plotted are the relations between the different choices of metrics measuring the amount of perturbation and the performance of PT BART-Base model tested on the perturbed data.
    }
    \label{fig:pretrained_bart_metrics}
\end{figure}

\begin{figure}[H]
    \centering
  
    \includegraphics[width=\columnwidth]{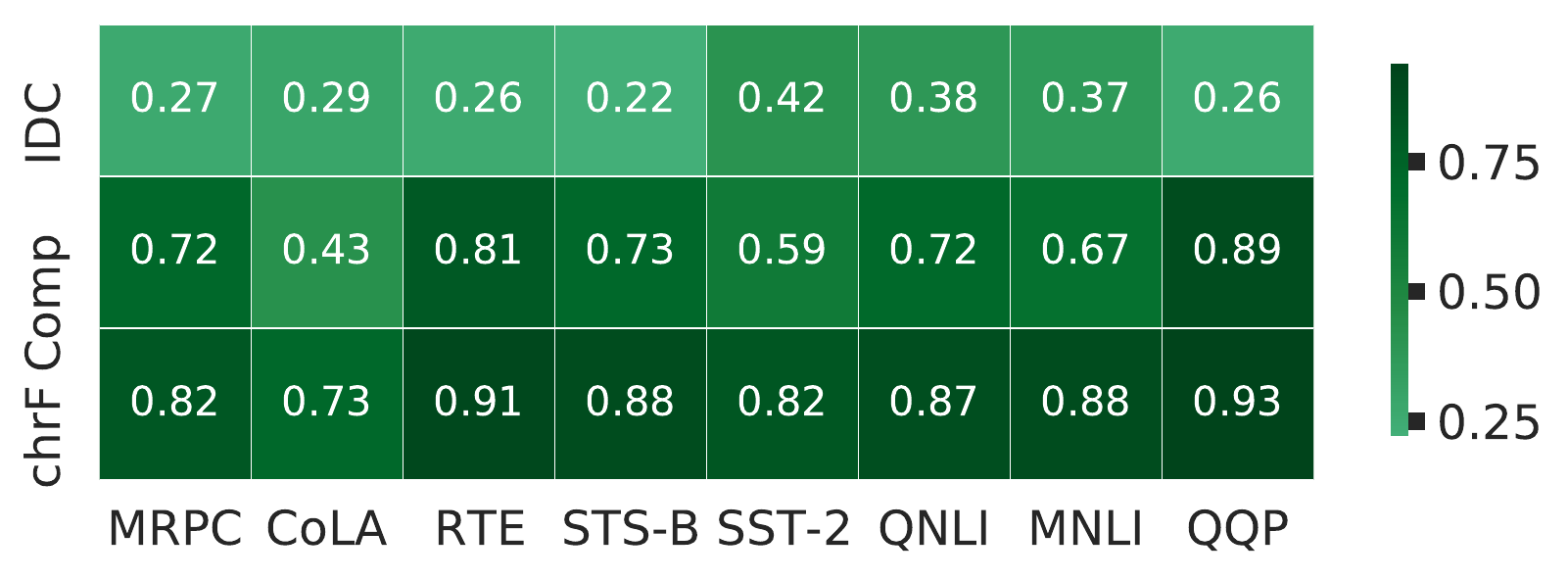}
    \caption{Rank correlation matrix between perturbations measured by different metrics and the performance on the different GLUE tasks of the PT BART model.
    }
    \label{fig:task_to_scores_bart}
\end{figure}

\newpage
\subsection{NPT Transformer}\label{app:npt_transformer}

\begin{figure}[H]
    \centering
    \includegraphics[width=0.35\textwidth]{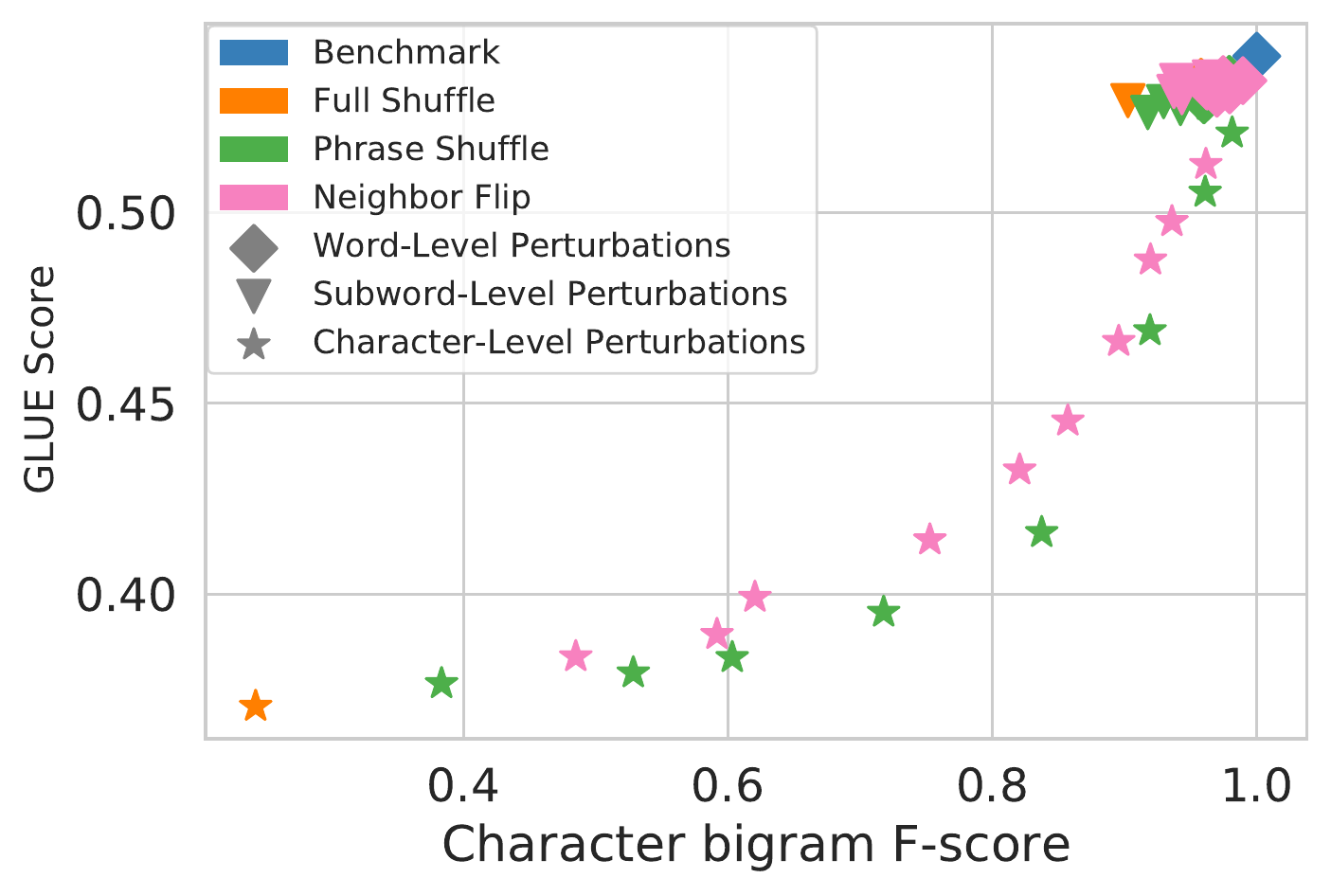}\hfill
    \includegraphics[width=0.31\textwidth]{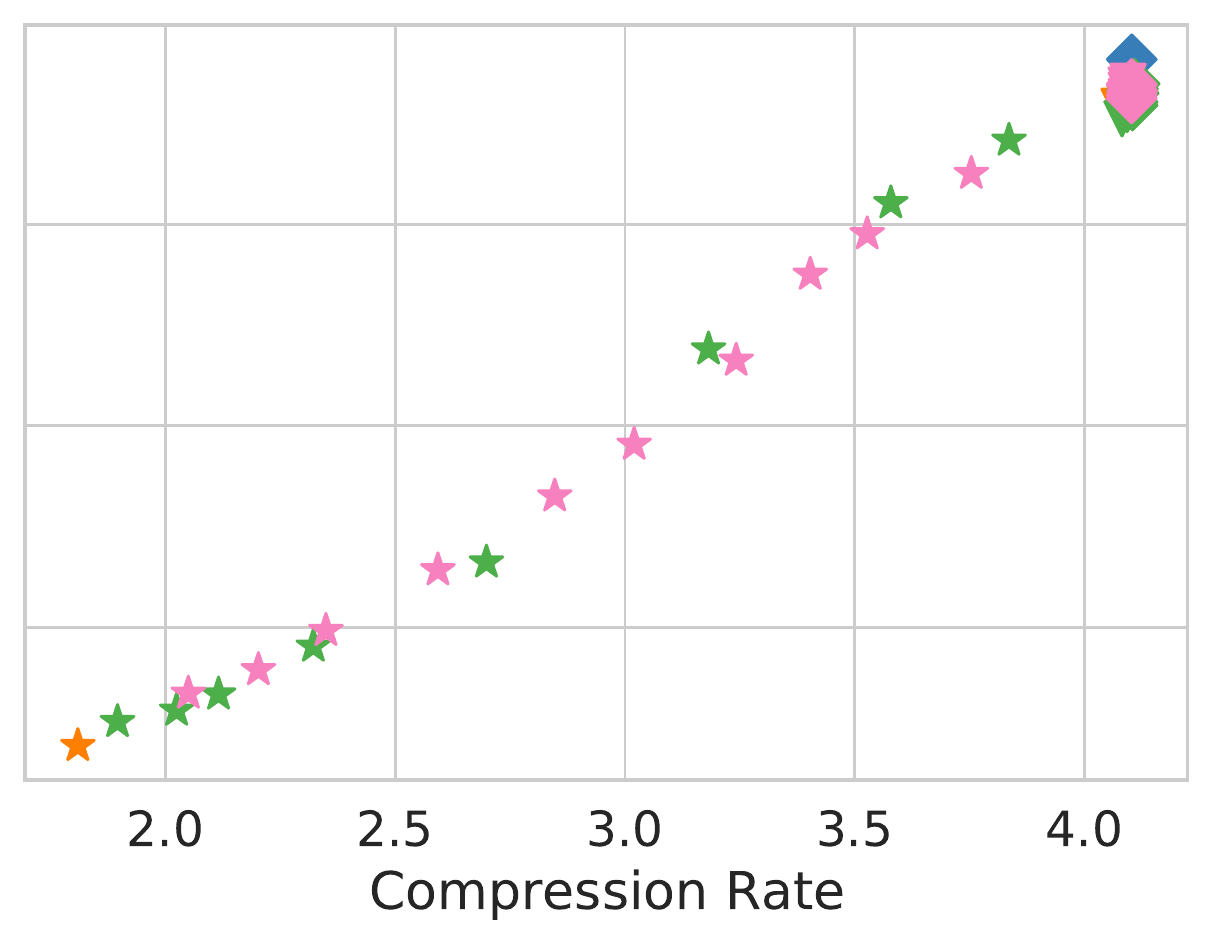}\hfill
    \includegraphics[width=0.31\textwidth]{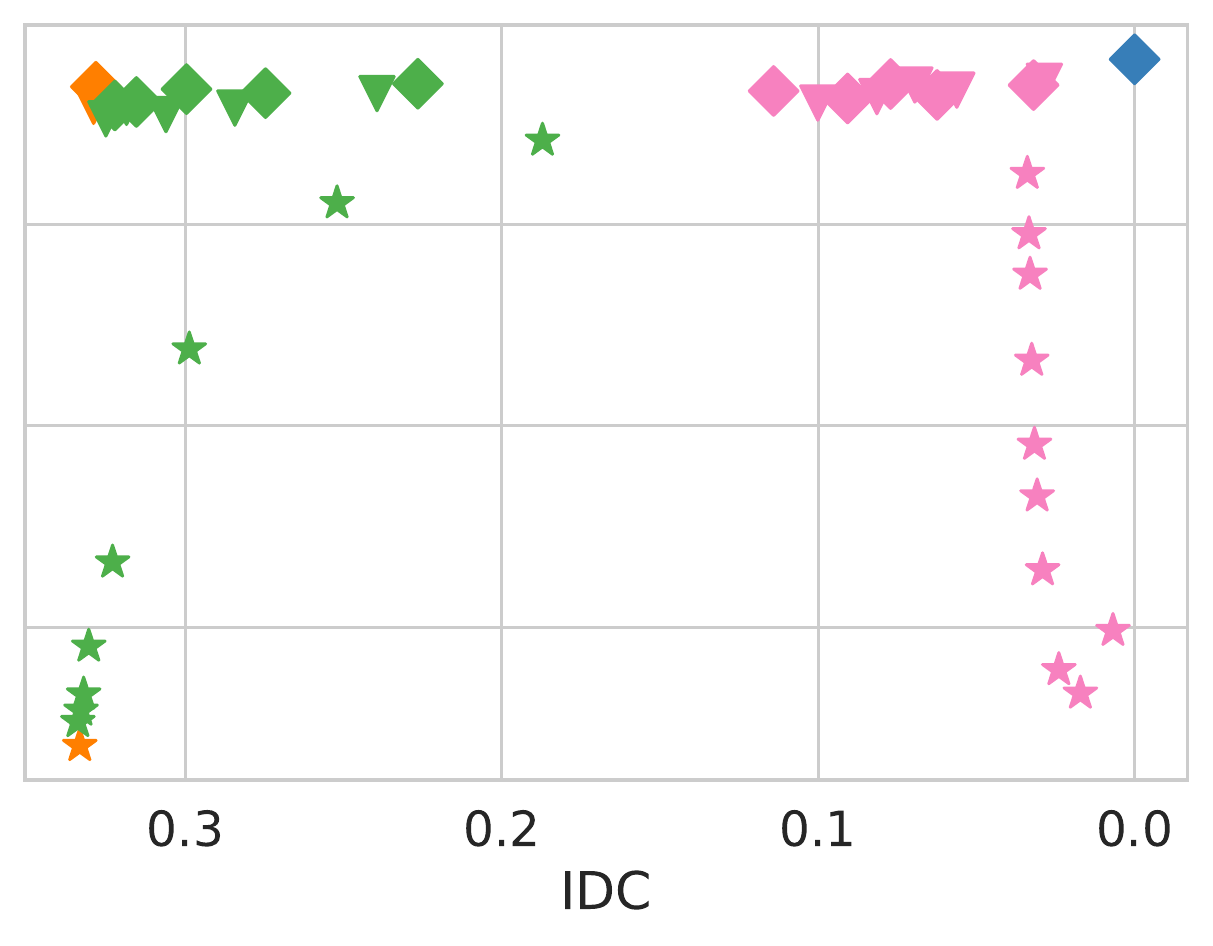}

    \caption{
    Plotted are the relations between the different choices of metrics measuring the amount of perturbation and the performance of NPT Transformer model tested on the perturbed data.
    The model does not seem to consider the position of tokens which explains why word and subword-level perturbation do not seem to affect the performances.
    }
    \label{fig:npt_transformer_metrics}
\end{figure}

\begin{figure}[H]
    \centering
  
    \includegraphics[width=\columnwidth]{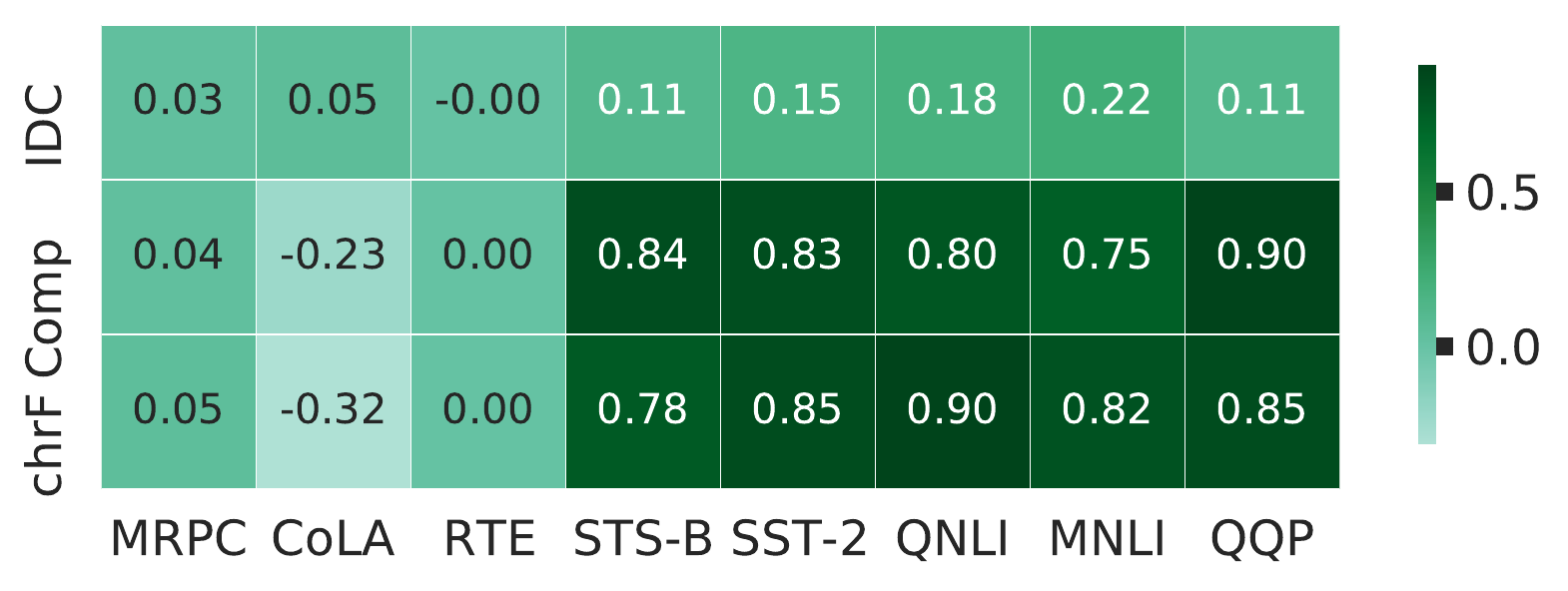}
    \caption{Rank correlation matrix between perturbations measured by different metrics and the performance on the different GLUE tasks of the NPT Transformer model.
    The model obtains a static chance score on the RTE task and extremely low scores on the MRPC and CoLA tasks which explains the strange correlations. 
    Those three tasks have seen the greatest improvement on the GLUE benchmark from the introduction of PT models.
    Those are also the three smallest tasks in the GLUE benchmark lending credence to the idea that positional embeddings are data hungry.
    }
    \label{fig:task_to_scores_npt_transformer}
\end{figure}

\begin{figure}[H]
    \centering
    \subfigure[NPT Transformer MRPC]{
    \includegraphics[width=0.37\textwidth]{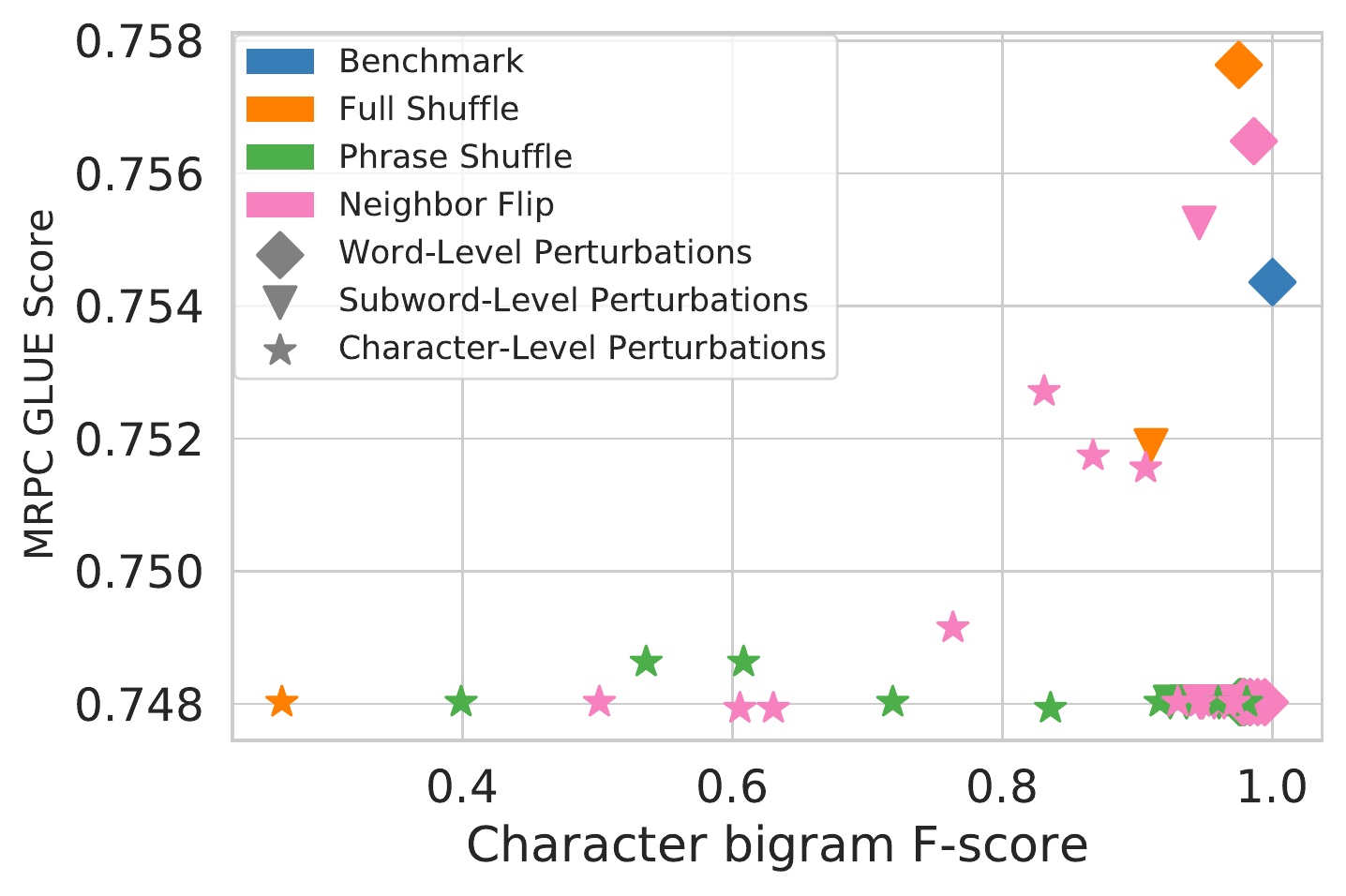}\hfill
    \includegraphics[width=0.32\textwidth]{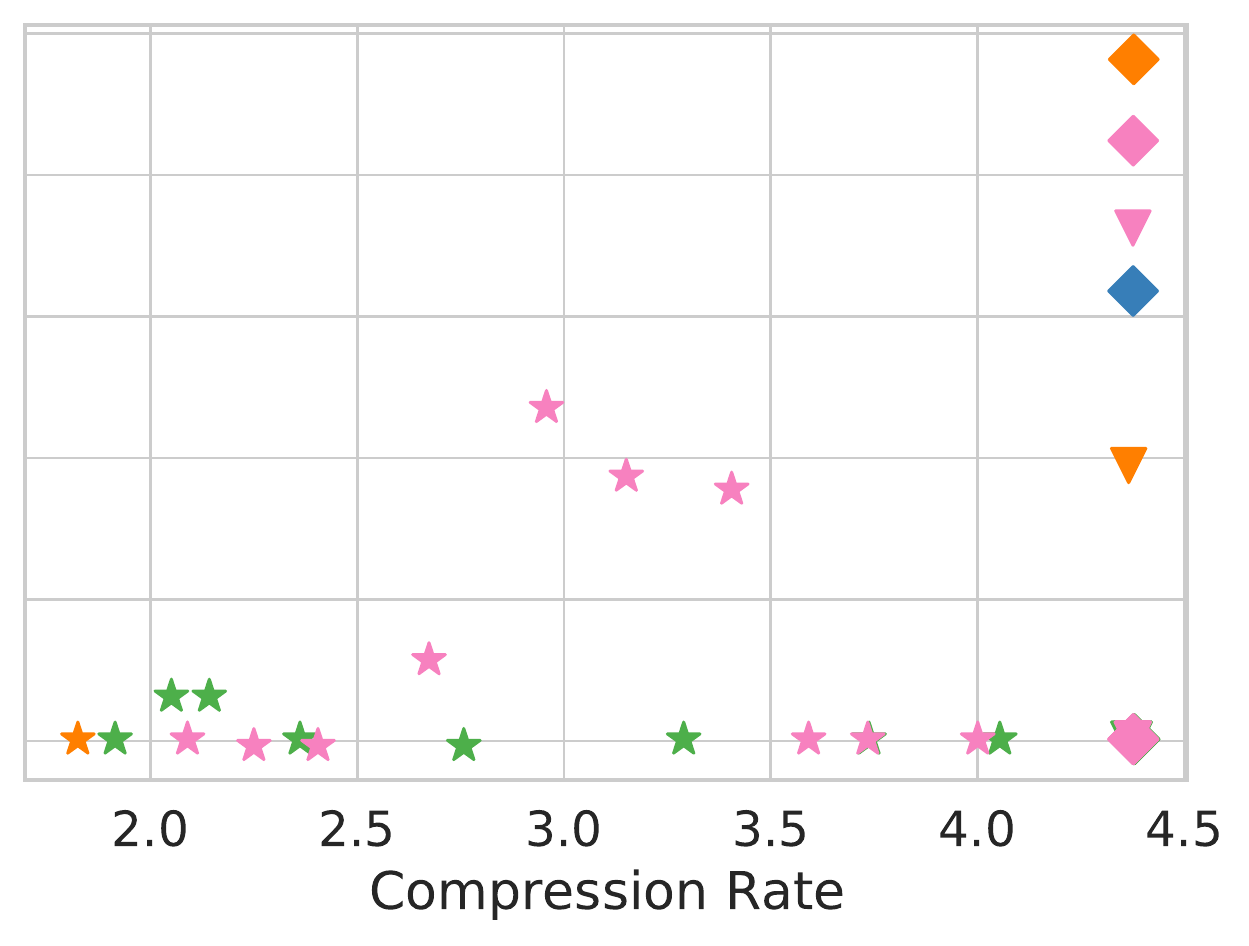}\hfill
    \includegraphics[width=0.31\textwidth]{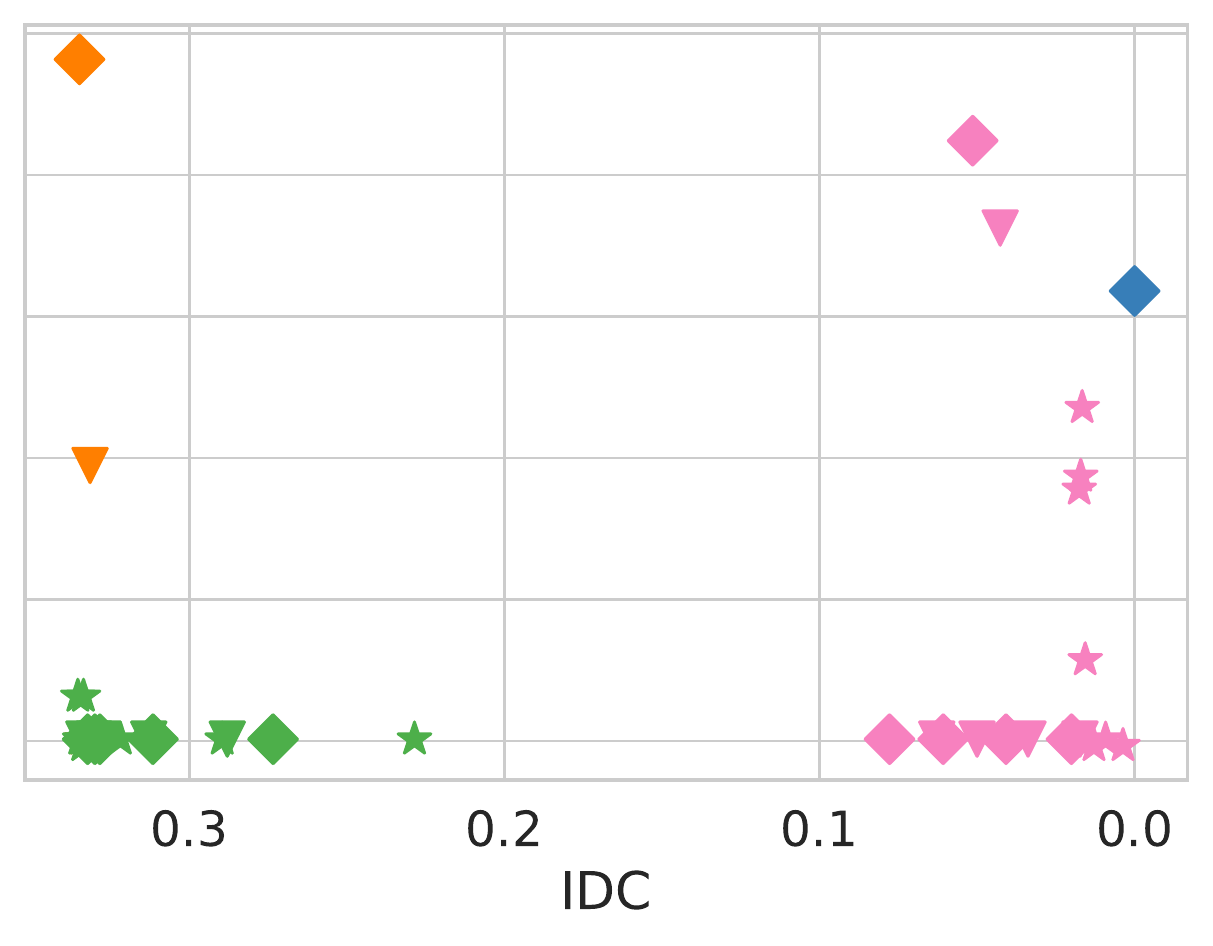}
}

    \subfigure[NPT Transformer CoLA]{
    \includegraphics[width=0.37\textwidth]{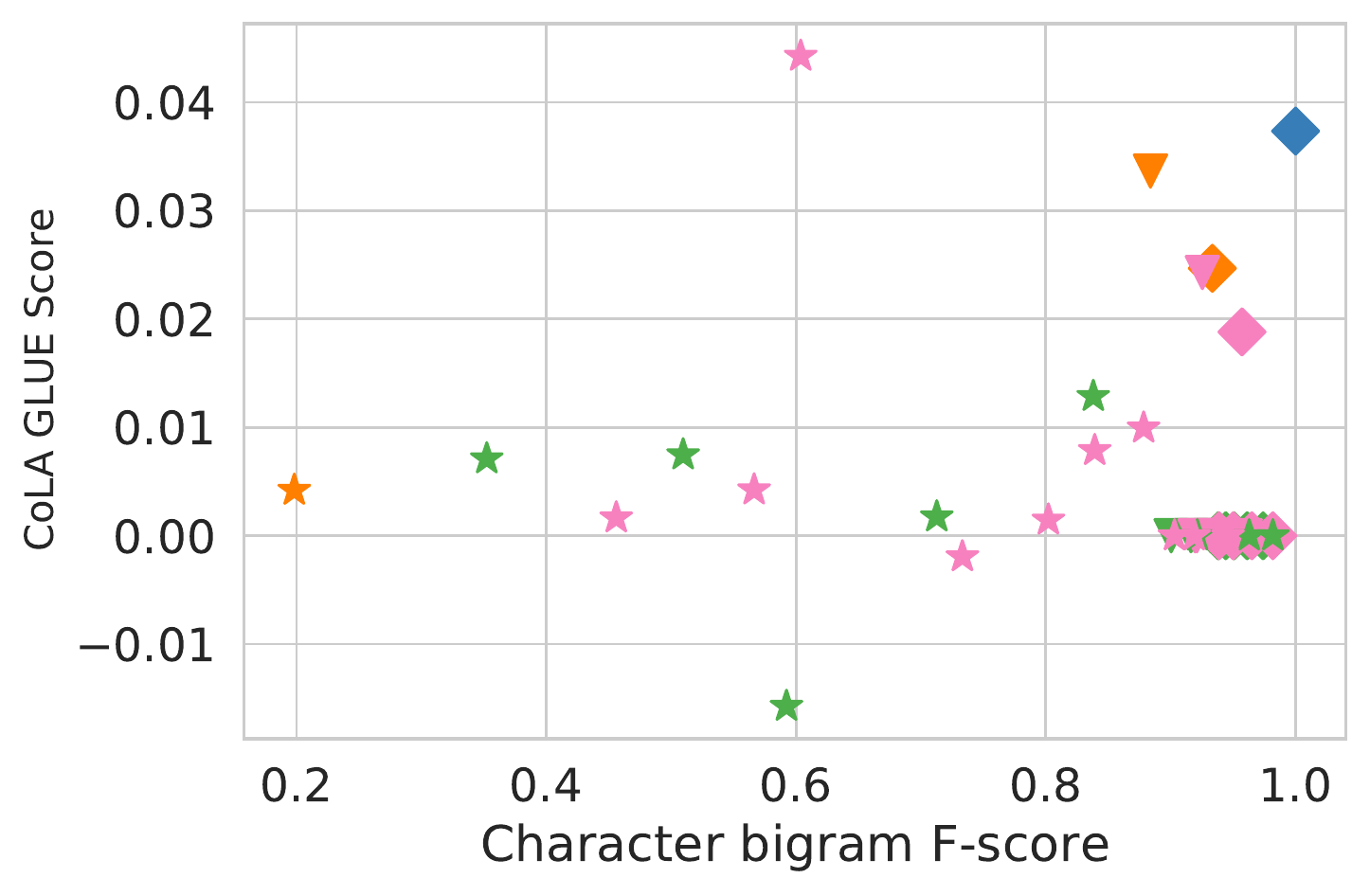}\hfill
    \includegraphics[width=0.31\textwidth]{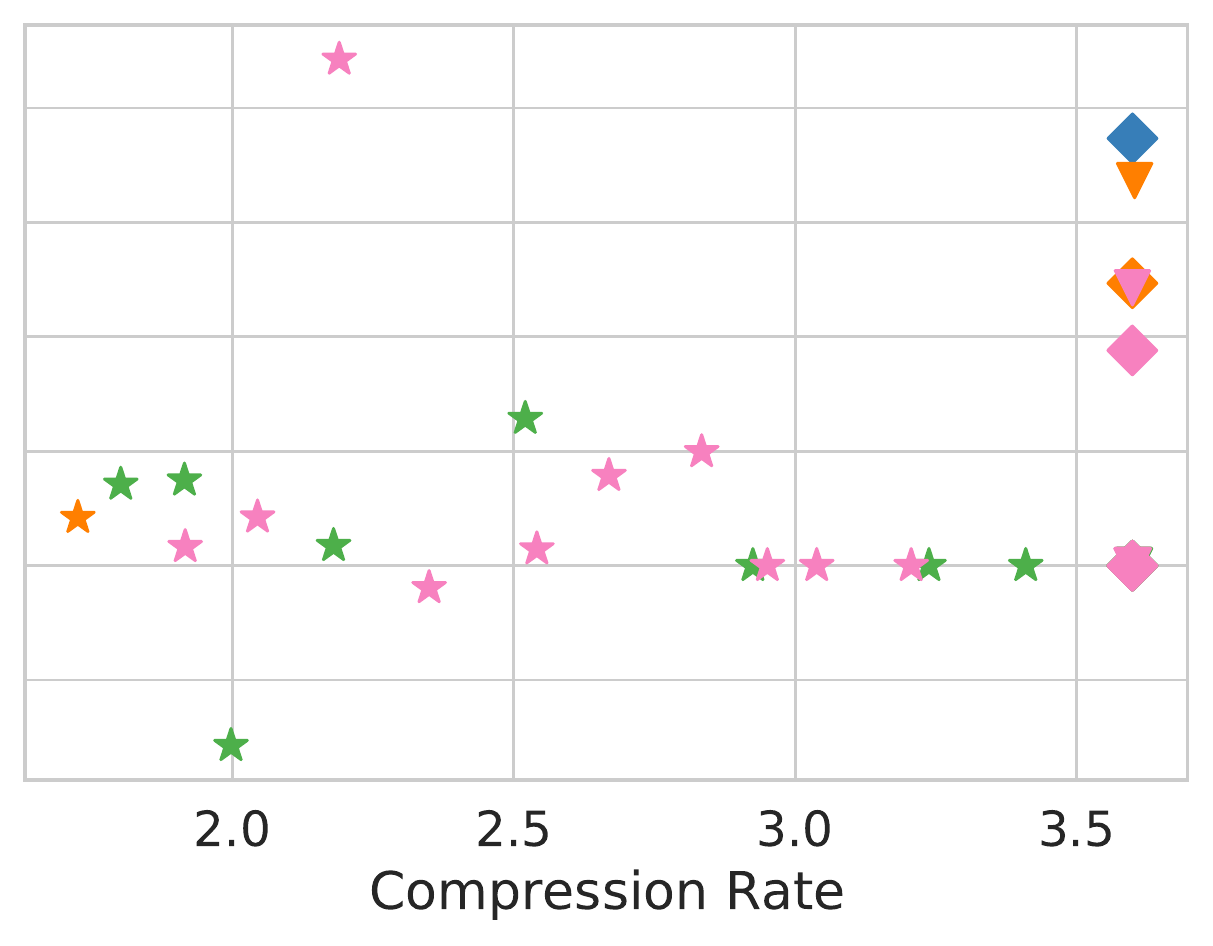}\hfill
    \includegraphics[width=0.31\textwidth]{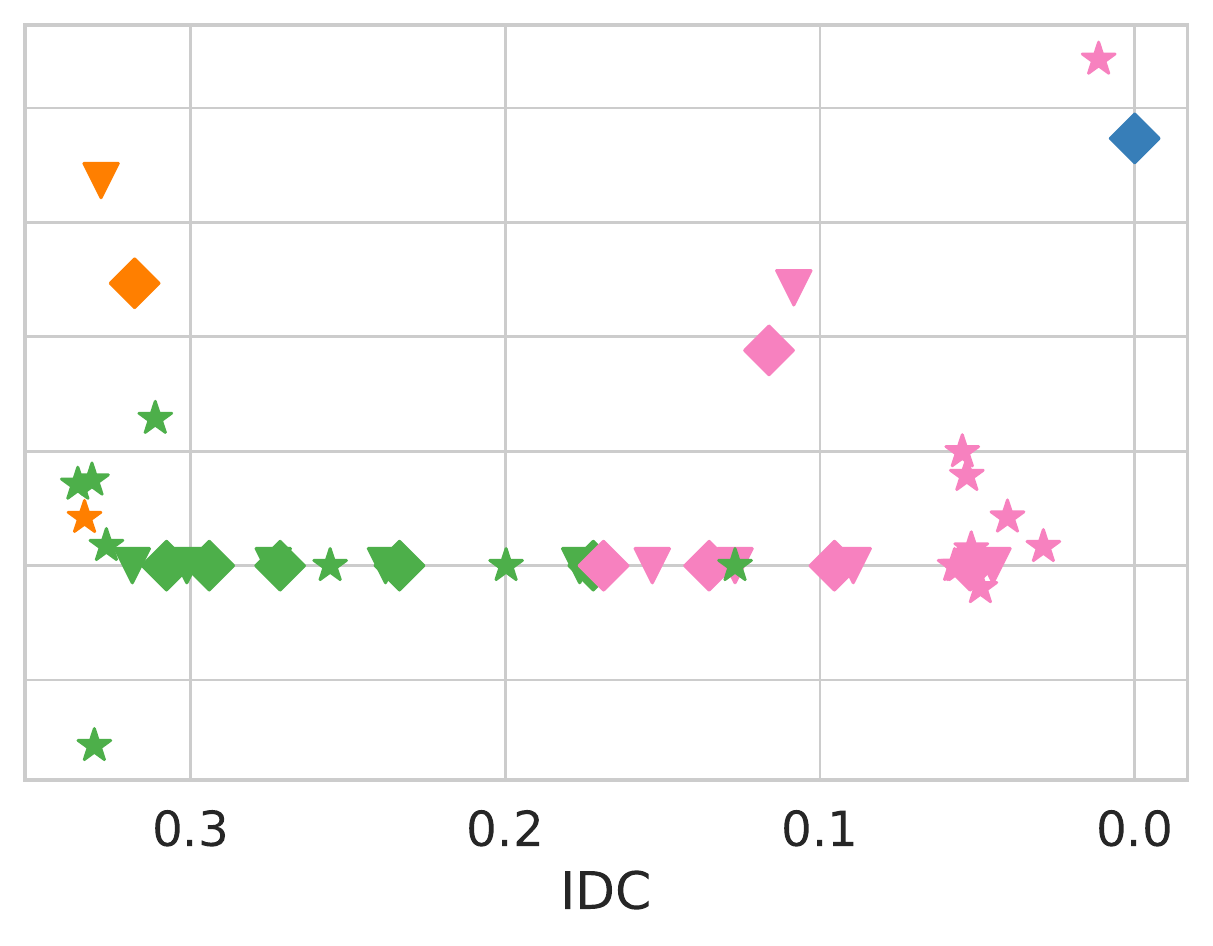}
}

    \subfigure[NPT Transformer RTE]{
    \includegraphics[width=0.37\textwidth]{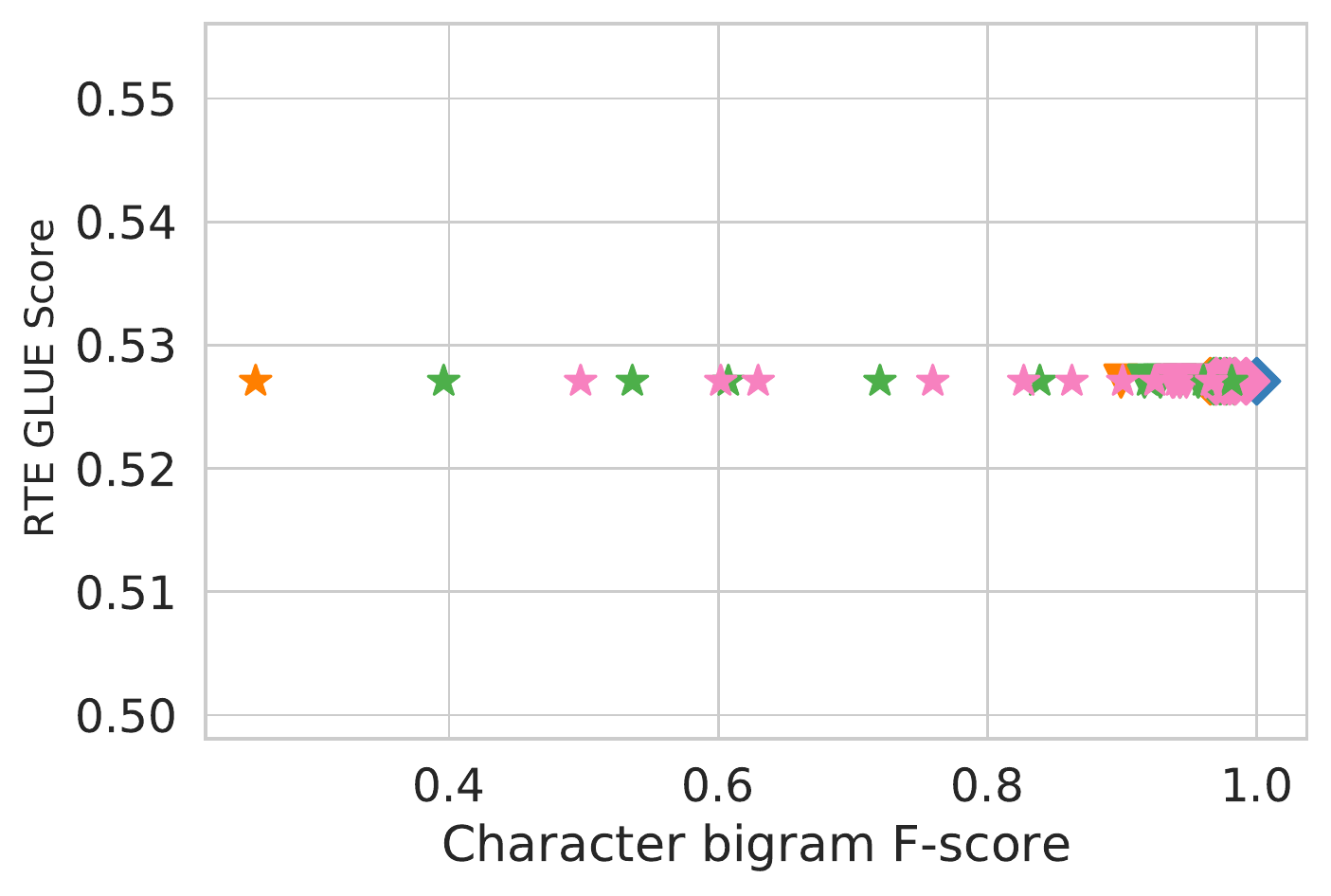}\hfill
    \includegraphics[width=0.31\textwidth]{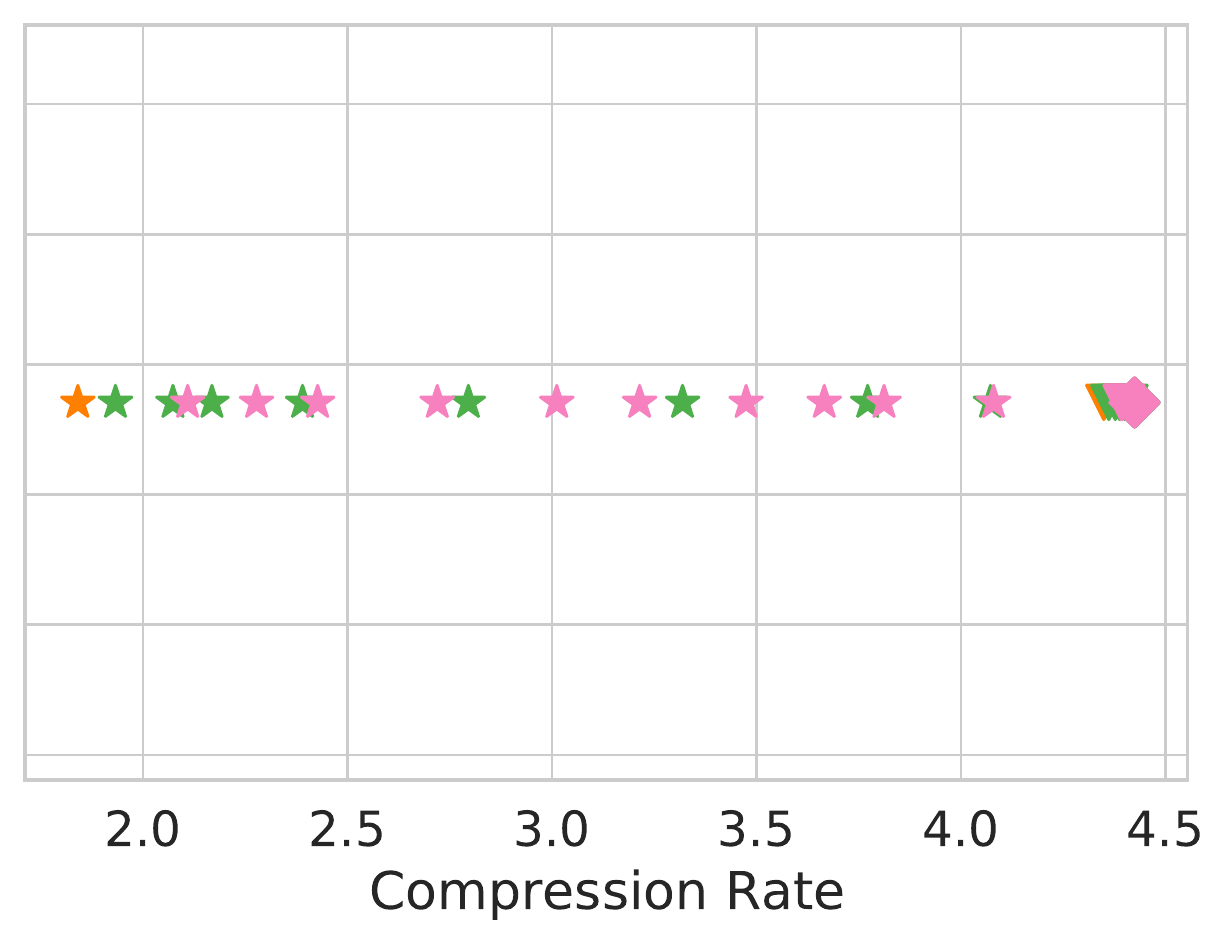}\hfill
    \includegraphics[width=0.31\textwidth]{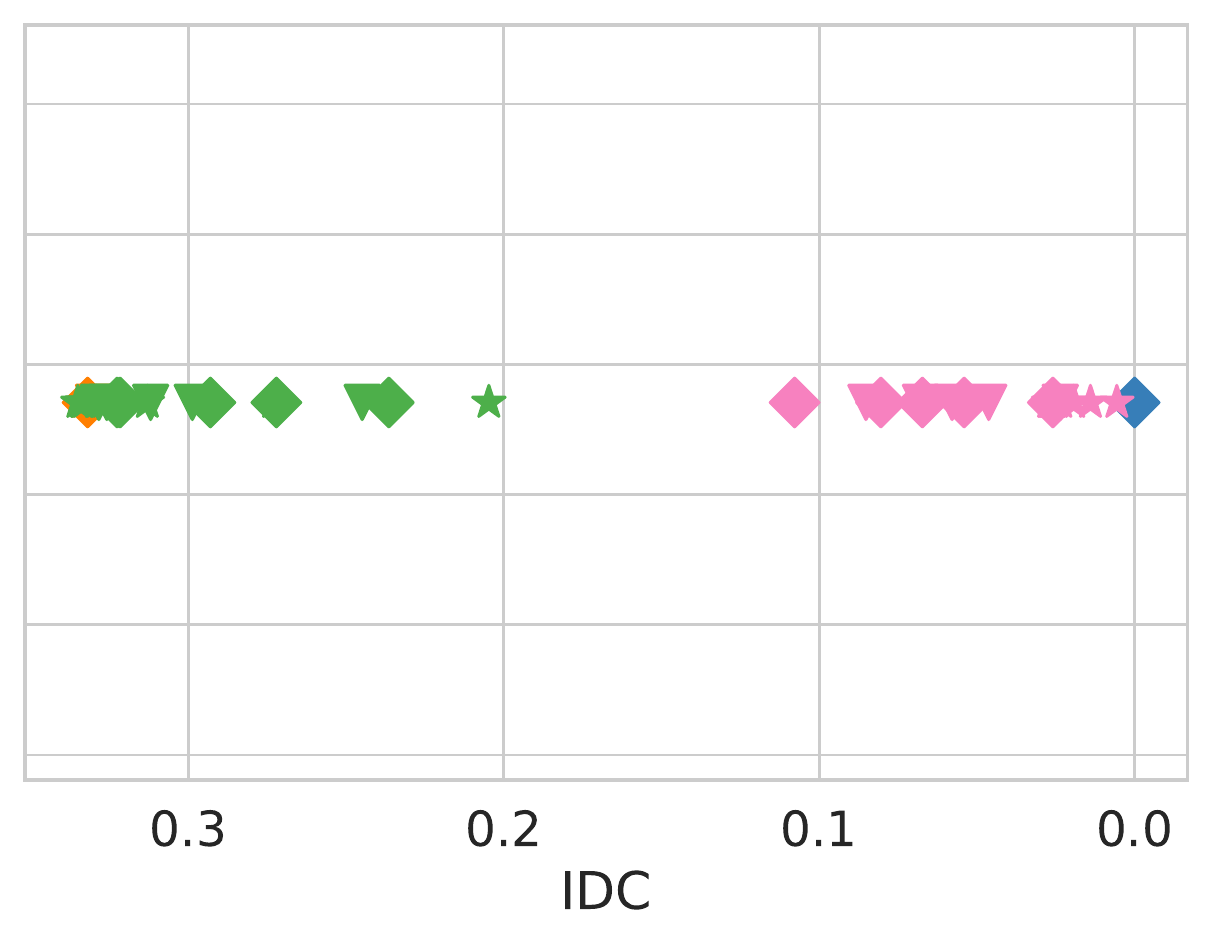}
}

    \caption{
    Plotted are the offending task for the strangeness in the NPT Transformer correlation.
    Those tasks seem to rely on the position of inputs more then other tasks which would explain the comparatively poor performance of the NPT Transformer.
    }
    \label{fig:npt_transformer_individual_tasks}
\end{figure}

\begin{figure}[H]
    \centering
    \includegraphics[width=0.37\textwidth]{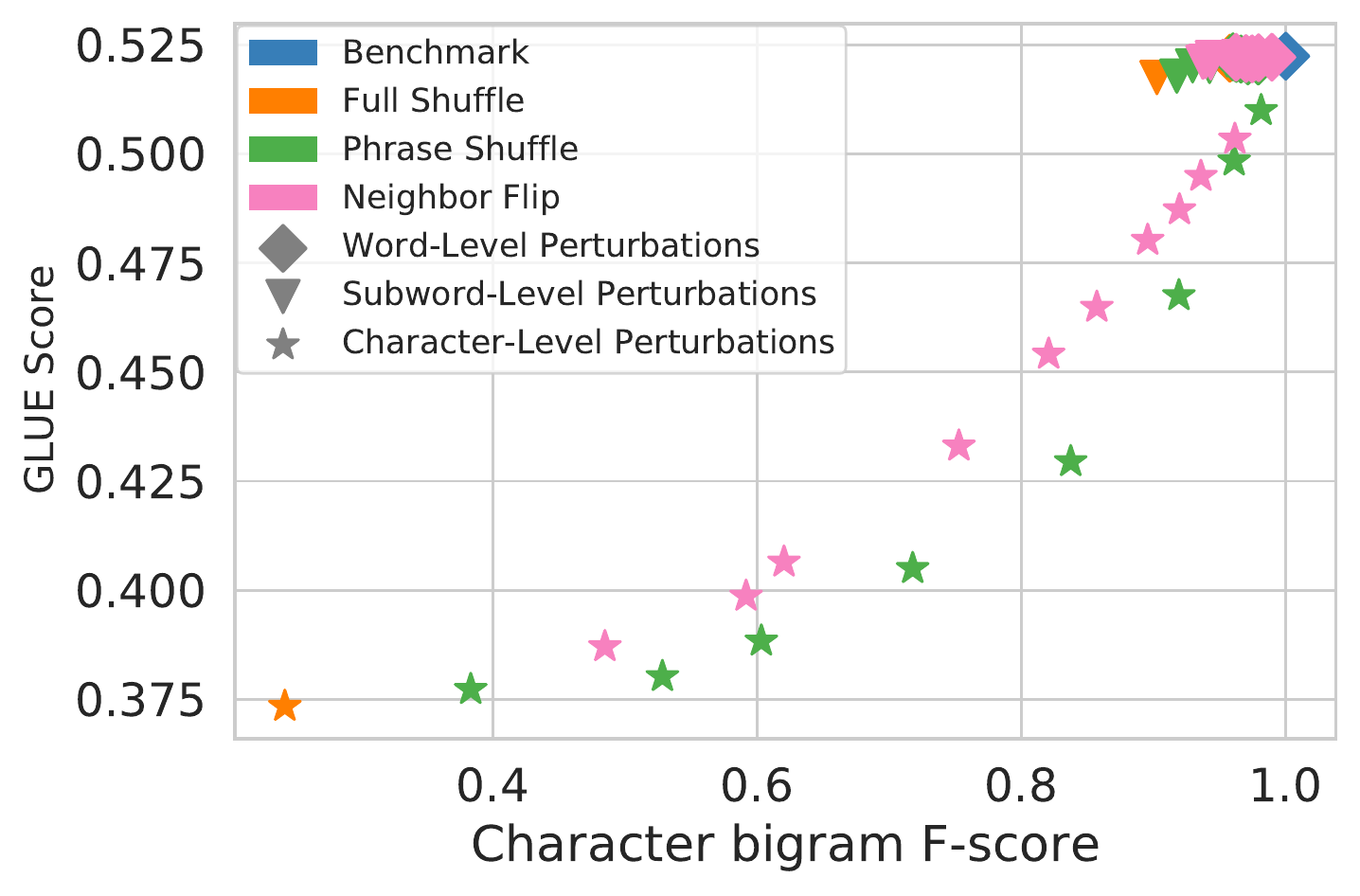}\hfill
    \includegraphics[width=0.31\textwidth]{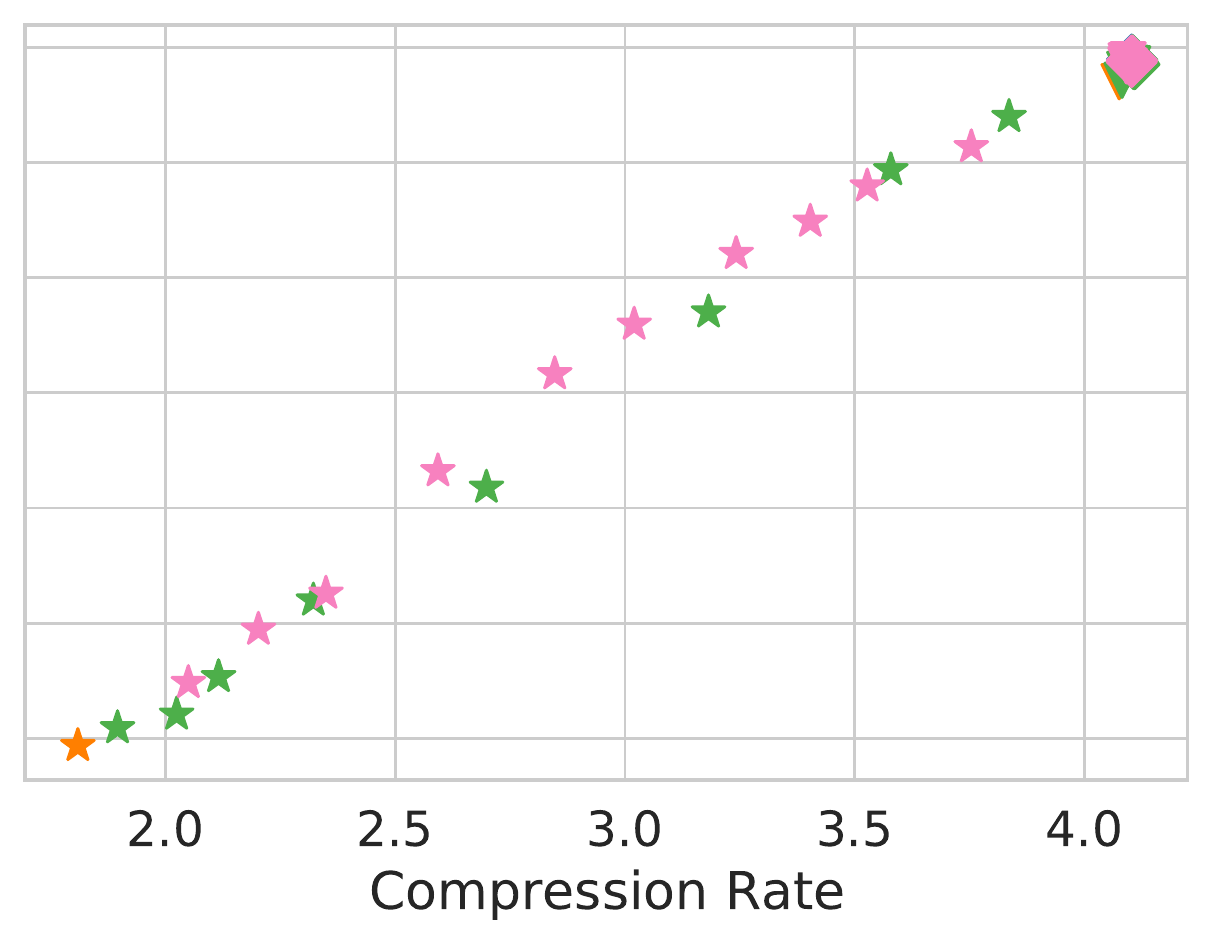}\hfill
    \includegraphics[width=0.31\textwidth]{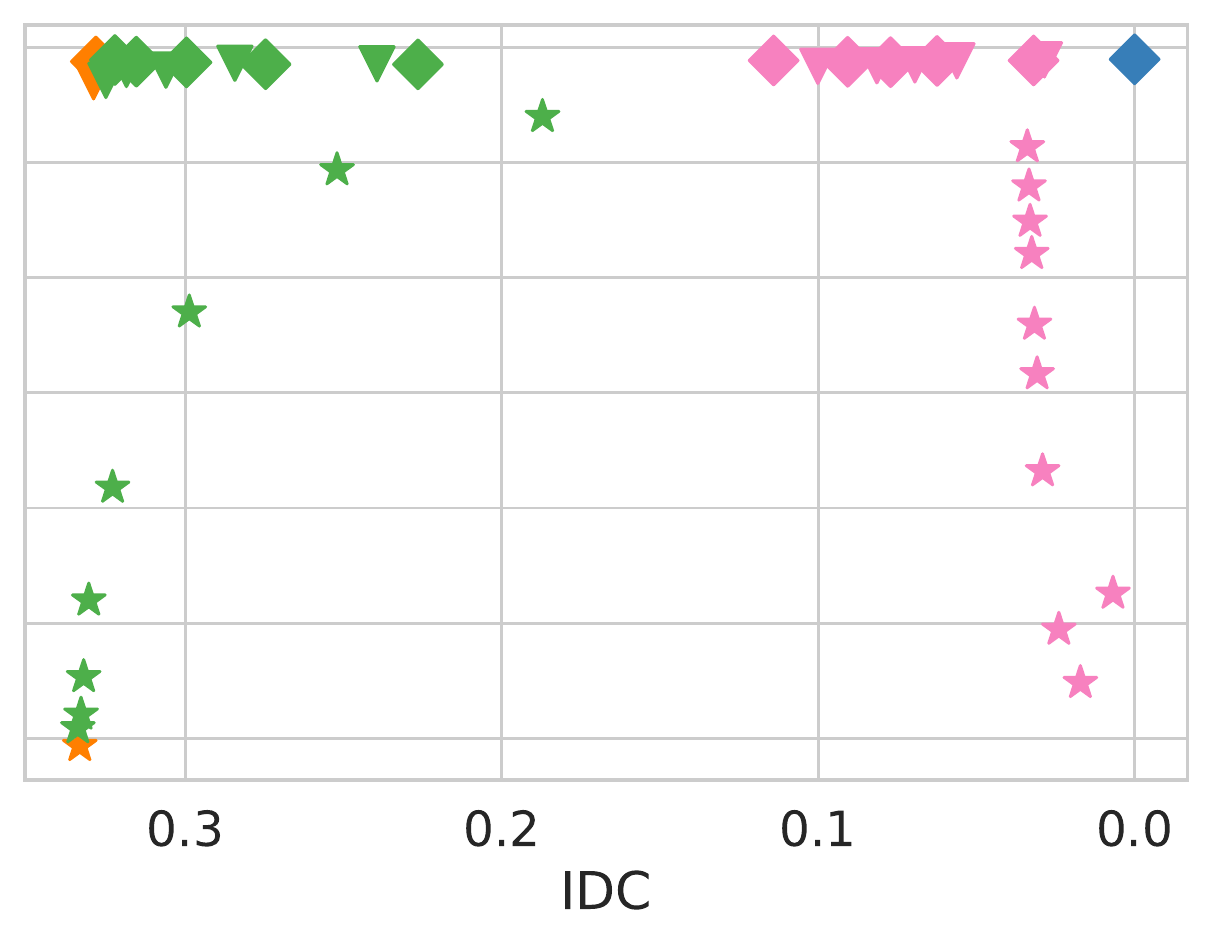}

    \caption{
    Plotted are the relations between the different choices of metrics measuring the amount of perturbation and the performance of NPT Transformer with positional embeddings frozen at $0$.
    We observe very similar results to the NPT Transformers with positional embeddings.
    }
    \label{fig:transformer_no_positional_embedding_metrics}
\end{figure}

\begin{figure}[H]
    \centering
  
    \includegraphics[width=\columnwidth]{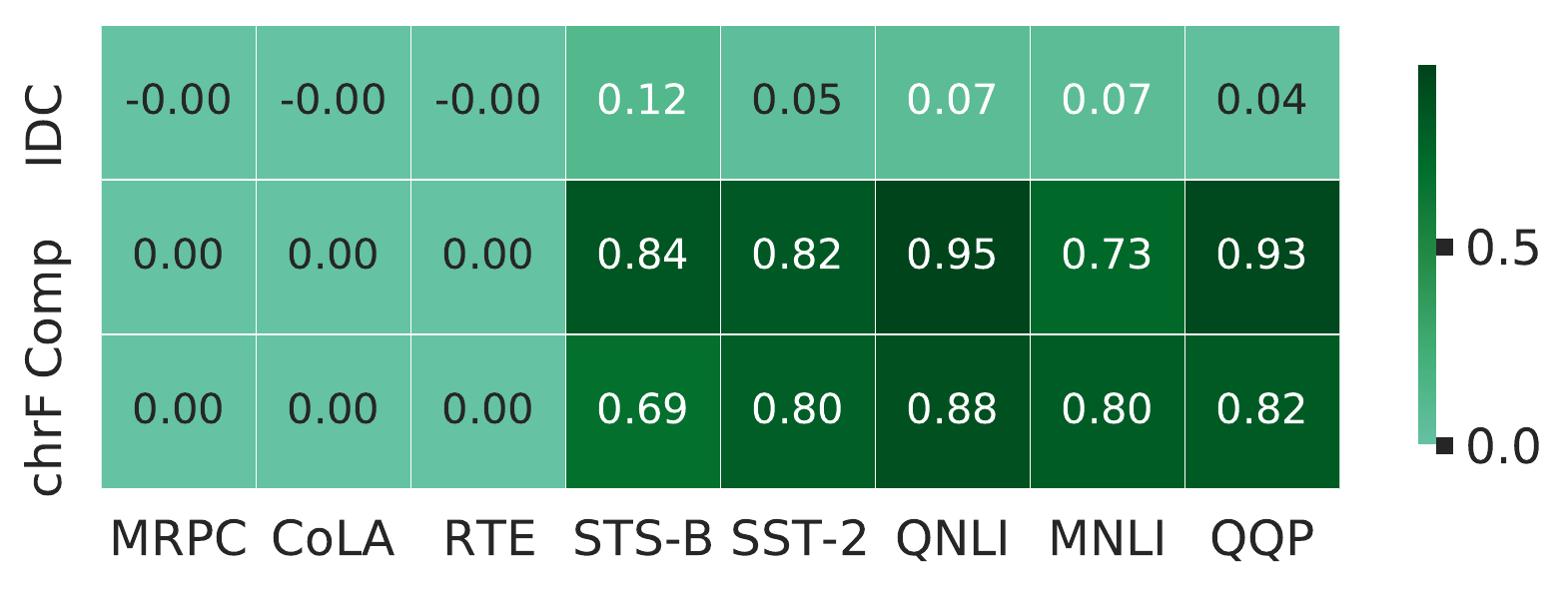}
    \caption{Rank correlation matrix between perturbations measured by different metrics and the performance on the different GLUE tasks of with the NPT Transformer with positional embeddings frozen at $0$.
    We observe very similar results to the NPT Transformers with positional embeddings.
    }
    \label{fig:task_to_scores_transformer_no_positional_embedding}
\end{figure}

The NPT Transformer has many interesting results that warrant additional analysis.
In Figure~\ref{fig:npt_transformer_metrics}, we can observe that no word or subword-level perturbation have any effect on the models performance, which implies that it considers inputs containing the same subwords in any order as equivalent.
In other words, it makes not use of the position of inputs.
Looking at individual tasks in Figure~\ref{fig:task_to_scores_npt_transformer}, we further observe that the correlations to the MRPC, CoLA and RTE tasks are all flat.
By observing those tasks performance individually in~\ref{fig:npt_transformer_individual_tasks}, we can see that the low correlation is simply caused by the fact that the model is incapable to obtain above-chance performances on any of the tasks.
Adding the results of the NPT Transformer with positional embeddings frozen to $0$, in Figure~\ref{fig:transformer_no_positional_embedding_metrics} and Figure~\ref{fig:task_to_scores_transformer_no_positional_embedding}, we can see little difference between the NPT Transformer with and without positional embedding.

\subsection{PT CharBERT}

The PT CharBERT seem roughly inline with the other PT models, with generally more importance to the chrF-$2$ and somewhat less importance to the compression rate.

\begin{figure}[H]
    \centering
    \includegraphics[width=0.35\textwidth]{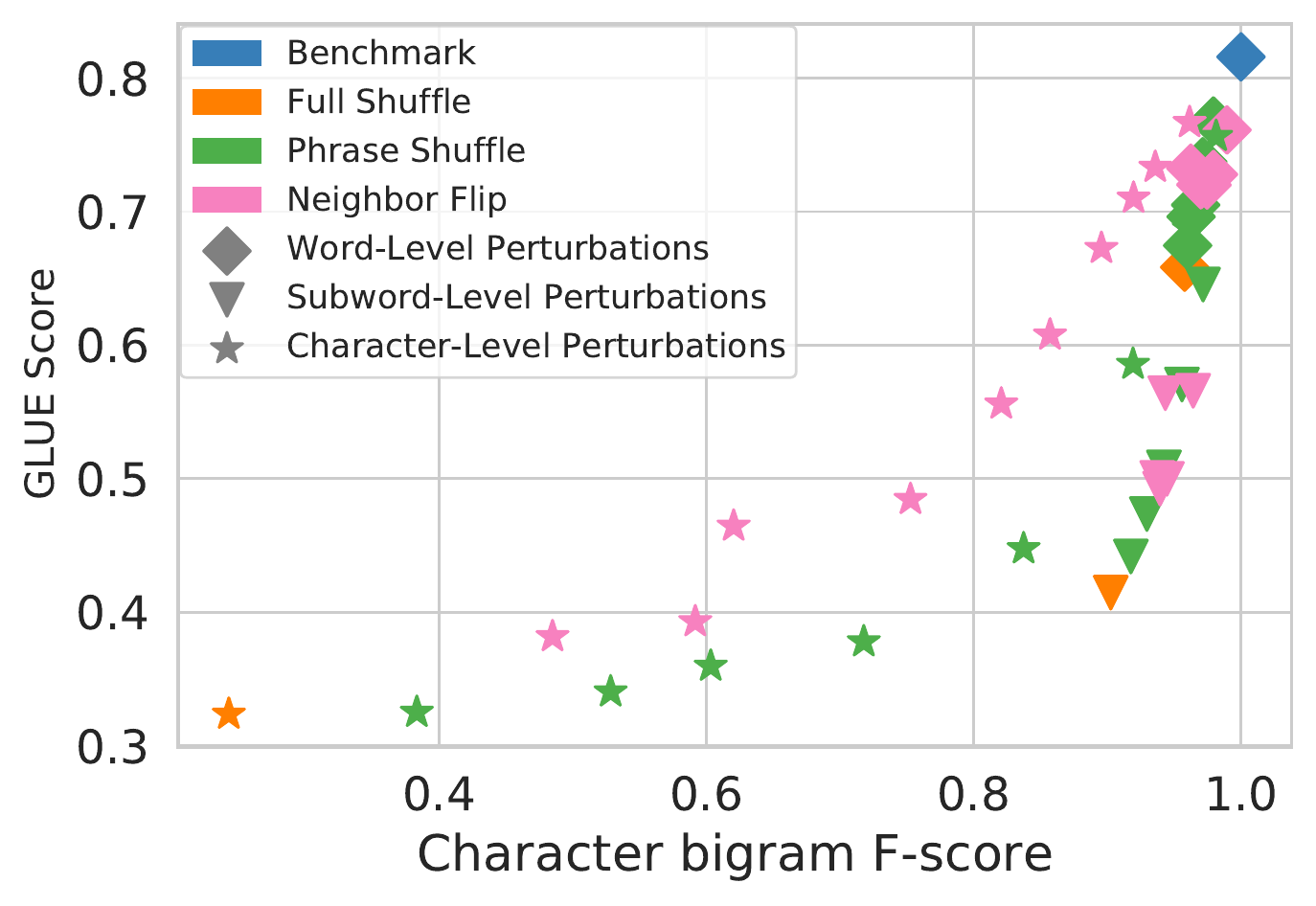}\hfill
    \includegraphics[width=0.31\textwidth]{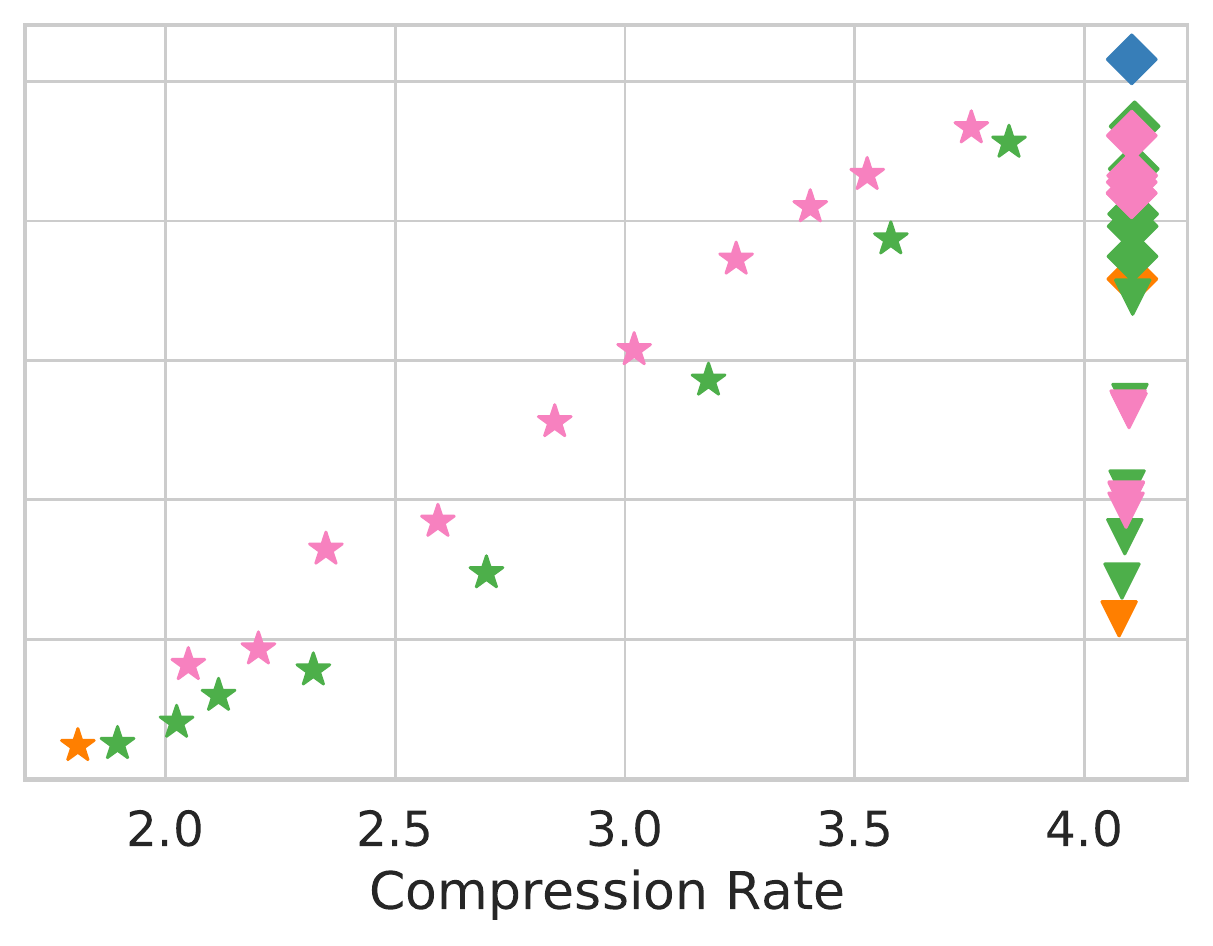}\hfill
    \includegraphics[width=0.31\textwidth]{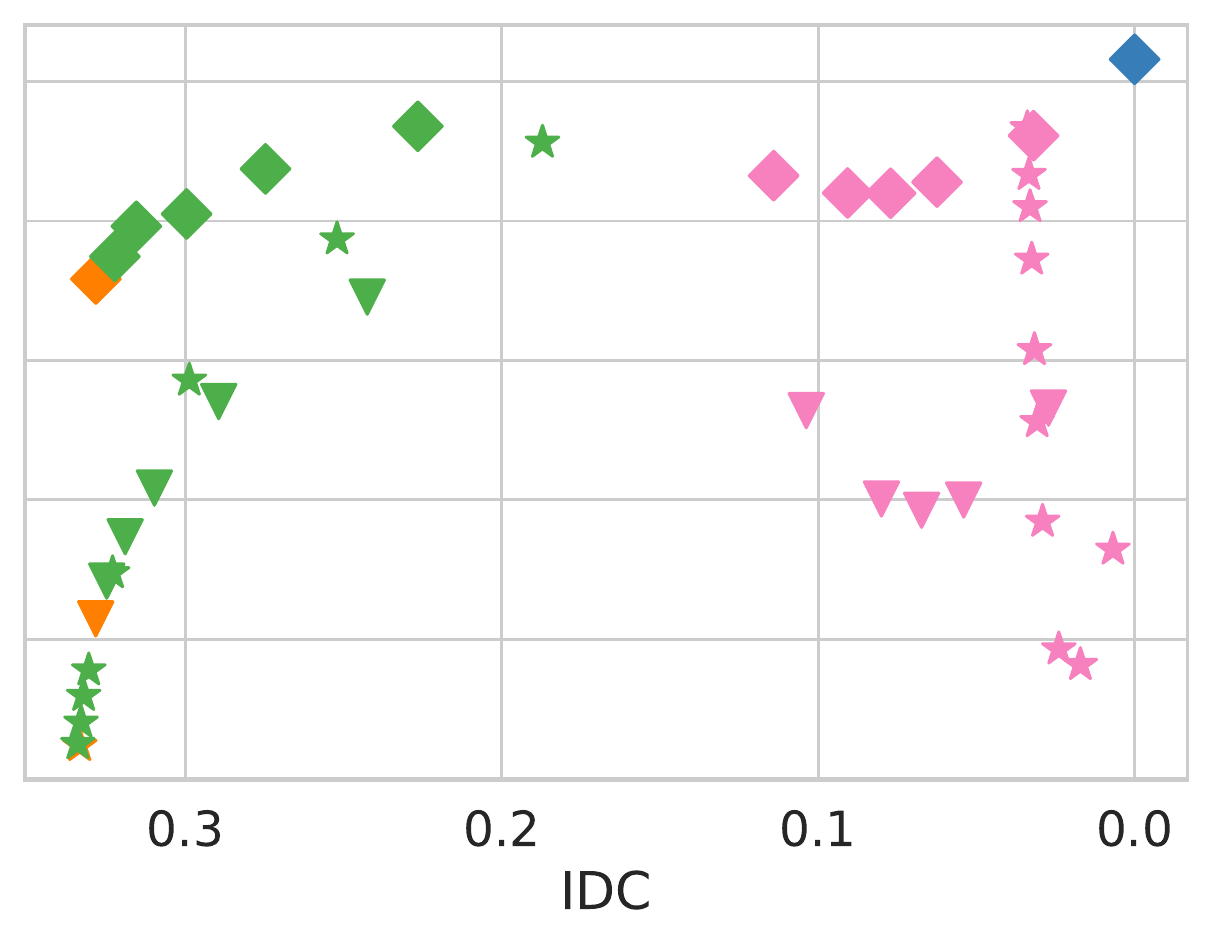}

    \caption{
    Plotted are the relations between the different choices of metrics measuring the amount of perturbation and the performance of PT CharBERT model tested on the perturbed data.
    }
    \label{fig:pt_charbert_metrics}
\end{figure}

\begin{figure}[H]
    \centering
  
    \includegraphics[width=\columnwidth]{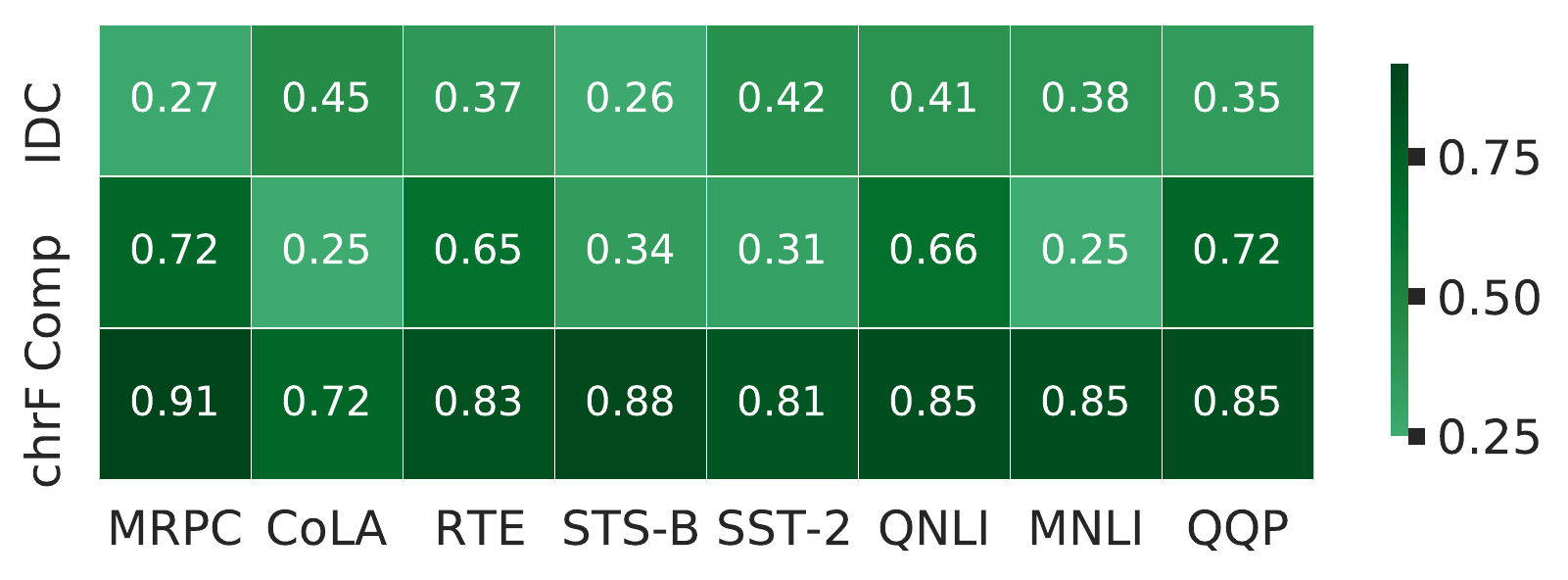}
    \caption{Rank correlation matrix between perturbations measured by different metrics and the performance on the different GLUE tasks of the PT CharBERT model.
    }
    \label{fig:task_to_scores_pt_charbert}
\end{figure}

\subsection{ConvNet}
The ConvNet is inline with other models, with the exception that it fails to obtain any kind of performance on the RTE task.

\begin{figure}[H]
    \centering
    \includegraphics[width=0.36\textwidth]{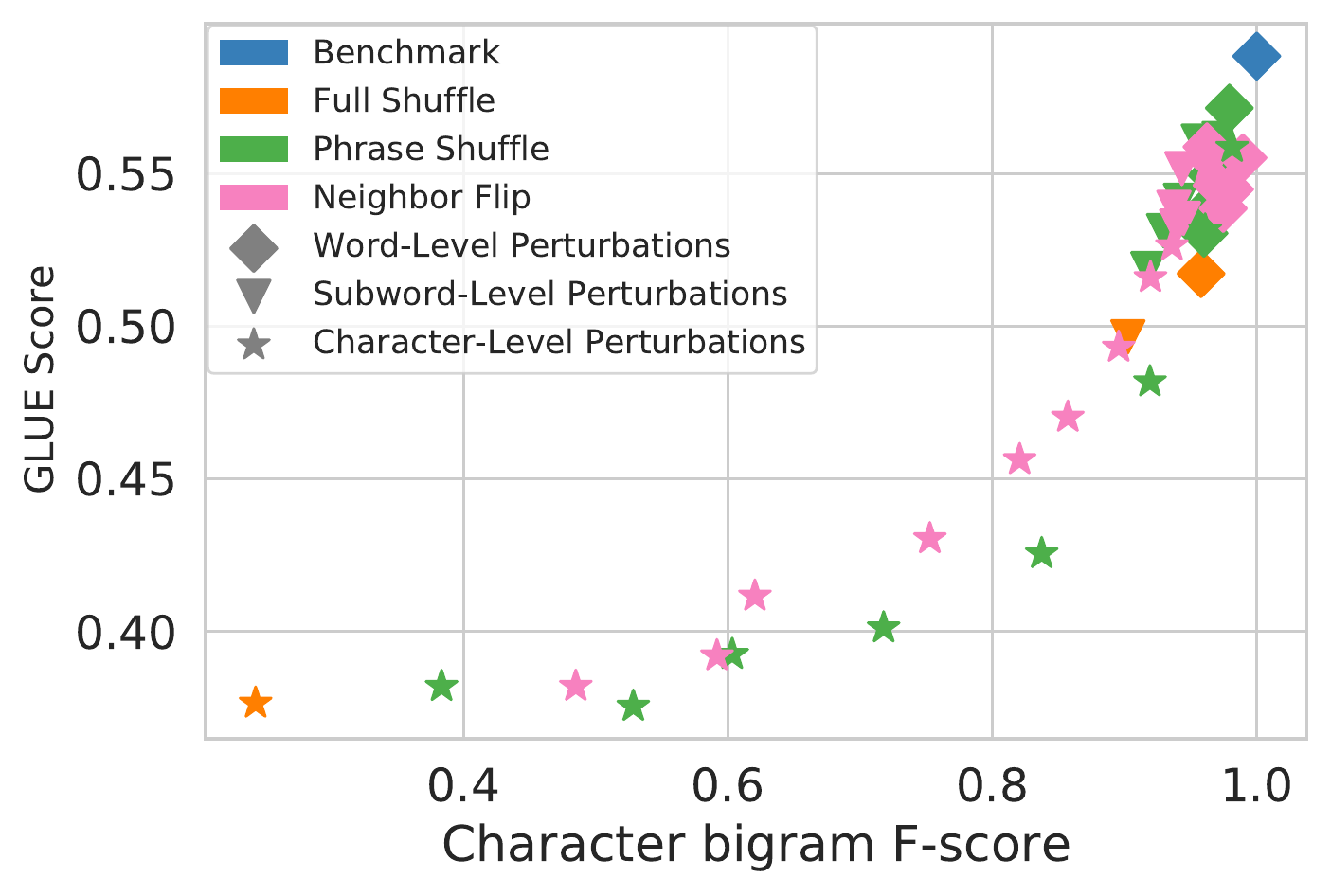}\hfill
    \includegraphics[width=0.31\textwidth]{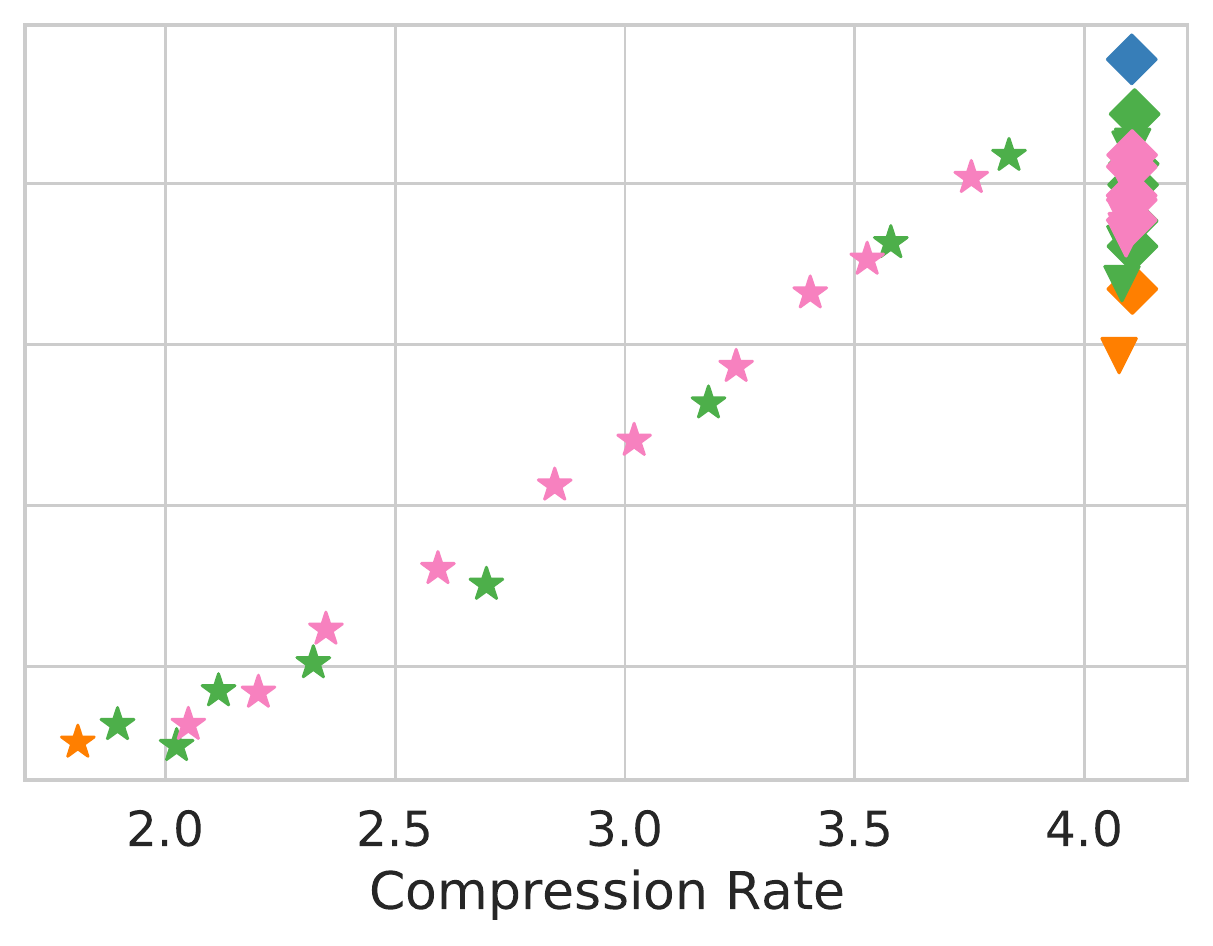}\hfill
    \includegraphics[width=0.31\textwidth]{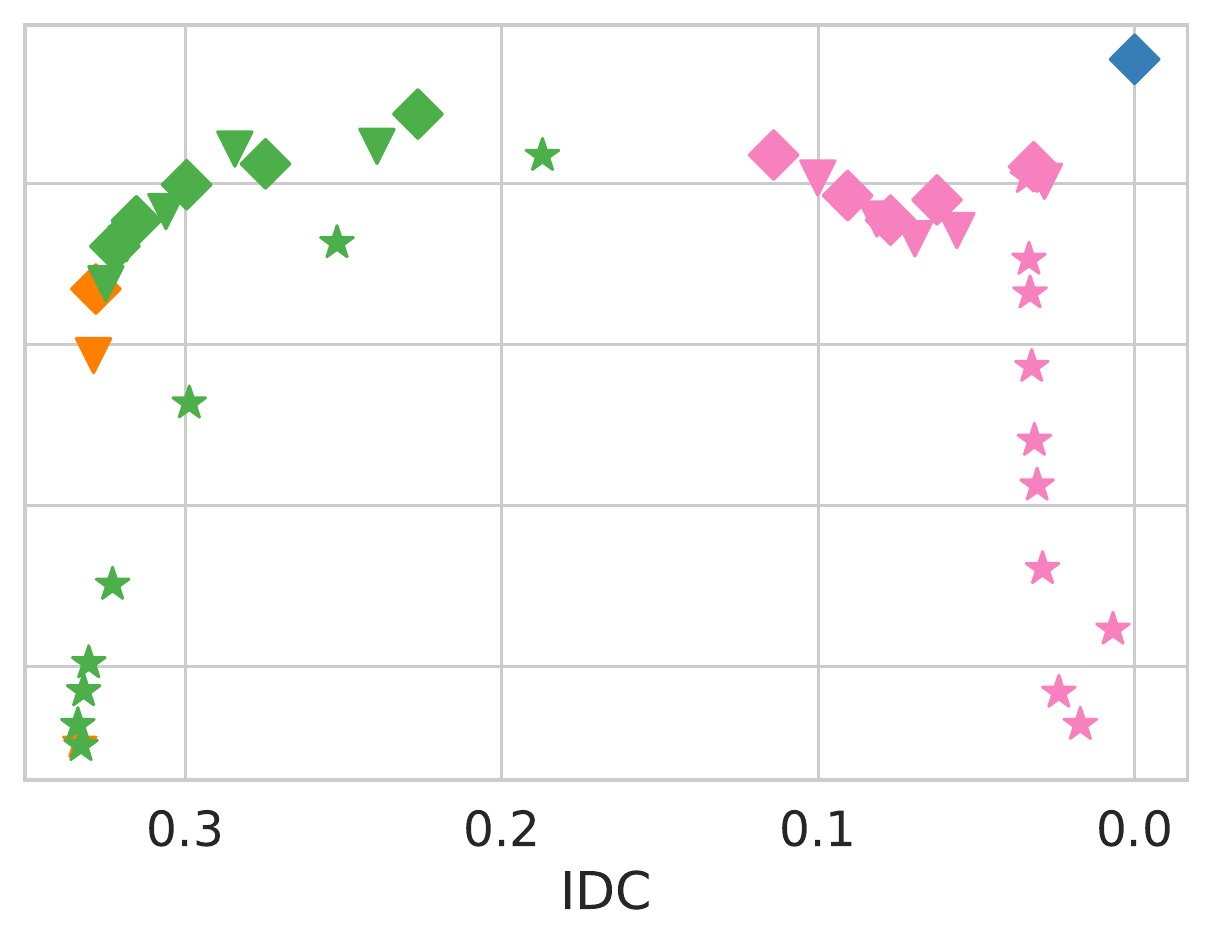}

    \caption{
    Plotted are the relations between the different choices of metrics measuring the amount of perturbation and the performance of ConvNet model tested on the perturbed data.
    }
    \label{fig:convnet_metrics}
\end{figure}

\begin{figure}[H]
    \centering
  
    \includegraphics[width=\columnwidth]{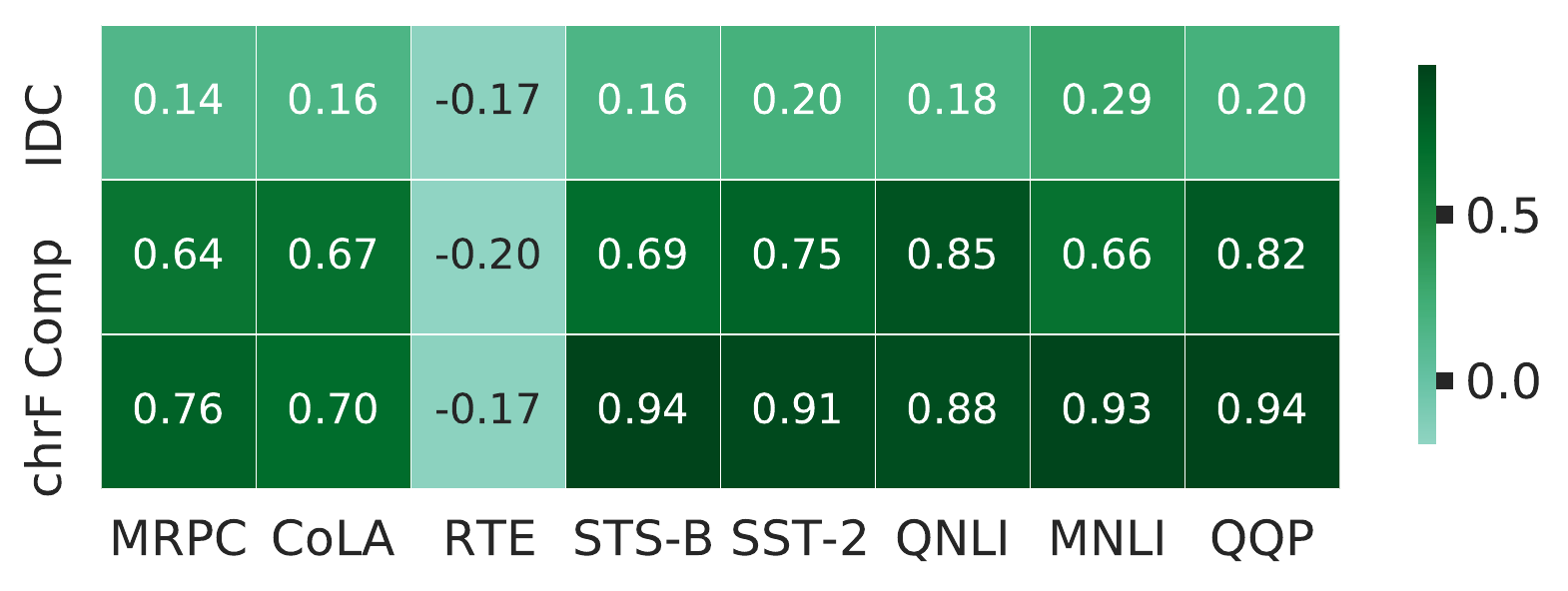}
    \caption{Rank correlation matrix between perturbations measured by different metrics and the performance on the different GLUE tasks of the ConvNet model.
    Much like the NPT Transformer, it is unable to obtain above chance-level on the RTE task.
    }
    \label{fig:task_to_scores_convnet}
\end{figure}

\subsection{BiLSTM}
The BiLSTM is inline with other models performances.
\begin{figure}[H]
    \centering
    \includegraphics[width=0.36\textwidth]{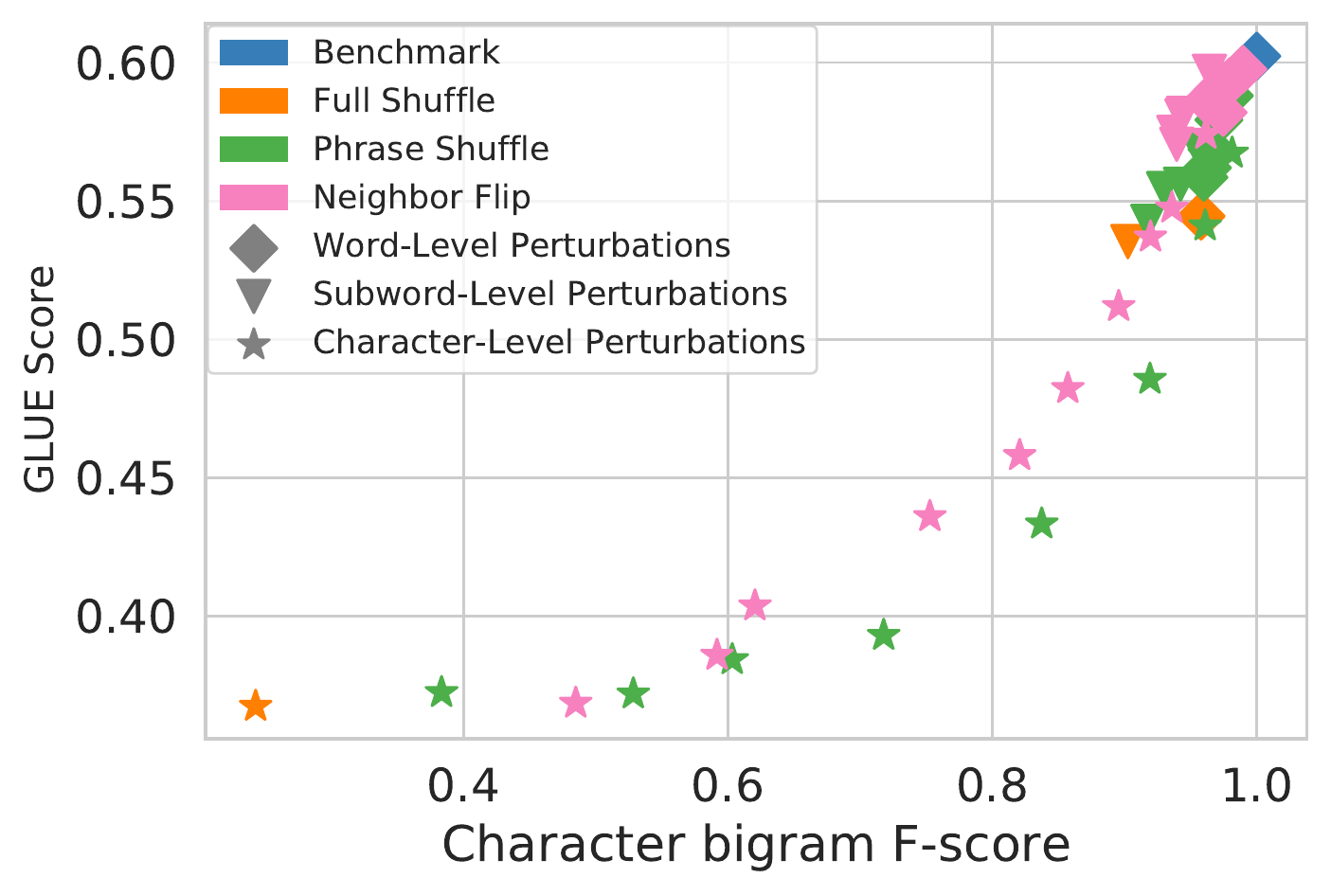}\hfill
    \includegraphics[width=0.31\textwidth]{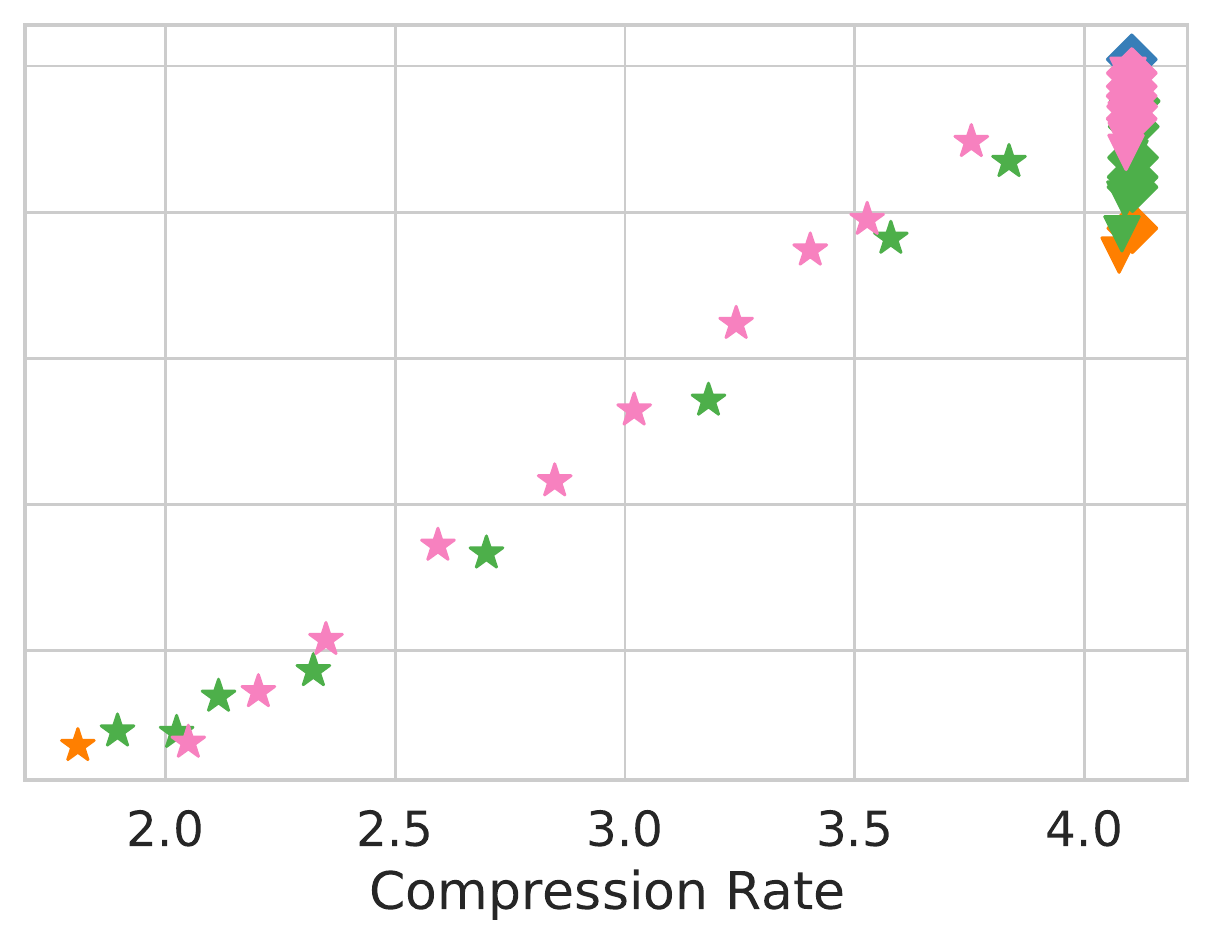}\hfill
    \includegraphics[width=0.31\textwidth]{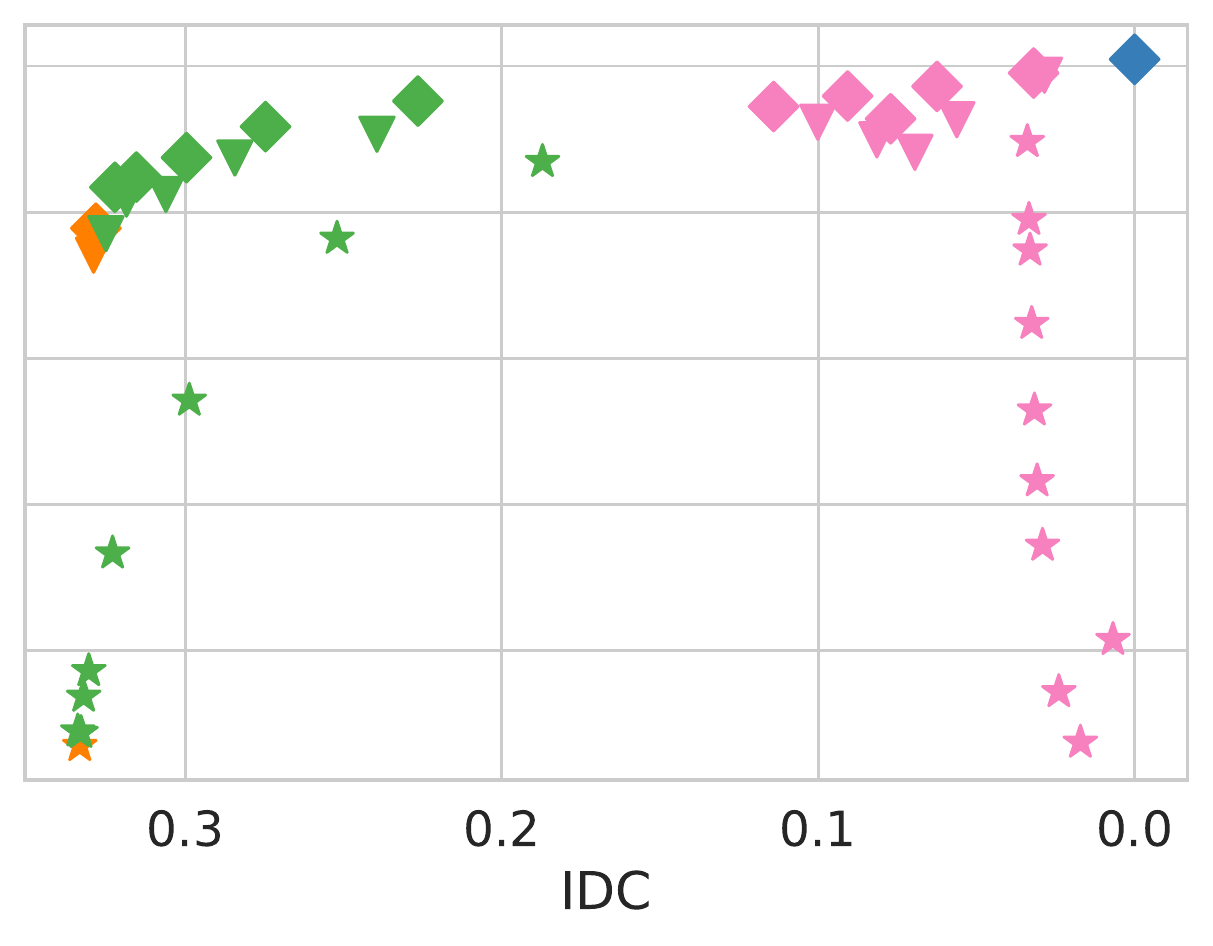}

    \caption{
    Plotted are the relations between the different choices of metrics measuring the amount of perturbation and the performance of BiLSTM model tested on the perturbed data.
    }
    \label{fig:bilstm_metrics}
\end{figure}

\begin{figure}[H]
    \centering
  
    \includegraphics[width=\columnwidth]{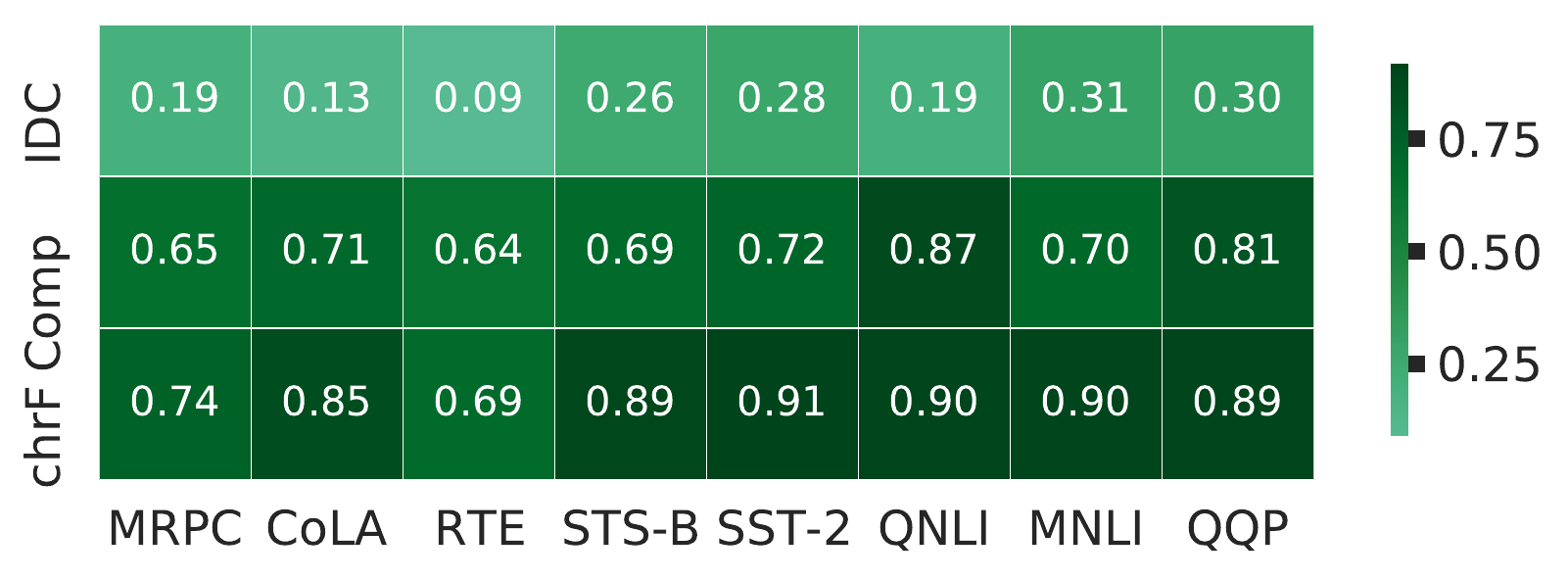}
    \caption{Rank correlation matrix between perturbations measured by different metrics and the performance on the different GLUE tasks of the BiLSTM model.
    }
    \label{fig:task_to_scores_bilstm}
\end{figure}

\subsection{ConvNet Character Embeddings}
\begin{figure}[H]
    \centering
    \includegraphics[width=0.36\textwidth]{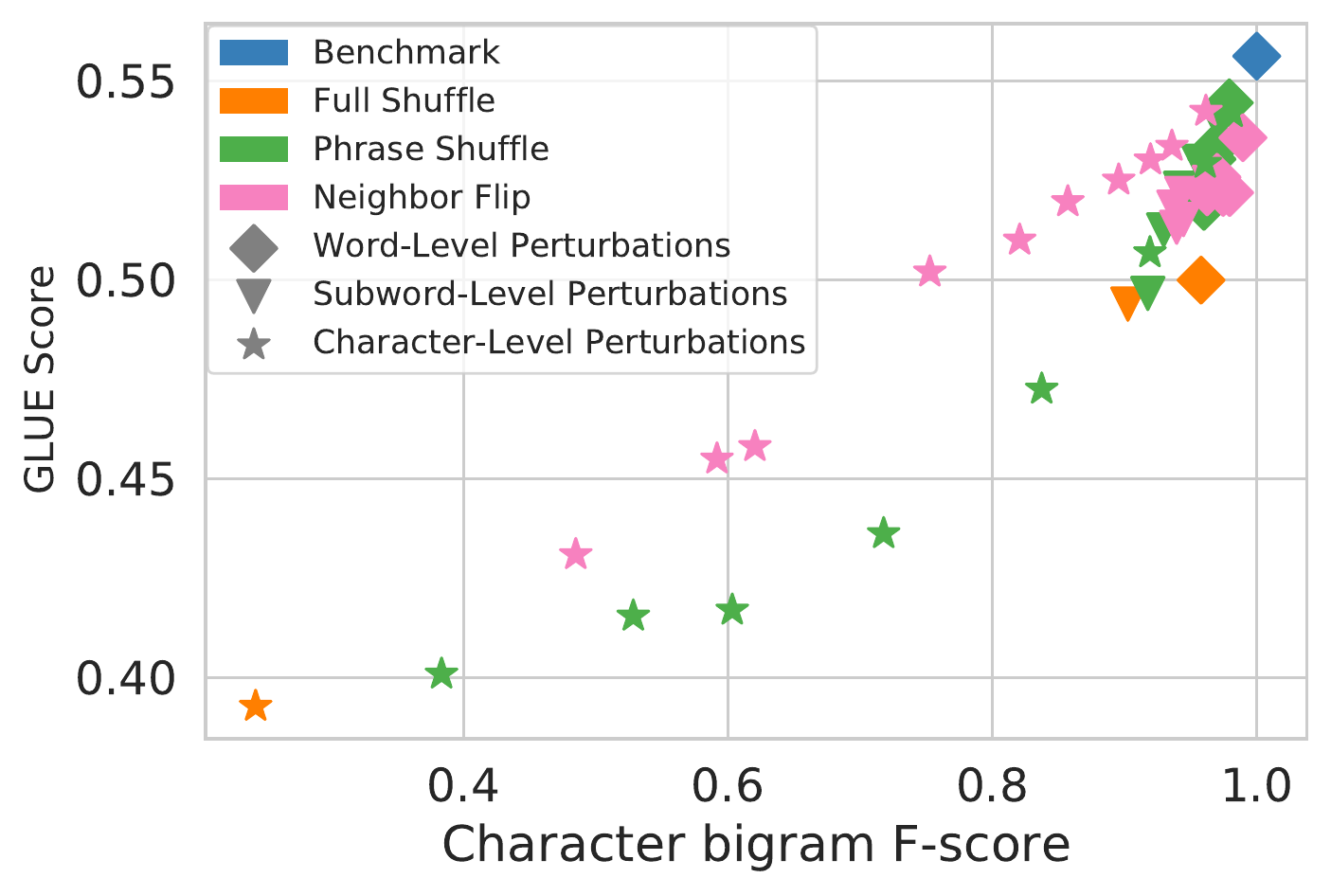}\hfill
    \includegraphics[width=0.31\textwidth]{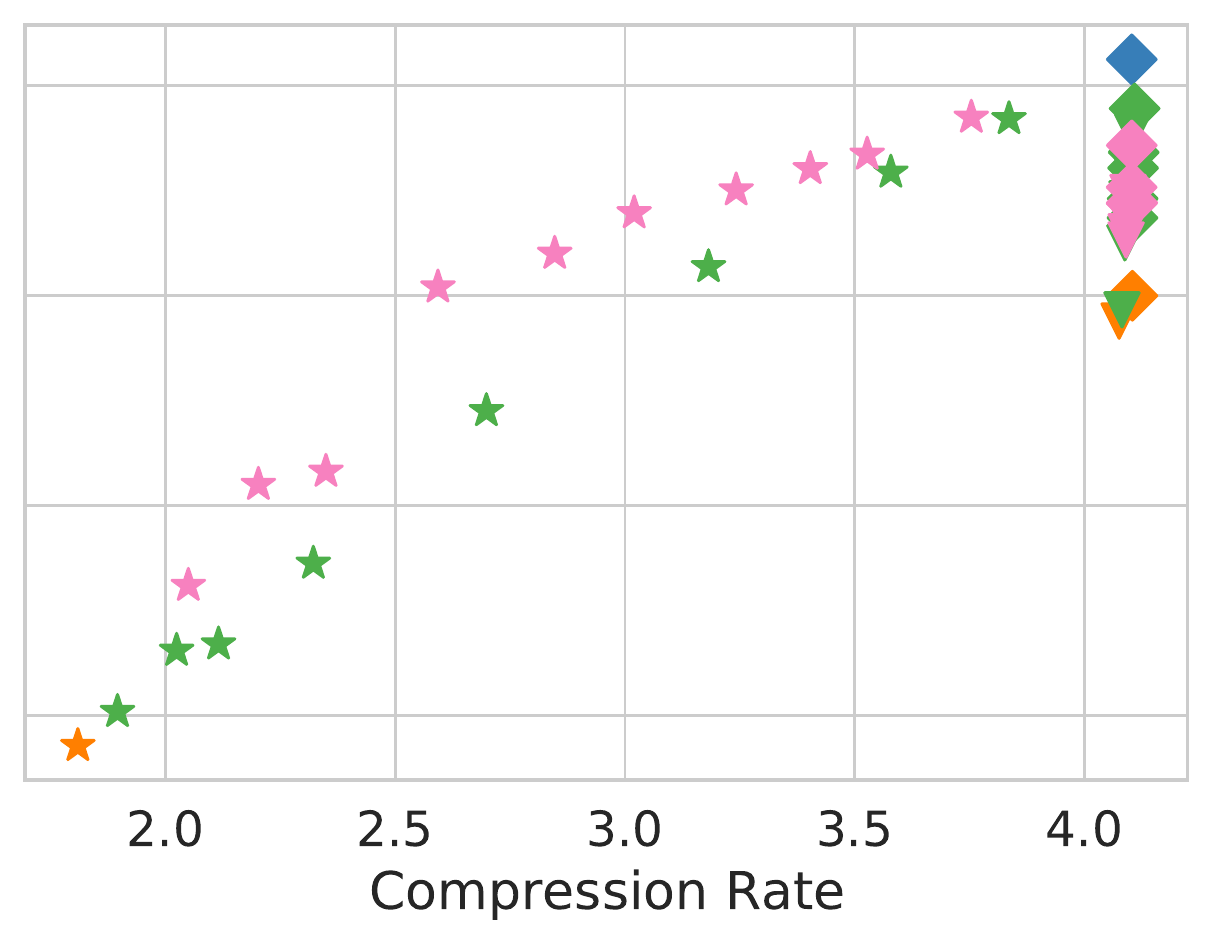}\hfill
    \includegraphics[width=0.31\textwidth]{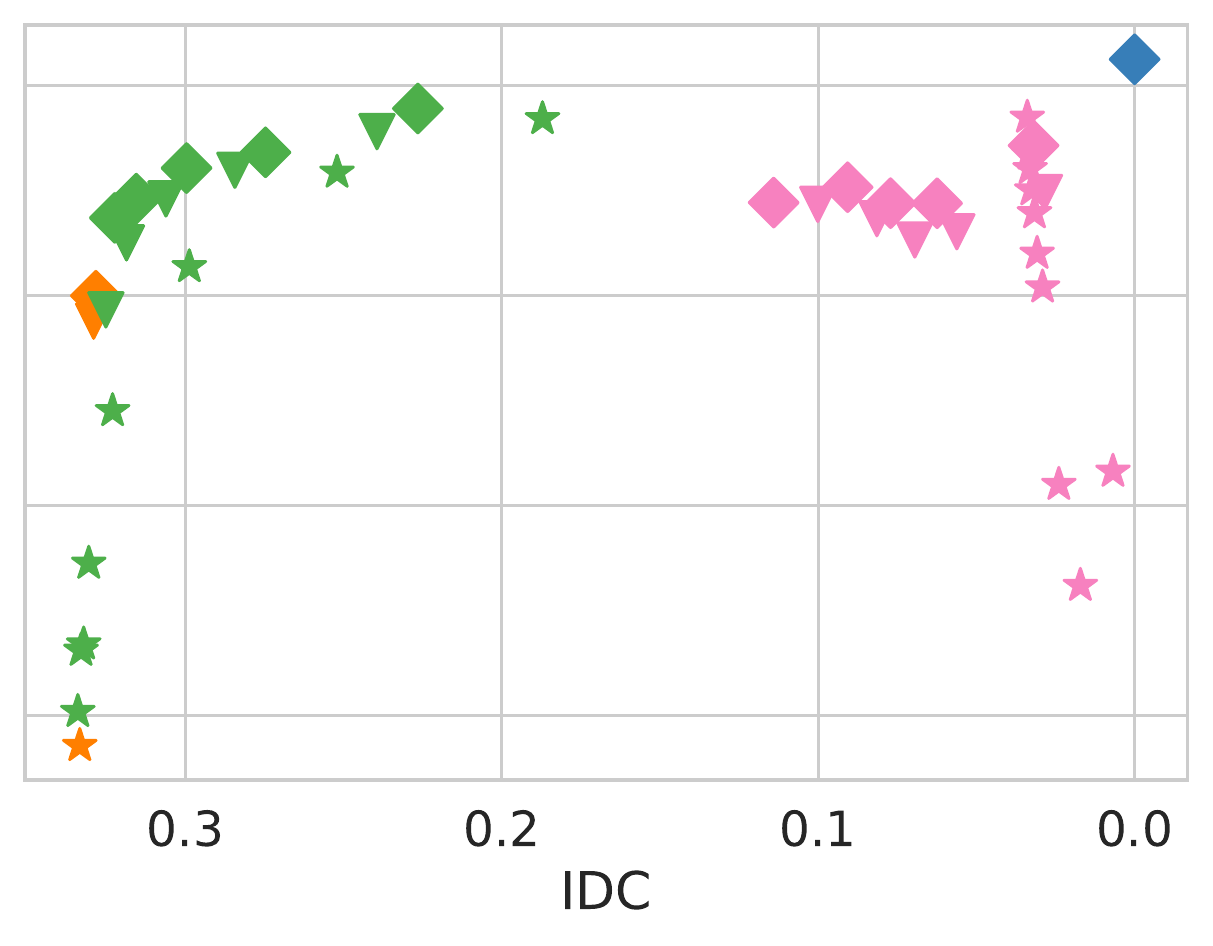}

    \caption{
    Plotted are the relations between the different choices of metrics measuring the amount of perturbation and the performance of BiLSTM model tested on the perturbed data.
    }
    \label{fig:bilstm_metrics}
\end{figure}

\begin{figure}[H]
    \centering
  
    \includegraphics[width=\columnwidth]{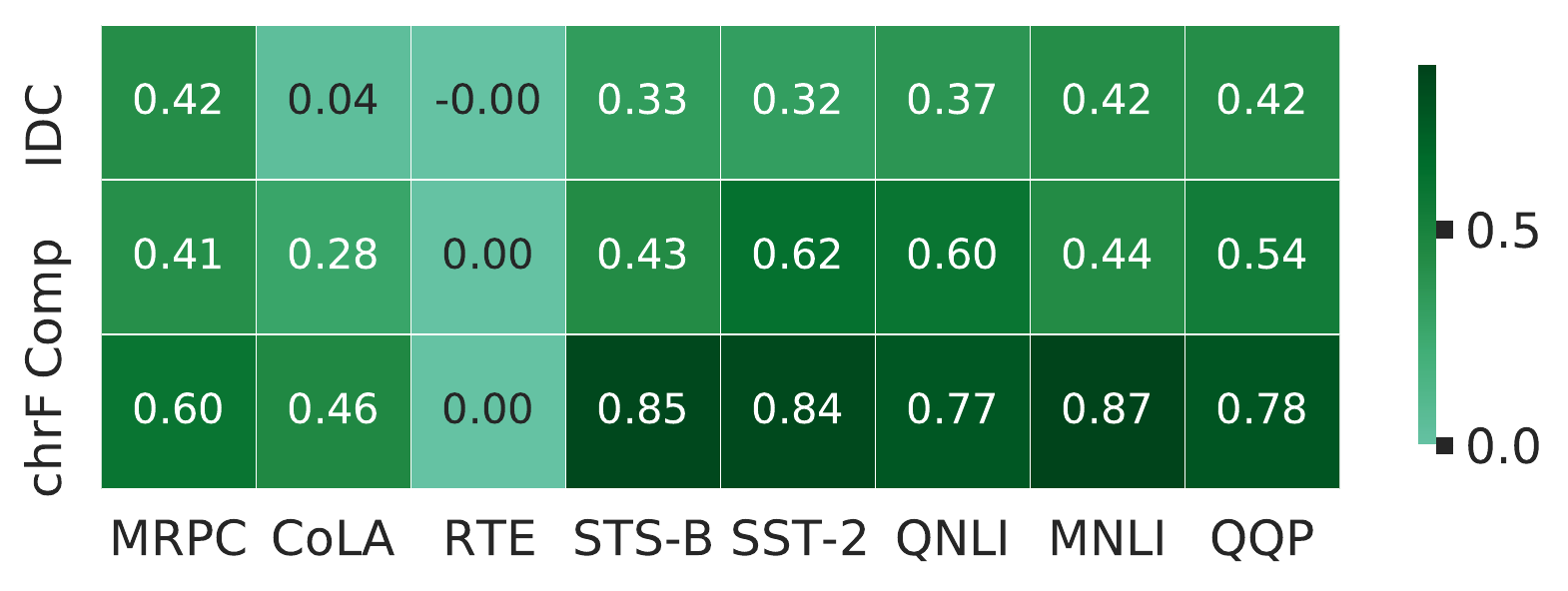}
    \caption{Rank correlation matrix between perturbations measured by different metrics and the performance on the different GLUE tasks of the BiLSTM model.
    }
    \label{fig:task_to_scores_bilstm}
\end{figure}

\subsection{BiLSTM with Character Embeddings}
The BiLSTM with Character Embeddings results seem roughly inline with the other models, with some failures on the CoLA, MRPC and RTE tasks.

\begin{figure}[H]
    \centering
    \includegraphics[width=0.36\textwidth]{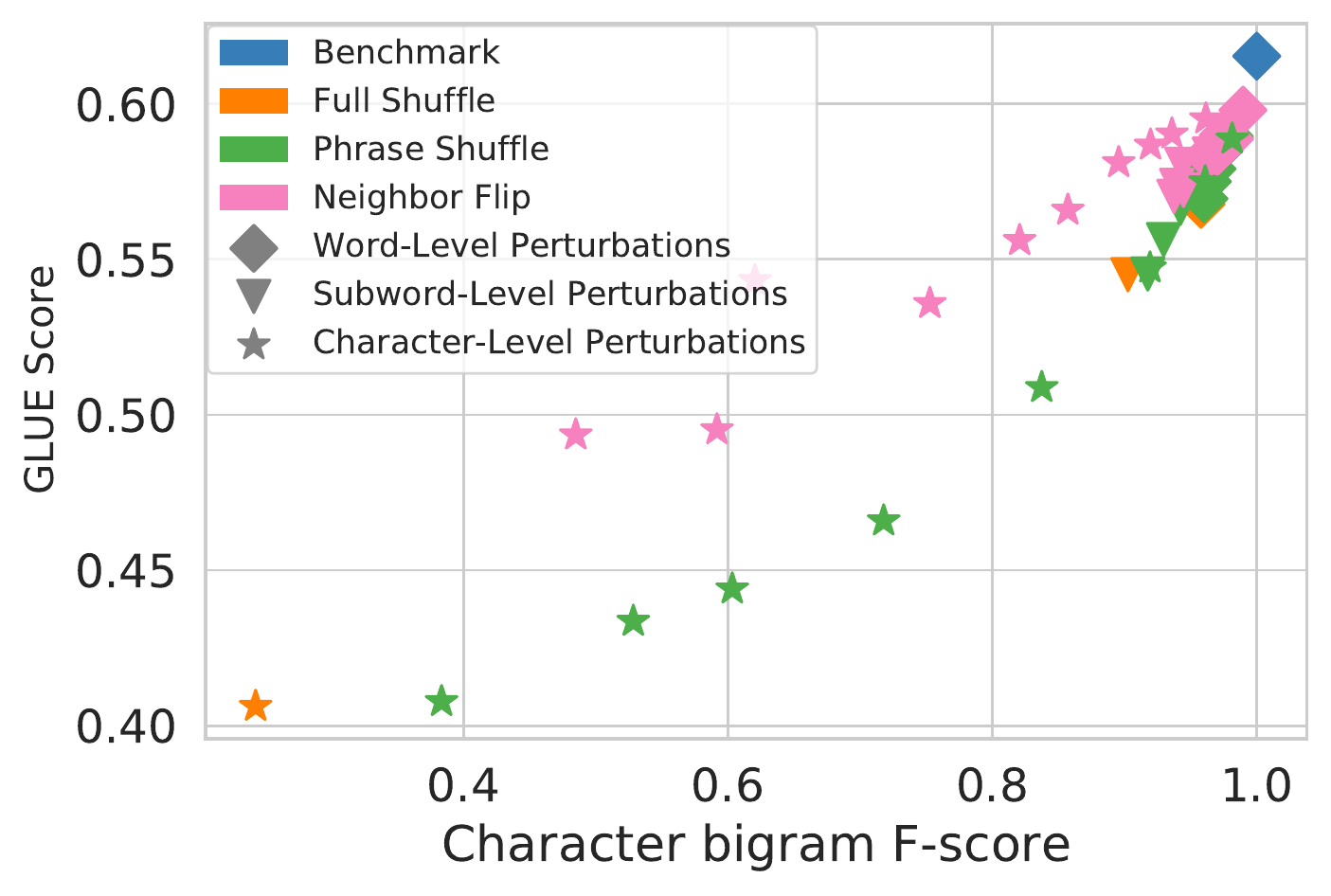}\hfill
    \includegraphics[width=0.31\textwidth]{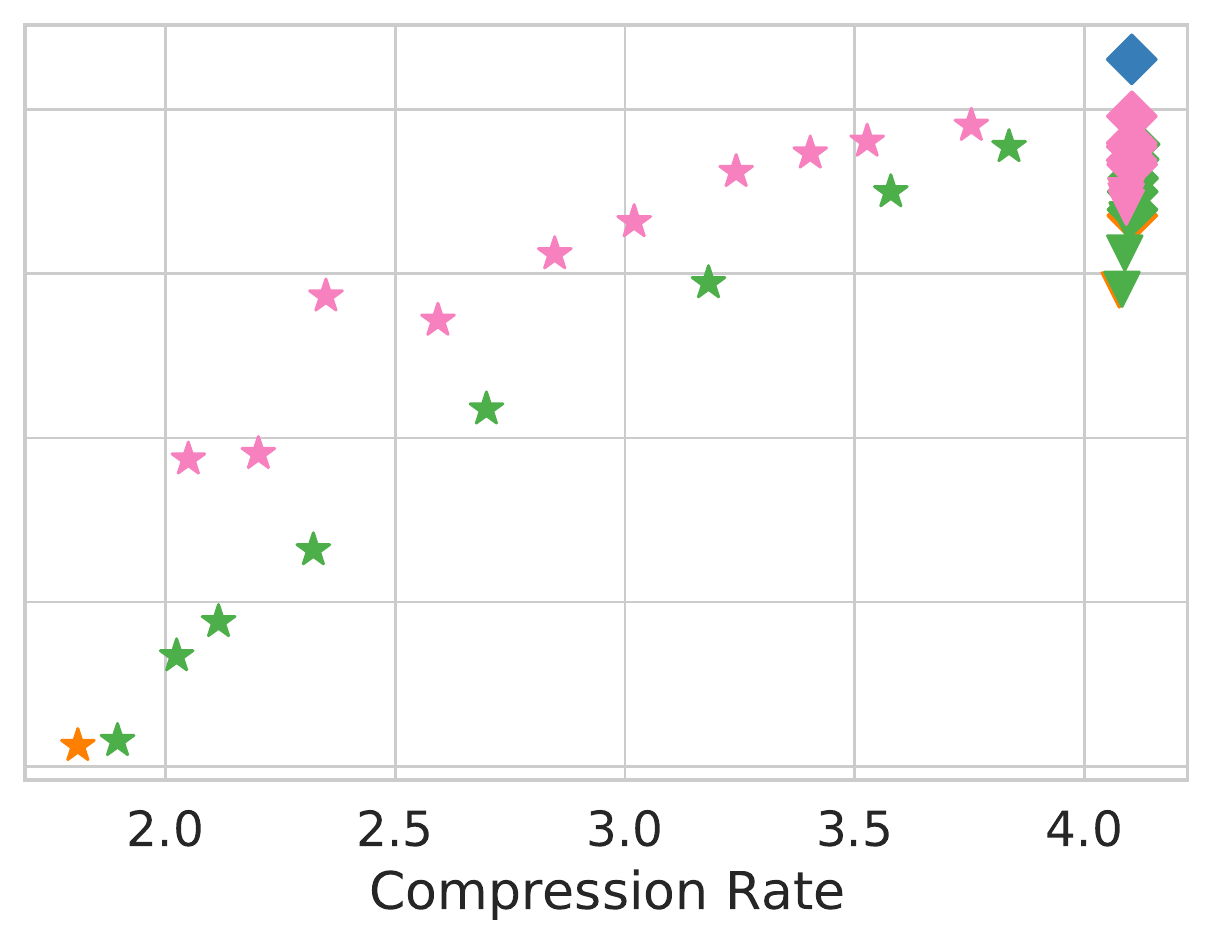}\hfill
    \includegraphics[width=0.31\textwidth]{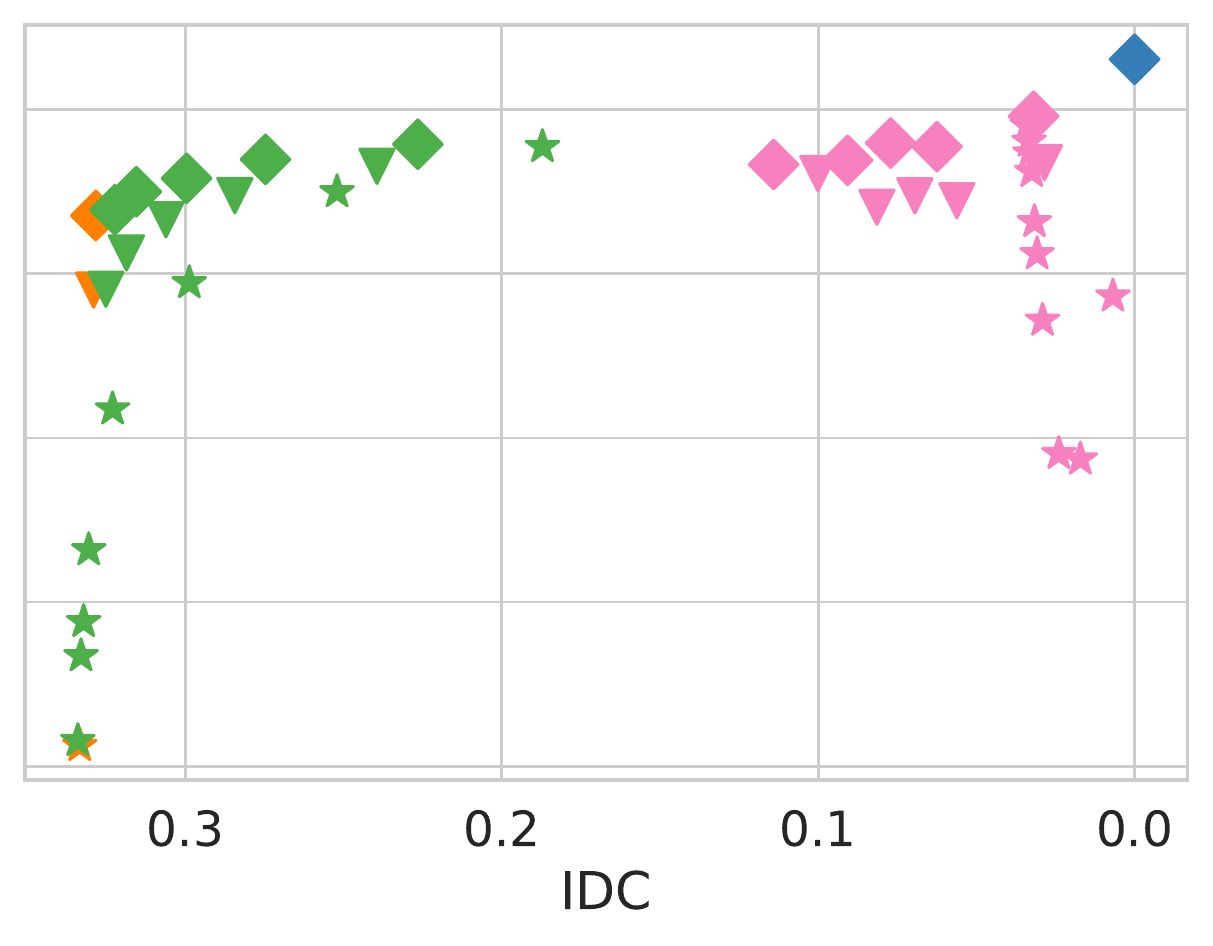}

    \caption{
    Plotted are the relations between the different choices of metrics measuring the amount of perturbation and the performance of BiLSTM with character embeddings model tested on the perturbed data.
    }
    \label{fig:bilstm_char_embeddings_metrics}
\end{figure}

\begin{figure}[H]
    \centering
  
    \includegraphics[width=\columnwidth]{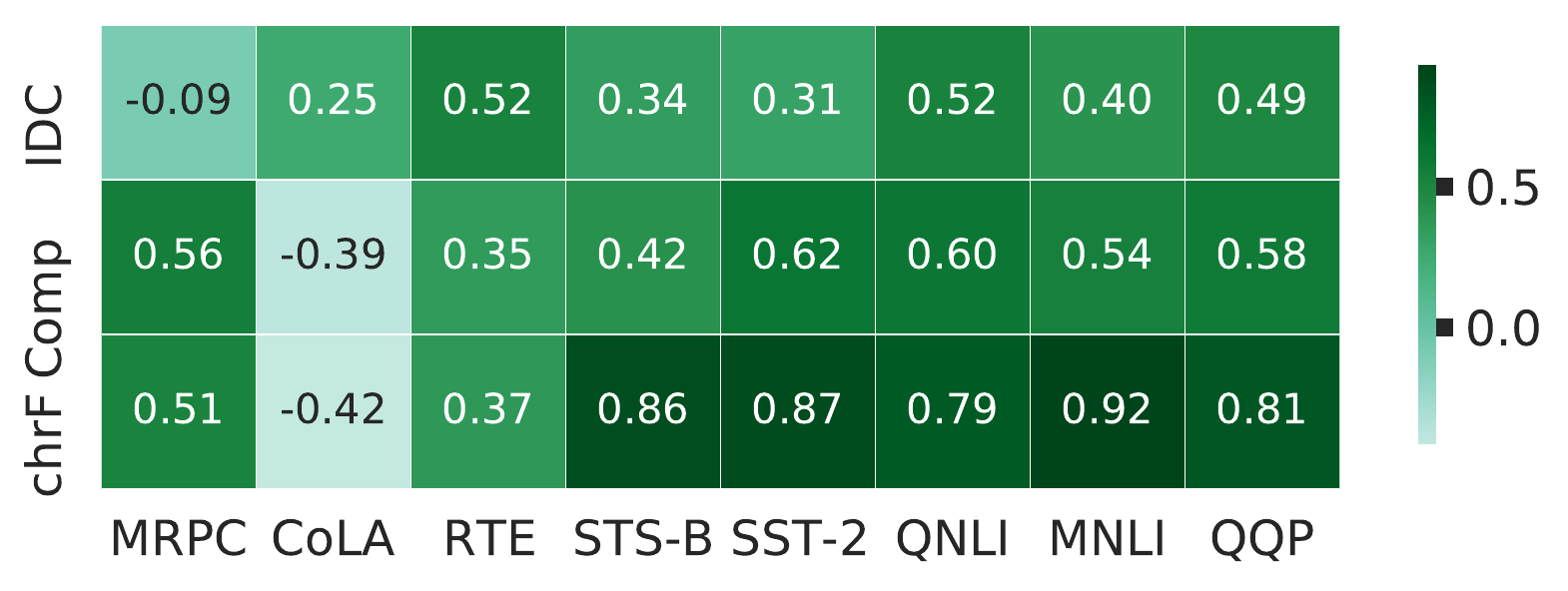}
    \caption{Rank correlation matrix between perturbations measured by different metrics and the performance on the different GLUE tasks of the BiLSTM with character embeddings model.
    In this case, the model struggles on the RTE and CoLA task.
    }
    \label{fig:task_to_scores_bilstm_char_embeddings}
\end{figure}

\end{document}